%% file: LatentPersonal.tex
\documentclass{article} 
\usepackage{iclr2027_conference,times}

\input{math_commands.tex}

\usepackage{hyperref}
\usepackage{url}

\usepackage{graphicx}
\usepackage{wrapfig}
\usepackage{algorithm}
\usepackage{algorithmic}
\usepackage{enumitem}
\usepackage{booktabs}
\usepackage{multirow}
\usepackage{subcaption}

\title{
Unlocking Latent Personalization in LLMs
}

\author{
Wei Chen,\quad 
Guanghui Zhu\thanks{Corresponding author.},\quad
Zhongliang Cai,\quad
Yihua Huang \\
State Key Laboratory for Novel Software Technology,\quad Nanjing University\\
\texttt{weichen@smail.nju.edu.cn,\quad zgh@nju.edu.cn}\\
}

\iclrfinalcopy 
\begin{document}

\maketitle
\lhead{}
\begin{abstract}
Large language models (LLMs) are increasingly expected to adapt to individual users, yet effective personalization remains challenging when only limited user-specific samples are available. In this work, we take an alternative perspective: pretrained LLMs may already possess latent capacity for personalization, and a few user samples may therefore suffice to guide the model toward user-aligned behavior with minimal user-specific adaptation. From this perspective, we propose \textit{LatentPersonal}, a framework that formulates personalization as navigation in a shared latent adaptation space. \textit{LatentPersonal} infers a compact latent representation from a few user samples to guide user-specific model adaptation, regularized with a variational information bottleneck to encourage compact preference representations. We instantiate \textit{LatentPersonal} with LoRA, leveraging its low-rank parameterization as a natural low-dimensional adaptation space for personalization. By simply inserting a user-specific guidance vector between the shared low-rank factors, the model can navigate toward personalized adaptations through lightweight inference of this compact representation, without updating the shared LoRA parameters. Experiments across multiple personalization datasets demonstrate that \textit{LatentPersonal} substantially reduces user-specific adaptation overhead while achieving effective personalization from only a few user-specific interactions, with particularly strong performance in the one-shot regime.
\end{abstract}

\input{chapters/introduction}

\input{chapters/related}

\input{chapters/methdology}

\input{chapters/experiment}

\input{chapters/conclusion}

\newpage
\bibliography{iclr2027_conference}
\bibliographystyle{iclr2027_conference}

\newpage
\appendix

\input{appendix/app_method}

\input{appendix/app_lora}

\input{appendix/app_details}

\input{appendix/app_exp}

\input{appendix/app_peft_instantiations}

\input{appendix/app_limitation}

\end{document}

%% file: math_commands.tex
\usepackage{amsmath,amsfonts,bm}

\def\eqref#1{equation~\ref{#1}}

\def\1{\bm{1}}

\DeclareMathAlphabet{\mathsfit}{\encodingdefault}{\sfdefault}{m}{sl}
\SetMathAlphabet{\mathsfit}{bold}{\encodingdefault}{\sfdefault}{bx}{n}



%% file: chapters/introduction.tex
\section{Introduction}
\label{Introduction}
Large language models (LLMs) have achieved remarkable success as general-purpose models and are increasingly being deployed in diverse real-world scenarios~\citep{brown2020language,wei2021finetuned,minaee2024large}, including personal assistants, domain-specific applications, and task-oriented intelligent systems. However, the transition from broadly capable models to user-centric systems introduces a fundamental challenge: a single pretrained model must accommodate heterogeneous preferences, behaviors, and requirements across users~\citep{houlsby2019parameter}. While general-purpose LLMs provide strong foundational capabilities, effectively tailoring them to individual users remains challenging, particularly when only limited user-specific interactions are available, making direct adaptation both data-inefficient and prone to overfitting~\citep{houlsby2019parameter,li2021prefix}.

Existing LLM personalization methods leverage user-specific information, including interaction histories, demonstrations, and feedback~\citep{zollo2025personalllm,aroca2025aligning}, to tailor model behavior to individual users via personalized prompting~\citep{zhang2024p4}, retrieval-based conditioning~\citep{salemi2024lamp}, user representation learning~\citep{liu2025llms+}, and parameter-efficient adaptation~\citep{tan2024democratizing}. Despite their different mechanisms, these approaches largely follow a similar intuition: personalization is achieved by further tailoring model behavior using user-specific information. Yet large-scale pretraining has endowed LLMs with broad and diverse capabilities, many of which may already support personalized responses but are not necessarily expressed in ways that align with individual users. Nevertheless, many existing personalization methods still rely on limited user-specific data to learn user-aligned behavior anew, potentially relearning behaviors that may already be supported by the pretrained model.

This raises a natural question: does effective personalization necessarily require learning user-specific behavior anew from limited user information?

\begin{figure*}[t]
    \centering
    \includegraphics[width=\textwidth]{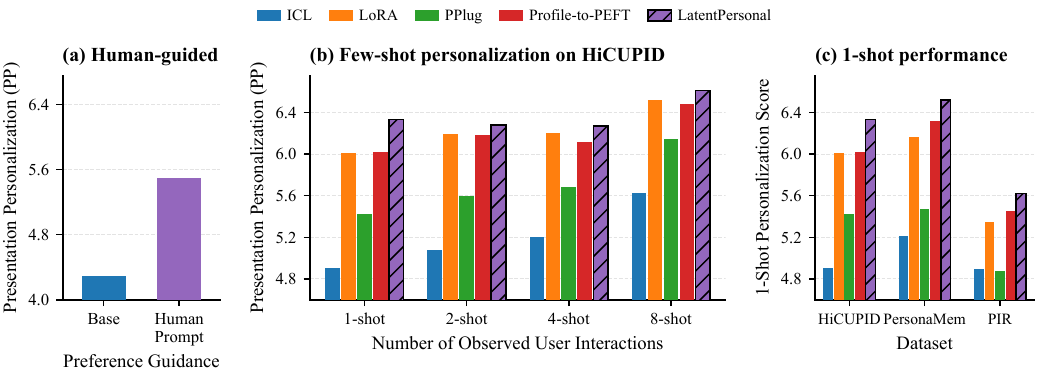}
    \caption{
\textbf{Personalized behaviors can be elicited from pretrained LLMs with limited user-specific evidence.}
\textbf{(a)} Providing a frozen pretrained LLM with a human-identified preference description derived from observed interactions improves personalization without user-specific parameter learning, suggesting that pretrained LLMs can exhibit personalization-relevant behaviors under appropriate guidance.
\textbf{(b)} On HiCUPID, \textit{LatentPersonal} achieves strong personalization with as few as one observed interaction and remains competitive as additional interactions become available.
\textbf{(c)} Its strong one-shot performance generalizes across three representative personalization datasets.
}    
\label{fig:intro_results}
\end{figure*}

We hypothesize that pretrained LLMs may already encode personalization-relevant behaviors that can be activated with limited user-specific evidence. This intuition is supported by Figure~\ref{fig:intro_results}(a), where a human-constructed preference prompt substantially improves personalization in a frozen pretrained LLM without user-specific parameter updates. If the personalization-relevant variation underlying these behaviors can be effectively represented, the resulting representation can serve as guidance for eliciting corresponding model behavior. From this perspective, personalization can be viewed as navigating a general-purpose model toward user-aligned behavior, rather than learning a separate adaptation from scratch for each user.

Parameter-efficient fine-tuning (PEFT) has become a widely adopted paradigm for adapting pretrained LLMs~\citep{houlsby2019parameter,hu2021lora,li2021prefix}. Importantly, its parameter-efficient adaptation form provides a concrete basis for the navigation perspective above. Rather than modifying the pretrained model as a whole, PEFT introduces a parameter-efficient adaptation that can be written as
\vspace{0.0cm}
\begin{equation}
	\theta=\theta_0+\Delta\theta,
\end{equation}
where $\theta_0$ denotes the pretrained model and $\Delta\theta$ represents the adaptation applied to it. Building on this view, if personalization-relevant characteristics can be effectively represented within the space of model adaptations $\Delta\theta$, then only a few user-specific samples may be sufficient to locate an appropriate adaptation for a target user.

To operationalize this navigation perspective, we introduce \textit{LatentPersonal}, a personalized adaptation framework that represents personalization-relevant variations in a shared low-dimensional latent space and uses user-specific interactions to navigate this space for personalized adaptation.

Our contributions can be summarized as follows:

\begin{itemize}[leftmargin=*]
	\item[{\bf i)}] We introduce a new perspective on LLM personalization by formulating user adaptation as navigation within a shared latent preference space toward user-specific adaptations. Building on this perspective, we propose \textit{LatentPersonal}, a latent-conditioned framework for personalized PEFT that learns compact user representations and uses them to navigate shared adaptation structures toward user-specific parameter updates.
	\item[{\bf ii)}] We instantiate \textit{LatentPersonal} across PEFT architectures, with a LoRA-based realization that directly integrates the user latent representation into the intermediate low-rank space, allowing the latent representation to modulate shared low-rank adaptation directions and thereby produce user-specific parameter updates within a shared LoRA structure. We further generalize this design to other PEFT architectures, including Adapter and Prompt Tuning.
	\item[{\bf iii)}] We empirically demonstrate the effectiveness of \textit{LatentPersonal} across multiple personalization datasets, PEFT architectures, and varying amounts of user-specific evidence. Our results show that \textit{LatentPersonal} is particularly effective in low-shot regimes, achieving strong personalization from only a few observed interactions. Moreover, our analyses reveal consistent and user-specific adaptation patterns in the learned latent space, providing empirical support for the latent navigation perspective.
\end{itemize}

As illustrated in Figure~\ref{fig:intro_results}(b,c), \textit{LatentPersonal} achieves effective personalization from only a few observed user interactions, supporting our view of personalization as navigating toward user-aligned behavior. Project Page: \url{https://pasalab.github.io/LatentPersonal/}.

%% file: chapters/related.tex
\section{Related Work}
\label{related work}

\paragraph{Parameter-Efficient Fine-Tuning of LLMs.}
Parameter-efficient fine-tuning (PEFT)~\citep{lialin2023scaling} has emerged as an effective alternative to full fine-tuning by adapting pretrained models with only a small number of trainable parameters. Existing approaches realize parameter-efficient adaptation through different forms of parameterization. Adapters~\citep{houlsby2019parameter} introduce lightweight trainable modules into pretrained models, while Prefix-Tuning~\citep{li2021prefix} and Prompt Tuning~\citep{lester2021power} optimize continuous representations with the pretrained parameters kept frozen. Intrinsic-dimension methods instead explore efficient adaptation by restricting optimization to low-dimensional subspaces of the full parameter space~\citep{aroca2025aligning}. LoRA~\citep{hu2021lora} takes a different approach by parameterizing weight updates with low-rank decomposition and has become one of the most widely adopted PEFT methods for LLMs.

\paragraph{Personalized Adaptation of LLMs.}
Personalization has become increasingly important as LLMs are deployed to serve users with diverse preferences and needs. Early works such as LaMP~\citep{salemi2024lamp} incorporated user-specific information into the model context through personalized prompting and retrieval from user histories. Subsequent approaches moved beyond explicit retrieval by learning compact user representations from historical interactions to condition model generation~\citep{ning2025user,liu2025llms+,cao2026beyond}. PerFit~\citep{liu2026perfit} further explores low-rank personalization shifts in the hidden representation space and performs user-specific adaptation through representation-level interventions. In parallel, parameter-efficient fine-tuning has enabled personalization directly through model parameters. OPPU~\citep{tan2024democratizing} learns a separate PEFT module for each user, while subsequent approaches such as Personalized Pieces~\citep{tan2024personalized} and PROPER~\citep{zhang2025proper} improve the scalability of personalized adaptation through parameter composition and progressive group-to-individual adaptation, respectively. More recently, Profile-to-PEFT~\citep{tan2026instant} generates personalized PEFT parameters directly from user profiles through a hypernetwork, avoiding per-user optimization at deployment time.

\paragraph{Latent Representations for Model Adaptation.}
Latent-variable modeling provides a general mechanism for exposing low-dimensional structure underlying high-dimensional observations, and has become a fundamental modeling principle across representation learning~\citep{bengio2013representation} and generative modeling~\citep{kingma2013auto}. Latent representations have also been widely explored in language models, including for language modeling~\citep{fang2019implicit}, text generation~\citep{hu2022fuse}, and controllable generation~\citep{gu2023controllable}. Similar ideas have recently appeared in personalized adaptation. Profile-to-PEFT~\citep{tan2026instant} encodes user profiles into compact representations and uses them to condition a hypernetwork that generates selected user-specific LoRA parameters. However, this formulation treats user adaptation primarily as a parameter prediction problem and does not explicitly characterize the shared low-dimensional structure underlying personalization across users.

%% file: chapters/methdology.tex
\section{Methodology}
\label{Methodology}

In this section, we introduce \textit{LatentPersonal}, a framework for efficient personalization from limited user interactions. Our core intuition is that pretrained LLMs may already encode behavioral variations relevant to individual preferences, while these variations are not necessarily expressed for a particular user. We therefore model personalization as navigating user-relevant variations in a shared latent adaptation space, rather than independently optimizing a separate adaptation for each user.

We organize the methodology as follows. Section~\ref{sec:latent_navigation} formulates personalized adaptation as navigation in a shared low-dimensional latent preference space. Section~\ref{sec:latent_adap} introduces an information-bottlenecked latent adaptation mechanism that infers compact user-specific representations from observed interactions and maps them to personalized model updates. Finally, Section~\ref{sec:lora_instantiation} presents LoRA as a representative parameter-efficient instantiation of \textit{LatentPersonal}.

\subsection{Personalization as Latent-Space Navigation}
\label{sec:latent_navigation}

\begin{wrapfigure}{r}{0.43\textwidth}
    \vspace{-10pt}
    \centering
    \includegraphics[width=0.4\textwidth]{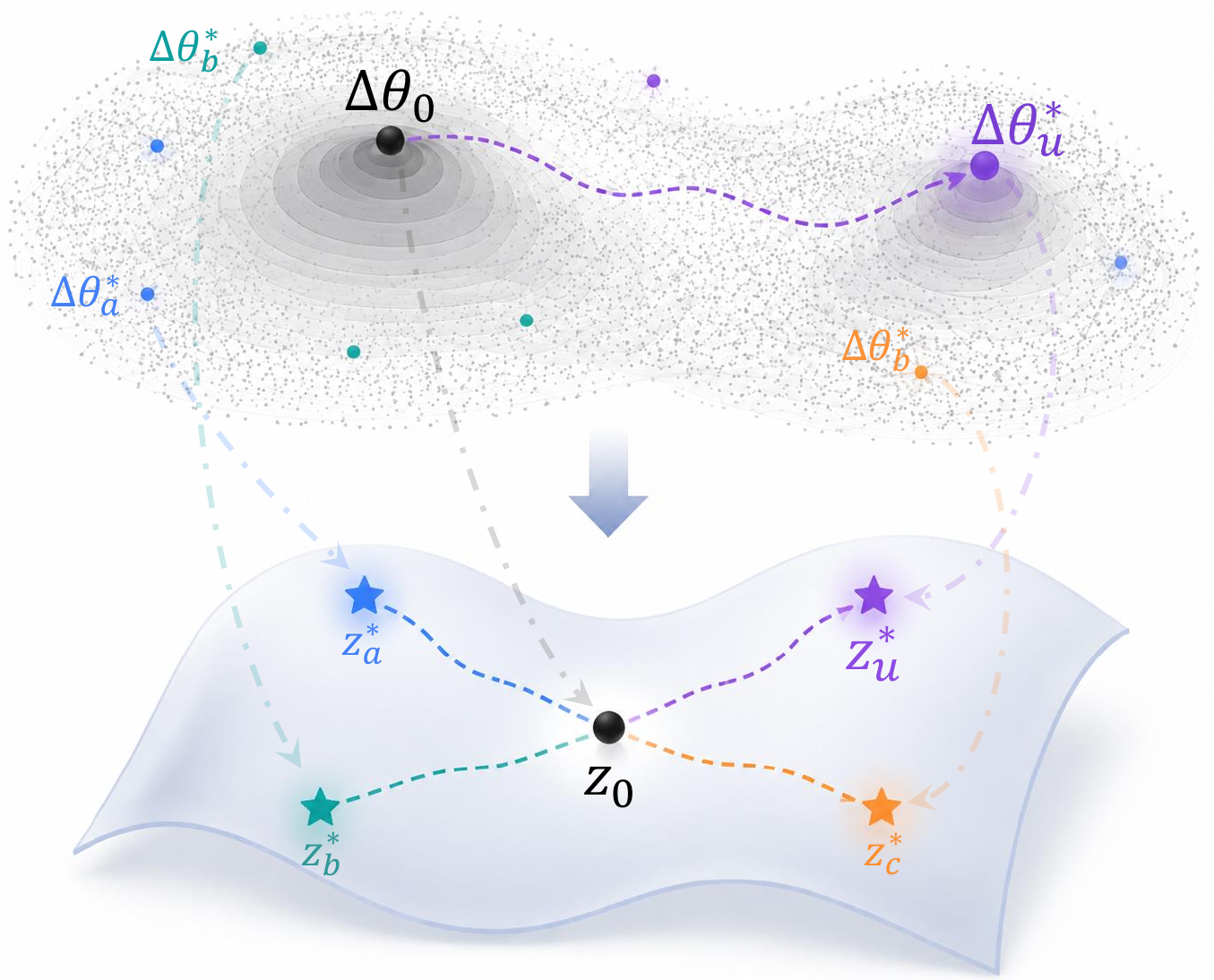}
    \vspace{-11pt}
    \caption{Illustration of personalized adaptation as navigation in the latent preference space of \textit{LatentPersonal}.}
    \label{fig:latentpersonal}
    \vspace{-1pt}
\end{wrapfigure}

We formulate personalized adaptation in a low-dimensional latent preference space.
Given a pretrained LLM with parameters $\theta_0$, each user $u$ provides a set of interactions $D_u=\{(x_i,y_i)\}_{i=1}^{K}$, where $x_i$ denotes the input, $y_i$ denotes the corresponding user-specific target response, and  $K$ is the number of available interaction samples. The objective is to obtain a personalized model that adapts the pretrained model to the user's preferences.

Instead of independently learning user-specific adaptations, we model personalization-relevant variations through a shared low-dimensional latent preference space. Specifically, we introduce a latent variable
\begin{equation}
z_u\in\mathbb{R}^{d},
\end{equation}
where $d$ is substantially smaller than the dimensionality of the parameter adaptation, and $z_u$ encodes the preference characteristics inferred from the user's observed interactions. This formulation is motivated by the intuition that personalization primarily involves capturing user-specific variations while largely preserving the pretrained model's general capabilities.

Under this formulation, personalized adaptation is expressed through a latent-to-parameter mapping:
\begin{equation}
\theta_u=\theta_0+g(z_u),
\end{equation}
where $g(\cdot)$ maps the latent preference representation to the corresponding model adaptation. Personalization can therefore be formulated as locating an optimal user-specific representation in the latent preference space:
\begin{equation}
z_u^*
=
\arg\min_{z}
\mathcal{L}(D_u,\theta_0+g(z)).
\end{equation}

This formulation provides a natural interpretation of personalization as navigation: the observed user interactions guide the model toward an appropriate location $z_u^*$ in the shared latent preference space, which in turn determines the corresponding personalized adaptation through $g(\cdot)$.

\subsection{Information-Bottlenecked Latent Adaptation}
\label{sec:latent_adap}

Given the latent-space formulation above, the remaining challenge is to infer a compact preference representation $z_u$ from the observed interactions $D_u$ and map it into an effective personalized adaptation. \textit{LatentPersonal} models this process through a latent-conditioned mapping:
\begin{equation}
	\Delta\theta_u=g_\phi(z_u),
\end{equation}
where $g_\phi(\cdot)$ denotes a shared latent-to-adaptation mapping learned across users. Rather than optimizing independent adaptation parameters for each user, this shared mapping produces user-specific adaptations conditioned on their latent preference representations.

\paragraph{Adaptation-sufficient preference representation.}
During training, for each user $u$, we partition the available interactions into a support set
$D_u^{\mathrm{sup}}$ and a disjoint query set $D_u^{\mathrm{qry}}$. Although the support interactions
$D_u^{\mathrm{sup}}$ provide evidence about user preferences, they also contain interaction-specific
variations, such as task content and local contextual details, that are not necessarily relevant to
personalization. Meanwhile, the pretrained LLM encodes broad linguistic and reasoning capabilities
in a high-dimensional parameter space. In contrast, user-specific preferences correspond to only a
limited subset of the model behaviors that need to be adapted.

To obtain a compact preference representation, we introduce an information bottleneck (IB)~\citep{tishby2000information} over the latent variable $z_u$. The bottleneck constrains the information encoded in $z_u$ to focus on the factors relevant to personalized adaptation. Let $\Delta\theta_u^*$ denote the optimal user-specific adaptation for user $u$. Ideally, $z_u$ should retain all information in $D_u^{\mathrm{sup}}$ that is relevant to $\Delta\theta_u^*$. An ideal representation would satisfy the following adaptation-sufficiency criterion:
\begin{equation}
	I(D_u^{\mathrm{sup}};\Delta\theta_u^* \mid z_u)=0 .
\label{eq:adaptation_sufficiency}
\end{equation}
Among such adaptation-sufficient representations, we further prefer compact representations that discard interaction-specific information unnecessary for personalization.

Motivated by this ideal representation, we introduce a learnable preference encoder $q_\psi(z_u \mid D_u^{\mathrm{sup}})$ to characterize the latent representation inferred from user interactions. We seek to maximize the information retained by $z_u$ about the desired user-specific adaptation, while simultaneously restricting its dependence on the observed interactions. This encourages the latent variable to capture adaptation-relevant preference factors rather than indiscriminately encoding the interaction history. This formulation serves as a design principle rather than an exact characterization of the objective optimized in practice, since $\Delta\theta_u^*$ is neither observed nor uniquely defined.
 Motivated by this ideal, we use the following information-bottleneck formulation as a conceptual objective:

\begin{equation}
\max_{\psi}\;\underbrace{I(z_u;\Delta\theta_u^*)}_{\text{adaptation preservation}}
-
\beta
\underbrace{
I(z_u;D_u^{\mathrm{sup}})
}_{\text{interaction compression}},
\label{eq:adaptation_ib}
\end{equation}

where $\beta>0$ controls the strength of the information bottleneck and balances adaptation preservation against interaction compression.

Direct optimization of Eq.~(\ref{eq:adaptation_ib}) is intractable in practice, as the optimal user-specific adaptation $\Delta\theta_u^*$ is not directly available and both mutual information terms are difficult to evaluate. We therefore derive tractable surrogate objectives for the two terms below.

\paragraph{Adaptation preservation objective.}

We first consider the adaptation preservation term $I(z_u;\Delta\theta_u^*)$, which encourages the latent preference representation to preserve information relevant to personalized adaptation. However, the optimal adaptation $\Delta\theta_u^*$ is not directly available during training, making direct supervision or explicit estimation of this mutual information infeasible. We therefore exploit the functional role of the latent representation established above: if $z_u$ preserves the information required for personalized adaptation, the adaptation induced by $z_u$ should enable the personalized model exhibit user-specific behaviors on held-out interactions from the same user.

To operationalize this criterion, for each training user $u$, we use the support interaction set $D_u^{\mathrm{sup}}$ to infer a compact latent preference representation through the preference encoder:
\begin{equation}
z_u \sim q_\psi(z_u \mid D_u^{\mathrm{sup}}),
\label{eq:preference_posterior}
\end{equation}
The resulting representation is then translated into a user-specific adaptation through the shared latent-to-adaptation mapping:
\begin{equation}
\Delta\theta_u = g_\phi(z_u),
\label{eq:latent_adaptation}
\end{equation}
where $g_\phi$ is the shared latent-to-adaptation mapping introduced above.

Rather than matching the unavailable $\Delta\theta_u^*$ directly, we optimize the induced personalized model on the query interactions:
\begin{equation}
\mathcal{L}_{\mathrm{adapt}} = 
\mathbb{E}_{z_u \sim q_\psi(z_u \mid D_u^{\mathrm{sup}})}
\left[
\mathcal{L}
\left(
D_u^{\mathrm{qry}};
\theta_0 + g_\phi(z_u)
\right)
\right].
\label{eq:adaptation_preservation}
\end{equation}
Here, $\mathcal{L}(D_u^{\mathrm{qry}};\theta_u)$ denotes the task loss under the personalized model, and $\mathcal{L}_{\mathrm{adapt}}$ serves as the adaptation-preservation objective. Minimizing $\mathcal{L}_{\mathrm{adapt}}$ encourages the latent representation inferred from $D_u^{\mathrm{sup}}$ to preserve information that is useful for inducing personalized behavior reflected in the observed interactions. We use this task loss as a tractable surrogate for the ideal adaptation-preservation term in Eq.~\ref{eq:adaptation_ib}, rather than as a formal bound on $I(z_u;\Delta\theta_u^*)$.

\paragraph{Interaction compression objective.}

While the adaptation preservation objective encourages $z_u$ to retain information necessary for generating effective personalized adaptations, the latent representation may still encode redundant interaction-specific variations that do not contribute to personalized adaptation. We therefore introduce a compression objective by minimizing the mutual information between the latent representation and the observed interactions, i.e., $I(z_u;D_u^{\mathrm{sup}})$.

Direct optimization of $I(z_u;D_u^{\mathrm{sup}})$ is intractable because it depends on the marginal distribution of the latent variable. We therefore introduce a variational prior $r(z)$ to upper-bound this mutual information and instantiate it as a standard Gaussian $r(z)=\mathcal N(0,I)$, yielding a tractable KL regularization term. The full derivation is provided in Appendix~\ref{app:interaction_compression}:
\begin{equation}
\mathcal{L}_{\mathrm{comp}}
=
\mathbb{E}_{u}
\left[
D_{\mathrm{KL}}
\left(
q_\psi(z_u|D_u^{\mathrm{sup}})
\|
r(z)
\right)
\right].
\end{equation}
Minimizing $\mathcal{L}_{\mathrm{comp}}$ restricts the interaction-specific information encoded in $z_u$, encouraging the latent representation to focus on preference factors relevant to personalized adaptation.

\paragraph{Learning the latent preference space.}

Combining the adaptation preservation and interaction compression objectives, we jointly optimize the preference encoder $q_\psi$ and the shared latent-to-adaptation mapping $g_\phi$ across training users. This joint training learns a shared latent preference space in which compact preference representations inferred from limited user interactions induce the corresponding personalized adaptations. The resulting objective is

\begin{equation}
\begin{aligned}
\mathcal{L}_{\mathrm{total}}
&=
\mathcal{L}_{\mathrm{adapt}}
+
\beta\mathcal{L}_{\mathrm{comp}}
\\
&=
\underbrace{
\mathbb{E}_{z_u\sim q_\psi(z_u\mid D_u^{\mathrm{sup}})}
\left[
\mathcal{L}
\left(
D_u^{\mathrm{qry}};
\theta_0+g_\phi(z_u)
\right)
\right]
}_{\text{adaptation preservation}}
+
\beta
\underbrace{
\mathbb{E}_{u}
\left[
D_{\mathrm{KL}}
\left(
q_\psi(z_u\mid D_u^{\mathrm{sup}})
\|
r(z)
\right)
\right]
}_{\text{interaction compression}}.
\end{aligned}
\label{eq:final_objective}
\end{equation}

We train LatentPersonal in a supervised manner, jointly optimizing the preference encoder $q_\psi$ and the shared adaptation mapping $g_\phi$ across training users using their observed interactions. The adaptation preservation term provides the learning signal for both $q_\psi$ and $g_\phi$, while the interaction compression term regularizes $q_\psi$ by constraining the information encoded in $z_u$. Consequently, $q_\psi$ learns to infer compact preference representations from limited user interactions, while $g_\phi$ learns to generate the corresponding personalized adaptations. At deployment, \textit{LatentPersonal} adapts to an unseen user by inferring $z_u$ from a small set of observed interactions and generating the corresponding adaptation through $g_\phi$, without requiring user-specific optimization.

\vspace{-0.3cm}
\subsection{LoRA Instantiation of \textit{LatentPersonal}}
\label{sec:lora_instantiation}

To instantiate the proposed \textit{LatentPersonal} framework, we adopt Low-Rank Adaptation (LoRA)~\citep{hu2021lora} as a representative PEFT technique~\citep{houlsby2019parameter,zaken2022bitfit}. The detailed implementation and training procedure are provided in Appendix~\ref{app:lora_implementation}. LoRA provides a natural adaptation space for \textit{LatentPersonal} due to its intrinsic low-rank parameterization. Given a pretrained weight matrix $W_0^{(l)}$, LoRA represents the adaptation as

\begin{equation}
	\Delta W^{(l)}=B^{(l)}A^{(l)}.
\end{equation}

where $A^{(l)}$ and $B^{(l)}$ are trainable low-rank factors. Instead of learning independent adapters for different users, \textit{LatentPersonal} uses the latent preference representation $z_u$ to modulate this adaptation space and generate user-conditioned updates:
\begin{equation}
	\Delta W_u^{(l)} = B^{(l)} \operatorname{diag}(z_u) A^{(l)}.
	\label{eq:main_latent_lora}
\end{equation}

Here, $\{A^{(l)},B^{(l)}\}$ parameterize the shared adaptation subspace learned from the user population, while $z_u$ determines the user-specific modulation within this space. In this instantiation, the LoRA structure itself realizes the latent-to-adaptation mapping $g_\phi$, avoiding the need to introduce an additional parameter generator. Compared with standard LoRA, the only structural modification is the insertion of the latent modulation $\operatorname{diag}(z_u)$ between the two low-rank factors. During training, the shared LoRA factors $\{A^{(l)},B^{(l)}\}$ are learned jointly with the lightweight preference encoder $q_\psi$, which maps the support interactions $D_u^{\mathrm{sup}}$ to the latent representation $z_u$. Thus, extending standard LoRA to \textit{LatentPersonal} requires only a lightweight preference encoder and a corresponding modification to the training objective.

At inference time, the shared LoRA factors remain fixed, and a new user's adaptation is obtained solely by inferring $z_u$ from the available interactions through $q_\psi$. No user-specific optimization of $A^{(l)}$ or $B^{(l)}$ is required. Consequently, \textit{LatentPersonal} can construct user-specific parameter updates from only a few interactions by leveraging the shared adaptation structure learned during training.

%% file: chapters/experiment.tex
\section{Experiments}
\label{Experiment}
\subsection{Experimental Setup}
\paragraph{Datasets.} We evaluate \textit{LatentPersonal} on four personalization benchmarks: HiCUPID, PRISM, PersonaMem and Persona-Iterative-Responses (PIR).

\begin{itemize}[leftmargin=*, , nosep]

	\item \textbf{HiCUPID}: HiCUPID~\citep{mok2025exploring} is a conversational benchmark for personalized AI assistants. It evaluates personalized response generation based on user-specific information captured through prior conversations.

    \item \textbf{PRISM}: PRISM~\citep{kirk2024prism} is a human-feedback dataset that captures individual preferences through real-world interactions with diverse LLMs. Each participant engages in multiple conversations and provides personalized ratings and fine-grained feedback on model responses.

    \item \textbf{PersonaMem}: PersonaMem~\citep{jiang2025personamem} is a benchmark for personalized response generation from long-term user interactions. We adopt PersonaMem-v2, which emphasizes implicit user preferences embedded in interaction histories rather than explicitly provided user profiles.

    \item \textbf{PIR}: Persona-Iterative-Responses (PIR) is a large-scale persona-conditioned preference dataset containing diverse questions paired with persona descriptions and alternative responses. The dataset captures how different persona characteristics induce distinct response preferences, providing a complementary setting for evaluating whether personalized models can infer and reproduce user-specific response tendencies from limited observations.
\end{itemize}

\paragraph{Baselines.} We compare \textit{LatentPersonal} with the base model without personalization and four representative personalization baselines. Unless otherwise specified, \textit{LatentPersonal} refers to its LoRA-based instantiation used throughout the main experiments. 

\begin{itemize}[leftmargin=*, , nosep]

    \item \textbf{ICL}: In-context learning (ICL) \citep{brown2020language} performs few-shot personalization by conditioning the base model on user-specific demonstrations without parameter updates.
    
    \item \textbf{LoRA}: We directly fine-tune LoRA parameters on user-specific samples, providing a standard
parameter-efficient baseline for personalized adaptation.

    \item \textbf{PPlug}: Persona-Plug (PPlug)~\citep{liu2025llms+} learns a user-specific representation from user samples and conditions the LLM on the resulting personal embedding for personalized generation.

    \item \textbf{Profile-to-PEFT}: Profile-to-PEFT \citep{tan2026instant} encodes user-specific information into a user representation and employs a shared hypernetwork to generate personalized PEFT parameters for efficient adaptation.
\end{itemize}

All experiments are conducted on a single NVIDIA A100-SXM4-80GB GPU. Additional implementation details, including training hyperparameters and model configurations, are provided in Appendix~\ref{app:model_config}, while detailed definitions of the evaluation metrics are provided in Appendix~\ref{app:evaluation_metrics}. 

\subsection{Results and Analysis}

\paragraph{Personalization Performance.} We report the main results using Llama-3.1-8B-Instruct~\citep{grattafiori2024llama} as the backbone and select three representative metrics for a concise comparison. Results with additional backbone models and the complete set of evaluation metrics are provided in Appendix~\ref{app:additional_performance_results}. We first report the zero-shot performance of the base model in Table~\ref{tab:base_results}, followed by the main personalization results under different numbers of user-specific observations in Table~\ref{tab:main_results}. 

Overall, \textit{LatentPersonal} achieves strong performance across all four benchmarks, with particularly consistent advantages in the few-shot regime. As more user observations become available, the gap to competing methods generally narrows, further highlighting the effectiveness of latent navigation when user-specific evidence is limited.
\input{tab/main_result_base.tex}

\input{tab/main_result.tex}

\paragraph{Component Analysis.}
We ablate three key designs of \textit{LatentPersonal}. with the main results reported in Table~\ref{tab:ablation}.
For \textit{w/o Latent}, we remove the latent modulation $z_u$ from \textit{LatentPersonal} while retaining the same optimization framework, such that personalization is performed directly through the LoRA components.
For \textit{w/o Navigation}, we retain the inferred user latent representation $z_u$ but keep it fixed during personalization, and instead update the LoRA components, thereby replacing latent-space navigation with direct parameter adaptation.
For \textit{w/o IB}, we remove the information bottleneck regularization by setting $\beta=0$.

\input{tab/ablation.tex}
Removing either the latent representation or the information bottleneck leads to substantial performance degradation, while replacing latent-space navigation with direct LoRA adaptation results in a smaller but consistent drop, supporting the contribution of all three components. Detailed ablation results are provided in Appendix~\ref{app:additional_ablation}, with additional sensitivity analyses on the bottleneck coefficient $\beta$ and latent dimension $d_z$ in Appendix~\ref{app:sensitivity_analysis}.

\paragraph{Test-Time Personalization Efficiency.}
We further evaluate the efficiency of test-time adaptation across personalization methods, with the main results reported in Table~\ref{tab:test_time_efficiency}. Personalization latency measures the time required to process the observed user interactions and obtain the user-specific adaptation before response generation. \textit{LatentPersonal} substantially reduces personalization overhead with only 1-4 observed interactions, and its cost increases moderately as more observations are incorporated. Detailed results on both personalization and generation latency are provided in Appendix~\ref{app:test_time_efficiency}.

\input{tab/time.tex}

\begin{figure*}[!th]
    \centering

    \begin{subfigure}[t]{0.48\textwidth}
        \centering
        \includegraphics[
            width=0.9\linewidth
        ]{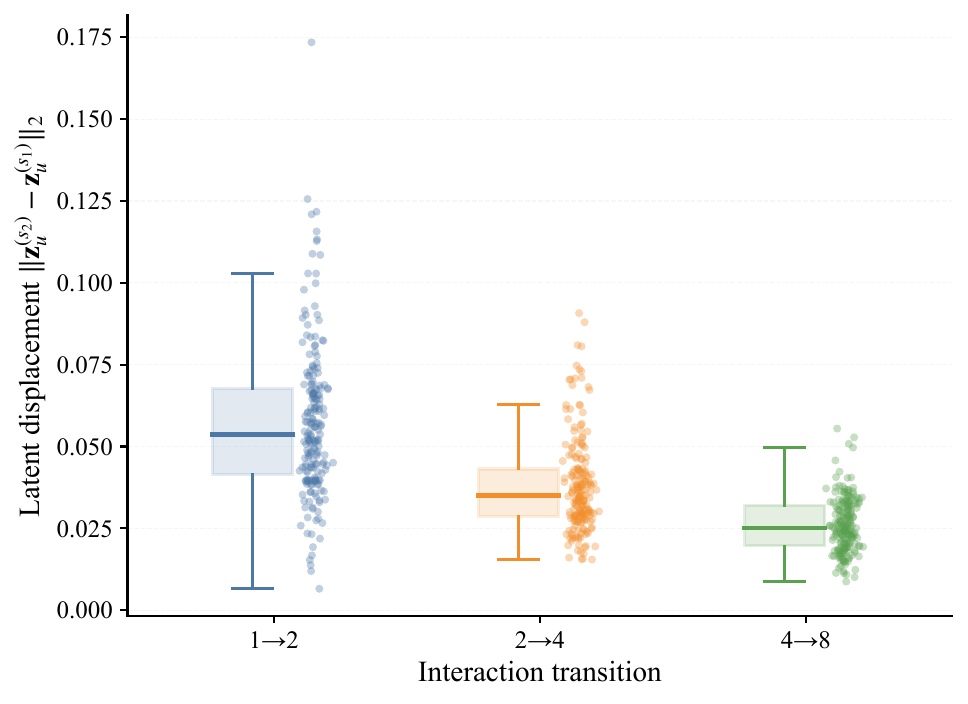}
        \caption{Shot-wise latent displacement.}
    \end{subfigure}
    \hfill
    \begin{subfigure}[t]{0.48\textwidth}
        \centering
        \includegraphics[
            width=0.9\linewidth
        ]{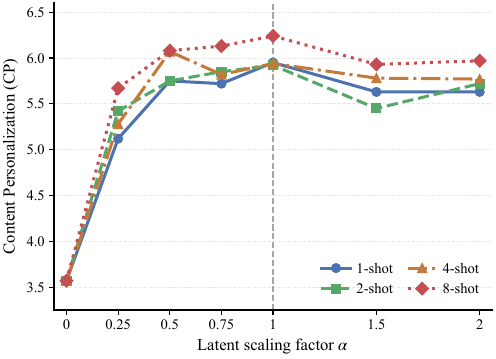}
        \caption{Effect of latent scaling.}
        \label{fig:latent_scaling}
    \end{subfigure}

    \caption{
        \textbf{Structure and functional behavior of the learned latent space on HiCUPID.}
        (a) Shot-wise latent displacement decreases with additional user observations.
        (b) Personalization performance under different latent scaling factors $\alpha$.
}
    \label{fig:latent_structure}

\end{figure*}

\begin{figure*}[!th]
\vspace{-0.5cm}
    \centering

    \begin{subfigure}[t]{0.65\textwidth}
        \centering
        \includegraphics[
            width=0.95\linewidth
        ]{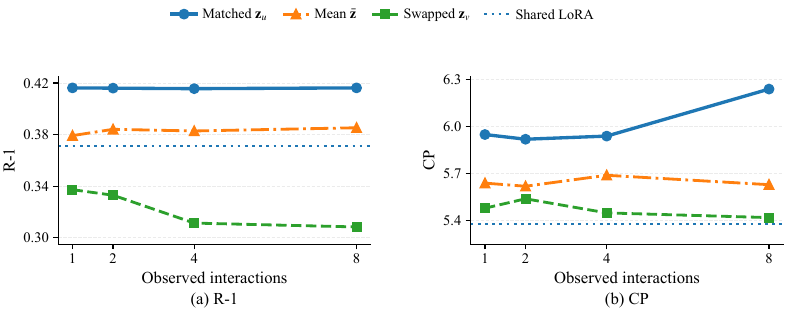}
        \caption{User-specificity analysis.}
        \label{fig:user_specificity}
    \end{subfigure}
    \hfill
    \begin{subfigure}[t]{0.32\textwidth}
        \centering
        \includegraphics[
            width=0.95\linewidth
        ]{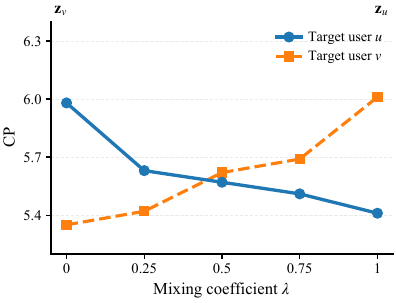}
        \caption{Inter-user interpolation.}
        \label{fig:latent_interpolation}
    \end{subfigure}

    \caption{
\textbf{User specificity and continuous navigation in the learned latent adaptation space on HiCUPID.}
(a) Matched user representations outperform swapped, population-mean, and shared LoRA alternatives.
(b) Interpolation between two representative users smoothly shifts personalization performance from one target user toward the other.
}
    \label{fig:latent_user_analysis}

\end{figure*}

\paragraph{Structure of the Learned Latent Space.}
We further examine the learned latent space in Figure~\ref{fig:latent_scaling}. We scale the inferred user representation as $\mathbf{z}'_u=\alpha\mathbf{z}_u$, where $\alpha$ controls the strength of latent guidance. Latent updates progressively decrease as more user observations are incorporated, suggesting that user representations gradually stabilize with increasing evidence. Meanwhile, removing the latent signal ($\alpha=0$) substantially degrades personalization, while moderate scaling maintains strong performance. We further analyze user-specific navigation in Figure~\ref{fig:latent_user_analysis}. Matched user representations consistently outperform swapped, population-mean, and shared LoRA alternatives, while interpolating between two user representations smoothly shifts personalization performance from one target user toward the other. Together, these results suggest that the learned latent space is not only structured and stable, but also supports continuous user-specific navigation. More extensive analyses are provided in Appendix~\ref{app:latent_analysis}.

\paragraph{Generalization across PEFT Architectures.}
Finally, we evaluate Adapter-based and Prompt Tuning-based instantiations of \textit{LatentPersonal}, with results reported in Appendix~\ref{app:peft_evaluation}. The consistent improvements across different PEFT architectures demonstrate that \textit{LatentPersonal} is not specific to its LoRA-based instantiation, but can be naturally extended to different parameter-efficient adaptation mechanisms. Implementation details of these variants are provided in Appendix~\ref{app:peft_architectures}.

%% file: tab/main_result_base.tex
\begin{table*}[!b]

\centering
\caption{
Performance of the frozen base model without personalized adaptation.
}
\label{tab:base_results}

\small
\setlength{\tabcolsep}{12pt}
\renewcommand{\arraystretch}{1.08}

\begin{tabular}{lccc}
\toprule
\textbf{Dataset}
& \textbf{ROUGE-1}
& \textbf{METEOR}
& \textbf{Content Personalization} \\
\midrule

HiCUPID
& 0.1614
& 0.2166
& 3.57 \\

PRISM
& 0.2558
& 0.2125
& 4.18 \\

PersonaMem
& 0.2368
& 0.2103
& 5.01 \\

PIR
& 0.2954
& 0.1966
& 4.08 \\

\bottomrule
\end{tabular}
\end{table*}

%% file: tab/main_result.tex
\begin{table*}[!th]
\centering
\caption{
Main results across four personalization benchmarks under different numbers of observed user interactions. R-1, MET., and CP denote ROUGE-1, METEOR, and Content Personalization, respectively. Best results are shown in \textbf{bold} and second-best results are \underline{underlined}.
}
\label{tab:main_results}

\scriptsize
\setlength{\tabcolsep}{3.0pt}
\renewcommand{\arraystretch}{1.08}

\begin{tabular}{cl*{12}{c}}
\toprule

\textbf{Dataset}
& \textbf{Method}
& \multicolumn{3}{c}{\textbf{1-shot}}
& \multicolumn{3}{c}{\textbf{2-shot}}
& \multicolumn{3}{c}{\textbf{4-shot}}
& \multicolumn{3}{c}{\textbf{8-shot}} \\

\cmidrule(lr){3-5}
\cmidrule(lr){6-8}
\cmidrule(lr){9-11}
\cmidrule(lr){12-14}

&
& R-1 & MET. & CP
& R-1 & MET. & CP
& R-1 & MET. & CP
& R-1 & MET. & CP \\

\midrule

\multirow{5}{*}{\textbf{HiCUPID}}
& ICL
& 0.1713 & 0.2377 & 3.85
& 0.1682 & 0.2294 & 4.25
& 0.1782 & 0.2429 & 4.64
& 0.1934 & 0.2730 & 5.52 \\

& LoRA
& 0.3886 & 0.3829 & 5.57
& 0.3993 & 0.3940 & 5.73
& 0.3912 & 0.3923 & 5.61
& \underline{0.3984} & 0.4027 & \textbf{6.26} \\

& PPlug
& 0.3018 & 0.2696 & 5.15
& 0.3116 & 0.2813 & 5.37
& 0.3276 & 0.3018 & 5.63
& 0.3315 & 0.3105 & 5.95 \\

& Profile-to-PEFT
& \underline{0.4015} & \underline{0.4022} & \underline{5.64}
& \underline{0.4018} & \underline{0.4052} & \underline{5.78}
& \underline{0.4039} & \underline{0.4014} & \underline{5.75}
& 0.3975 & \underline{0.4078} & 6.17 \\

& \textbf{LatentPersonal}
& \textbf{0.4162} & \textbf{0.4133} & \textbf{5.95}
& \textbf{0.4160} & \textbf{0.4142} & \textbf{5.92}
& \textbf{0.4156} & \textbf{0.4137} & \textbf{5.94}
& \textbf{0.4162} & \textbf{0.4135} & \underline{6.24} \\

\midrule

\multirow{5}{*}{\textbf{PRISM}}
& ICL
& \underline{0.2929} & \underline{0.2372} & 4.11
& 0.2917 & 0.2233 & \underline{4.71}
& 0.2998 & \underline{0.2263} & 4.82
& \underline{0.3009} & \underline{0.2276} & \textbf{5.84} \\

& LoRA
& 0.2843 & 0.2046 & 4.05
& 0.2649 & 0.1940 & 4.63
& 0.2568 & 0.1873 & 4.97
& 0.2638 & 0.1926 & 5.23 \\

& PPlug
& 0.2875 & 0.2033 & 4.11
& 0.2806 & 0.1951 & 4.55
& 0.2853 & 0.1950 & 4.97
& 0.2889 & 0.1945 & 5.54 \\

& Profile-to-PEFT
& 0.2895 & 0.2200 & \underline{4.15}
& 0.2888 & 0.2141 & 4.42
& 0.2949 & 0.2221 & \underline{5.28}
& 0.2928 & 0.2187 & 5.68 \\

& \textbf{LatentPersonal}
& \textbf{0.3081} & \textbf{0.2421} & \textbf{4.28}
& \textbf{0.3036} & \textbf{0.2380} & \textbf{4.82}
& \textbf{0.3043} & \textbf{0.2385} & \textbf{5.30}
& \textbf{0.3049} & \textbf{0.2391} & \underline{5.79} \\
\midrule

\multirow{5}{*}{\textbf{PersonaMem}}
& ICL
& 0.2514 & 0.2352 & 4.87
& 0.2682 & 0.2457 & 5.51
& 0.2618 & 0.2457 & 6.07
& 0.2759 & 0.2618 & 6.39 \\

& LoRA
& \underline{0.3448} & \underline{0.2767} & \underline{5.97}
& 0.3442 & 0.2740 & \underline{6.09}
& 0.3447 & 0.2800 & \underline{6.25}
& 0.3446 & 0.2782 & \underline{6.40} \\

& PPlug
& 0.2949 & 0.2265 & 5.17
& 0.2902 & 0.2216 & 5.19
& 0.2939 & 0.2238 & 5.53
& 0.2953 & 0.2226 & 5.88 \\

& Profile-to-PEFT
& 0.3398 & 0.2749 & 5.74
& \underline{0.3480} & \underline{0.2806} & 5.70
& \underline{0.3523} & \underline{0.2827} & 5.51
& \underline{0.3561} & \underline{0.2859} & 6.37 \\

& \textbf{LatentPersonal}
& \textbf{0.3534} & \textbf{0.2803} & \textbf{6.14}
& \textbf{0.3631} & \textbf{0.2930} & \textbf{6.27}
& \textbf{0.3647} & \textbf{0.2996} & \textbf{6.35}
& \textbf{0.3718} & \textbf{0.3031} & \textbf{6.42} \\

\midrule

\multirow{5}{*}{\textbf{PIR}}
& ICL
& 0.3028 & 0.2066 & 4.19
& 0.3090 & 0.2114 & 5.16
& 0.3125 & 0.2129 & \underline{5.89}
& 0.3082 & 0.2082 & 6.51 \\

& LoRA
& 0.3805 & 0.2962 & \underline{4.28}
& 0.3795 & 0.2931 & 5.25
& 0.3836 & 0.2983 & 5.72
& 0.3846 & 0.2977 & \textbf{6.67} \\

& PPlug
& 0.3711 & 0.2495 & 3.96
& 0.3633 & 0.2401 & 5.16
& 0.3645 & 0.2436 & 5.86
& 0.3646 & 0.2409 & 6.56 \\

& Profile-to-PEFT
& \underline{0.3891} & \underline{0.3066} & 4.08
& \textbf{0.3952} & \textbf{0.3111} & 5.13
& \underline{0.3857} & \underline{0.3024} & 5.75
& \textbf{0.3880} & \underline{0.3045} & \underline{6.63} \\

& \textbf{LatentPersonal}
& \textbf{0.3932} & \textbf{0.3087} & \textbf{4.38}
& \underline{0.3926} & \underline{0.3094} & \textbf{5.30}
& \textbf{0.3912} & \textbf{0.3084} & \textbf{5.90}
& \underline{0.3852} & \textbf{0.3163} & 6.58 \\

\bottomrule
\end{tabular}

\end{table*}

%% file: tab/ablation.tex
\begin{table*}[!th]
\centering
\caption{
Ablation study of key components in \textit{LatentPersonal} on HiCUPID with Llama-3.1-8B-Instruct. R-1, MET., and CP denote ROUGE-1, METEOR, and Content Personalization.
}
\label{tab:ablation}

\small
\setlength{\tabcolsep}{4pt}
\renewcommand{\arraystretch}{1.08}

\begin{tabular}{l*{9}{c}}
\toprule

\textbf{Variant}
& \multicolumn{3}{c}{\textbf{1-shot}}
& \multicolumn{3}{c}{\textbf{2-shot}}
& \multicolumn{3}{c}{\textbf{4-shot}} \\

\cmidrule(lr){2-4}
\cmidrule(lr){5-7}
\cmidrule(lr){8-10}

& R-1 & MET. & CP
& R-1 & MET. & CP
& R-1 & MET. & CP \\

\midrule

\textit{LatentPersonal}
& 0.4162 & 0.4133 & 5.95
& 0.4160 & 0.4142 & 5.92
& 0.4156 & 0.4137 & 5.94 \\

w/o Latent
& 0.3550 & 0.3667 & 4.72
& 0.3499 & 0.3841 & 4.86
& 0.3308 & 0.3423 & 5.29 \\

w/o Navigation
& 0.3812 & 0.3777 & 5.66
& 0.3910 & 0.3929 & 5.71
& 0.3909 & 0.3992 & 5.92 \\

w/o IB
& 0.2684 & 0.3423 & 4.57
& 0.2703 & 0.3422 & 4.65
& 0.2695 & 0.3444 & 4.74 \\

\bottomrule
\end{tabular}
\end{table*}

%% file: tab/time.tex
\begin{table}[!th]
\centering

\caption{Test-time personalization latency (s) on HiCUPID.}
\label{tab:test_time_efficiency}

\small
\setlength{\tabcolsep}{5pt}
\renewcommand{\arraystretch}{1.05}

\begin{tabular}{lcccc}
\toprule
\textbf{Method}
& \textbf{1-shot}
& \textbf{2-shot}
& \textbf{4-shot}
& \textbf{8-shot} \\
\midrule

LoRA
& 0.3718 & 0.3622 & 0.3692 & 0.3640 \\

Profile-to-PEFT
& 0.0539 & 0.0538 & 0.0539 & \textbf{0.0544} \\

\textit{LatentPersonal}
& \textbf{0.0219}
& \textbf{0.0304}
& \textbf{0.0404}
& 0.0676 \\

\bottomrule
\end{tabular}
\end{table}

%% file: chapters/conclusion.tex
\section{Conclusion}
\label{Conclusion}

In this work, we revisit LLM personalization through the lens of latent navigation, motivated by the hypothesis that pretrained LLMs may support a broad range of user-specific behaviors that can be elicited with limited user-specific evidence. From this perspective, we introduce \textit{LatentPersonal}, which infers compact user representations and performs personalization through a shared latent adaptation space. Our results show that user-specific adaptations can be effectively represented within this shared space, where representations inferred from only a few interactions enable effective personalization for unseen users. Experiments across multiple personalization benchmarks further demonstrate that \textit{LatentPersonal} is particularly effective in the few-shot regime, while analyses of its components, efficiency, and learned latent space provide empirical support for the proposed formulation. Moreover, its extension across different PEFT architectures suggests that the latent-navigation formulation is not tied to a particular adaptation mechanism. Overall, our results highlight latent navigation as a promising perspective for efficient few-shot personalization of LLMs.

%% file: appendix/app_method.tex
\newpage
\section{Method Supplement}
\subsection{Derivation of Variational Preference Compression}
\label{app:interaction_compression}

We provide the derivation of the variational upper bound used for the interaction compression objective. Under the joint distribution $p(D_u^{\mathrm{sup}})q_\psi(z_u\mid D_u^{\mathrm{sup}})$, the mutual information between the observed interactions $D_u^{\mathrm{sup}}$ and the latent preference representation $z_u$ can be written as

\begin{equation}
I(z_u;D_u^{\mathrm{sup}})=\mathbb{E}_{p(D_u^{\mathrm{sup}})}
	\left[D_{\mathrm{KL}}
		\left(
			q_\psi(z_u|D_u^{\mathrm{sup}})\|q_\psi(z_u)
		\right)
	\right],
\end{equation}

where the aggregated posterior is defined as

\begin{equation}
q_\psi(z_u)
=
\int p(D_u^{\mathrm{sup}})q_\psi(z_u|D_u^{\mathrm{sup}})dD_u^{\mathrm{sup}} .
\end{equation}

Since the aggregated posterior $q_\psi(z_u)$ is generally intractable, we introduce a tractable variational prior $r(z)$. The corresponding KL term can be decomposed as

\begin{align}
&
\mathbb{E}_{p(D_u^{\mathrm{sup}})}
\left[
D_{\mathrm{KL}}
\left(
q_\psi(z_u|D_u^{\mathrm{sup}})
\|
r(z)
\right)
\right]
\nonumber\\
&=
\mathbb{E}_{p(D_u^{\mathrm{sup}})q_\psi(z_u|D_u^{\mathrm{sup}})}
\left[
\log
\frac{q_\psi(z_u|D_u^{\mathrm{sup}})}
{r(z)}
\right]
\nonumber\\
&=
\mathbb{E}_{p(D_u^{\mathrm{sup}})q_\psi(z_u|D_u^{\mathrm{sup}})}
\left[
\log
\frac{q_\psi(z_u|D_u^{\mathrm{sup}})}
{q_\psi(z_u)}
+
\log
\frac{q_\psi(z_u)}
{r(z)}
\right]
\nonumber\\
&=
I(z_u;D_u^{\mathrm{sup}})
+
D_{\mathrm{KL}}
\left(
q_\psi(z_u)
\|
r(z)
\right).
\end{align}

Since $D_{\mathrm{KL}}(q_\psi(z_u)\|r(z))\geq0$, we obtain the variational upper bound

\begin{equation}
I(z_u;D_u^{\mathrm{sup}})
\leq
\mathbb{E}_{p(D_u^{\mathrm{sup}})}
\left[
D_{\mathrm{KL}}
\left(
q_\psi(z_u|D_u^{\mathrm{sup}})
\|
r(z)
\right)
\right].
\end{equation}

For implementation, we adopt a diagonal Gaussian posterior

\begin{equation}
q_\psi(z_u|D_u^{\mathrm{sup}})
=
\mathcal{N}
\left(
\mu_u,
\operatorname{diag}(\sigma_u^2)
\right),
\end{equation}

where $\mu_u\in\mathbb{R}^d$ and $\sigma_u\in\mathbb{R}_{+}^d$ denote the mean and standard deviation vectors predicted by the preference encoder, respectively. We use a standard Gaussian prior $r(z)=\mathcal{N}(0,I)$. The resulting KL regularization admits the closed-form expression:

\begin{equation}
D_{\mathrm{KL}}
\left(
q_\psi(z_u|D_u^{\mathrm{sup}})
\|
\mathcal{N}(0,I)
\right)
=
\frac{1}{2}
\sum_{j=1}^{d}
\left(
\mu_{u,j}^{2}
+
\sigma_{u,j}^{2}
-
\log\sigma_{u,j}^{2}
-
1
\right).
\end{equation}

%% file: appendix/app_lora.tex
\newpage

\section{Detailed LoRA-based Implementation of \textit{LatentPersonal}}
\label{app:lora_implementation}

Following the latent adaptation formulation introduced in Section~\ref{sec:latent_adap}, we provide a detailed implementation of \textit{LatentPersonal} based on LoRA. Although \textit{LatentPersonal} is not restricted to a specific parameter-efficient adaptation mechanism, we choose LoRA as a representative instantiation due to its wide adoption and computational efficiency.

LoRA parameterizes the update of a pretrained weight matrix as a low-rank decomposition:

\begin{equation}
\Delta W^{(l)}
=
B^{(l)}A^{(l)},
\end{equation}

where $B^{(l)}\in\mathbb{R}^{d_{\mathrm{out}}\times r}$ and $A^{(l)}\in\mathbb{R}^{r\times d_{\mathrm{in}}}$, with $r\ll\min(d_{\mathrm{out}},d_{\mathrm{in}})$. This low-rank formulation provides a natural parameterization for realizing the latent-to-adaptation mapping in \textit{LatentPersonal}.

In the following, we describe how user preference representations are incorporated into the LoRA adaptation process and how the resulting latent-conditioned LoRA model is jointly trained across users.

\subsection{Latent-conditioned LoRA Adaptation}
\label{app:latent_lora}

The low-rank structure of LoRA provides a natural adaptation space for realizing the latent-to-adaptation mapping in \textit{LatentPersonal}. Specifically, the rank $r$ determines the dimensionality of the intermediate adaptation space, while the low-rank factors induce a set of shared adaptation directions. For the $l$-th transformer layer, the LoRA factors can be expressed as

\begin{equation}
B^{(l)}A^{(l)}
=
\sum_{i=1}^{r}
b_i^{(l)}a_i^{(l)T},
\end{equation}

where each rank-one component $b_i^{(l)}a_i^{(l)T}$ defines an adaptation direction shared across users. Based on this decomposition, we align the latent preference representation with the LoRA rank space, i.e., $z_u\in\mathbb{R}^{r}$, and interpret $z_u$ as a user-specific coordinate vector that modulates these shared adaptation directions.

Therefore, rather than parameterizing an independent LoRA update for each user, \textit{LatentPersonal} constructs user-conditioned adaptations by modulating the contribution of the shared low-rank directions:

\begin{equation}
\Delta W_u^{(l)}
=
B^{(l)}
\operatorname{diag}(z_u)
A^{(l)}.
\label{eq:latent_lora}
\end{equation}

Equivalently, the personalized update can be written as

\begin{equation}
\Delta W_u^{(l)}
=
\sum_{i=1}^{r}
z_{u,i}
b_i^{(l)}
a_i^{(l)T}.
\end{equation}

Here, the LoRA factors $\{b_i^{(l)}a_i^{(l)T}\}_{i=1}^{r}$ define a shared adaptation basis learned from the population of users, while $z_{u,i}$ specifies the contribution of each basis direction for user $u$. Consequently, \textit{LatentPersonal} does not require independent user-specific adapters; instead, personalization is realized by inferring a compact latent representation that determines the user-specific combination of shared low-rank adaptation directions.

\subsection{Preference Representation Inference}
\label{app:preference_inference}

To infer a user preference representation from a variable-sized set of support interactions, we first transform $D_u^{\mathrm{sup}}=\{(x_i,y_i)\}_{i=1}^{K}$ into a fixed-dimensional representation. Specifically, we use the frozen pretrained LLM backbone as the interaction encoder. For each interaction $(x_i,y_i)$, we extract the final-layer hidden states and apply mean pooling over non-padding tokens to obtain an interaction-level representation $h_i$. The representations of the $K$ observed interactions are then aggregated by mean pooling,

\begin{equation}
	h_u = \frac{1}{K}\sum_{i=1}^{K} h_i,
\end{equation}

yielding a fixed-dimensional user representation independent of the number of observed interactions. The resulting representation $h_u$ is fed into the two-layer MLP preference encoder described in Table~\ref{tab:preference_encoder}, which predicts the parameters of the variational posterior,

\begin{equation}
(\mu_u,\log\sigma_u^2)=q_\psi(h_u),
\qquad
q_\psi(z_u\mid D_u^{\mathrm{sup}})
=
\mathcal{N}\!\left(
\mu_u,\operatorname{diag}(\sigma_u^2)
\right).
\end{equation}

This simple parameter-free aggregation naturally accommodates varying numbers of observed interactions while keeping preference inference lightweight.

\subsection{Training with Latent-Conditioned LoRA}

\label{app:training_lora}

During training, LatentPersonal jointly learns the preference encoder and the shared LoRA adaptation space across multiple users. For each training user $u$, we partition the available interactions into a support set $D_u^{\mathrm{sup}}$ and a disjoint query set $D_u^{\mathrm{qry}}$. The support interactions are used to infer the latent preference representation:

\begin{equation}
z_u\sim q_\psi(z_u|D_u^{\mathrm{sup}}),
\end{equation}

The inferred latent representation is then used to generate the user-conditioned LoRA updates described in Section~\ref{app:latent_lora}. The resulting personalized model is optimized on $D_u^{\mathrm{qry}}$, providing the learning signal for both the preference encoder and the shared LoRA adaptation space.

\begin{equation}
	\theta_u = \theta_0+\Delta\theta_u^{\mathrm{LoRA}}(z_u).
\end{equation}

The resulting personalized model is optimized on the query interactions $D_u^{\mathrm{qry}}$, providing the learning signal for both the preference encoder and the shared LoRA adaptation space. The support interactions $D_u^{\mathrm{sup}}$ are used for latent inference, while the disjoint query interactions $D_u^{\mathrm{qry}}$ provide supervision for learning the shared adaptation space. Specifically, instantiating the general latent-to-adaptation mapping in Eq.~(\ref{eq:latent_adaptation}) with the latent-conditioned LoRA parameterization leads to the following training objective:

\begin{equation}
\mathcal{L}_{\mathrm{total}} =
\mathbb{E}_{z_u\sim q_\psi(z_u\mid D_u^{\mathrm{sup}})}
\left[
\mathcal{L}
\left(
D_u^{\mathrm{qry}};
\theta_0+\Delta\theta_u^{\mathrm{LoRA}}(z_u)
\right)
\right]
+
\beta
\mathbb{E}_{u}
\left[
D_{\mathrm{KL}}
\left(
q_\psi(z_u\mid D_u^{\mathrm{sup}})
\|
r(z)
\right)
\right].
\label{eq:lora_objective}
\end{equation}

which corresponds to the general variational objective in Eq.~(\ref{eq:final_objective}) under the LoRA instantiation.

The preference encoder parameters $\psi$ and the shared LoRA parameters $\{A^{(l)},B^{(l)}\}_{l=1}^{L}$ are jointly optimized by minimizing $\mathcal{L}_{\mathrm{total}}$.

Different from conventional personalized fine-tuning, \textit{LatentPersonal} does not optimize independent LoRA parameters for each user. Instead, it learns a shared LoRA adaptation space across users, while user-specific personalization is achieved by inferring a compact latent preference representation from a small number of observed interactions.

The overall training procedure of \textit{LatentPersonal} with latent-conditioned LoRA is summarized in Algorithm~\ref{alg:personal_lora}.

\input{appendix/app_alg}

%% file: appendix/app_alg.tex
\begin{algorithm}[htbp]
\caption{Training Procedure of \textit{LatentPersonal} with Latent-Conditioned LoRA}
\label{alg:personal_lora}
\begin{algorithmic}[1]

\REQUIRE Pretrained model $\theta_0$, training users $\mathcal{U}_{\mathrm{train}}$

\STATE Initialize preference encoder $q_\psi$

\STATE Initialize shared LoRA parameters
$\phi=\{A^{(l)},B^{(l)}\}_{l=1}^{L}$

\WHILE{not converged}

    \STATE Sample a batch of users
    $\mathcal{B}\subset\mathcal{U}_{\mathrm{train}}$

    \STATE Initialize $\mathcal{L}_{\mathrm{batch}}\leftarrow0$

    \FOR{each user $u\in\mathcal{B}$}

		\STATE Retrieve interactions $D_u$ and split into $D_u^{\mathrm{sup}}$ and $D_u^{\mathrm{qry}}$

        \STATE Infer preference posterior:
        $(\mu_u,\sigma_u)\leftarrow q_\psi(D_u^{\mathrm{sup}})$

        \STATE Sample latent representation:
        $z_u=\mu_u+\sigma_u\odot\epsilon$,
        $\epsilon\sim\mathcal{N}(0,I)$

        \STATE Generate latent-conditioned LoRA adaptations:\\ \qquad
        $\Delta W_u^{(l)}
        =
        B^{(l)}
        \operatorname{diag}(z_u)
        A^{(l)},
        \quad \forall l=1,\dots,L$

        \STATE Compute adaptation loss:
        $\mathcal{L}_{\mathrm{adapt},u}
        =
        \mathcal{L}
        \left(
        D_u^{\mathrm{qry}};
        \theta_0+\Delta\theta_u^{\mathrm{LoRA}}(z_u)
        \right)$

        \STATE Compute compression loss:
        $\mathcal{L}_{\mathrm{comp},u}
        =
        D_{\mathrm{KL}}
        \left(
        \mathcal{N}(\mu_u,\operatorname{diag}(\sigma_u^2))
        \|
        \mathcal{N}(0,I)
        \right)$

        \STATE Accumulate training objective:
        $\mathcal{L}_{\mathrm{batch}}
        \leftarrow
        \mathcal{L}_{\mathrm{batch}}
        +
        \mathcal{L}_{\mathrm{adapt},u}
        +
        \beta\mathcal{L}_{\mathrm{comp},u}$

    \ENDFOR

    \STATE Update $\psi$ and $\phi$:
    $(\psi,\phi)
    \leftarrow
    (\psi,\phi)
    -
    \eta\nabla_{\psi,\phi}
    \mathcal{L}_{\mathrm{batch}}$

\ENDWHILE

\RETURN $q_\psi$ and shared LoRA parameters $\phi$

\end{algorithmic}
\end{algorithm}

%% file: appendix/app_details.tex
\newpage
\section{Implementation Details}
\label{app:impl_details}

\subsection{Model and Training Configuration}
\label{app:model_config}

\begin{table}[!h]
\centering

\begin{minipage}[t]{0.4\linewidth}
\centering
\caption{Training hyperparameters.}
\label{tab:training_hyperparameters}
\begin{tabular}{lc}
\toprule
\textbf{Hyperparameter} & \textbf{Value} \\
\midrule
LoRA rank $r$            & 16 \\
LoRA scaling $\alpha$    & 32 \\
LoRA dropout             & 0.05 \\
IB coefficient $\beta$   & 0.01 \\
Batch size               & 1 \\
Training epochs          & 3 \\
Learning rate            & $2\times10^{-4}$ \\
\bottomrule
\end{tabular}
\end{minipage}
\hfill
\begin{minipage}[t]{0.58\linewidth}
\centering
\caption{Configuration of the preference encoder.}
\label{tab:preference_encoder}
\begin{tabular}{lc}
\toprule
\textbf{Component} & \textbf{Configuration} \\
\midrule
Hidden dimension        & 512 \\
Number of hidden layers & 2 \\
Activation function     & GELU \\
Dropout                 & 0.1 \\
Mean head               & Linear ($512 \rightarrow d_z$) \\
Log-variance head       & Linear ($512 \rightarrow d_z$) \\
Latent dimension $d_z$  & 16 \\
\bottomrule
\end{tabular}
\end{minipage}

\end{table}

Unless otherwise specified, we use the default configurations of the corresponding pretrained models. For response generation, we use deterministic greedy decoding with \texttt{do\_sample=False}, \texttt{num\_beams=1}, and \texttt{max\_length=512}. Training and test users are strictly disjoint within each dataset. For each test user, 1, 2, 4, 8, or 16 interactions constitute the observed support set for personalization, while 4 additional held-out interactions are reserved exclusively for evaluation. Each experimental setting is evaluated on at least 200 held-out interactions across test users. We first average the evaluation scores across the held-out interactions for each user and then average the resulting user-level scores across all test users.

For each benchmark, we train all methods independently using only the corresponding training split and evaluate them on the held-out test users from the same dataset. This benchmark-specific protocol is applied consistently to all compared methods. Accordingly, the learned latent adaptation space is shared across users within each benchmark, rather than across heterogeneous datasets.

\paragraph{Training support--query construction.}
To align training with the few-shot personalization setting at test time, we adopt an episodic support--query training procedure. For each sampled training user $u$, we randomly sample a support size $K_{\mathrm{sup}}$ from $\{1,2,4,8,16\}$ and construct a support set $D_u^{\mathrm{sup}}$ containing $K_{\mathrm{sup}}$ interactions. We then sample $K_{\mathrm{qry}}=4$ additional interactions from the same user, disjoint from the support set, to form the query set $D_u^{\mathrm{qry}}$. The support set is used exclusively to infer the user-specific latent representation $z_u \sim q_\psi(z_u \mid D_u^{\mathrm{sup}})$, while the query set is used to compute the adaptation loss $\mathcal{L}(D_u^{\mathrm{qry}};\theta_0 + g_\phi(z_u))$. The information-bottleneck regularization is applied to the posterior inferred from $D_u^{\mathrm{sup}}$. Support and query interactions are resampled across training episodes, encouraging the inferred latent representation to capture user-specific preference information that generalizes beyond the particular interactions used for latent inference.

\paragraph{Baseline Selection.}
Our experiments focus on few-shot personalization for unseen users, where only a small support set is available at test time. We therefore select baselines that can be directly evaluated under this protocol. Methods such as OPPU~\citep{tan2024democratizing} and TAP-PER~\citep{cao2026beyond}, which primarily rely on persistent user histories or user-specific states, target a different personalization setting and are therefore not included as direct baselines. PerFit~\citep{liu2026perfit} similarly considers a history-rich personalization setting, selecting users with extensive interaction histories and performing a second-stage optimization of user-specific representation interventions. In contrast, our setting evaluates adaptation to unseen users from only 1--8 observed interactions, without user-specific optimization at deployment. Its reported results are therefore not directly comparable under our few-shot evaluation protocol.
\paragraph{Persona-Iterative-Responses (PIR).}
We use PIR from the publicly available Hugging Face repository
\texttt{sher222/persona-iterative-responses}.\footnote{\url{https://huggingface.co/datasets/sher222/persona-iterative-responses}}

\subsection{Evaluation Metrics}
\label{app:evaluation_metrics}

We evaluate model performance using five metrics that capture both  reference-based generation quality and user-specific personalization.  Specifically, we report \textbf{ROUGE-1}, \textbf{ROUGE-L}, and \textbf{METEOR} for generation quality, and introduce \textbf{Content Personalization (CP)} and \textbf{Presentation Personalization (PP)} to assess personalization from complementary perspectives~\citep{zheng2023judging}. Higher values indicate better performance for all metrics.

\paragraph{ROUGE-1 and ROUGE-L.}
We use ROUGE-1 and ROUGE-L to measure the lexical and sequential overlap between the generated response and the reference response. ROUGE-1 evaluates unigram-level overlap, while ROUGE-L measures similarity based on the longest common subsequence.

\paragraph{METEOR.}
METEOR measures the similarity between the generated and reference responses through token-level alignment, providing a complementary evaluation of lexical and semantic consistency beyond n-gram overlap.

\paragraph{Content Personalization (CP).}
CP evaluates whether the content of a generated response is aligned with the preferences reflected in the user's historical interactions. It considers the selection of topics, examples, information, emphasis, and ordering, and penalizes responses that are generic or inconsistent with the user's inferred interests and goals. Given the user history $H_u$, current input $x$, reference response $y$, and generated response $\hat{y}$, we define the sample-level CP score as
\begin{equation}
s_{\mathrm{CP}}=\mathcal{E}_{\mathrm{CP}}\left(H_u, x, y, \hat{y}\right),
\qquad s_{\mathrm{CP}} \in \{1,\ldots,10\},
\end{equation}
where $\mathcal{E}_{\mathrm{CP}}$ denotes the LLM-based evaluator. A higher CP score indicates stronger alignment between the generated content and the user's preferences.

\paragraph{Presentation Personalization (PP).}
PP evaluates whether the presentation of a generated response matches the patterns reflected in the user's interaction history, including tone, writing style, organization, level of detail, and formatting. Similarly, the sample-level PP score is defined as
\begin{equation}
s_{\mathrm{PP}}
=
\mathcal{E}_{\mathrm{PP}}
\left(H_u, x, y, \hat{y}\right),
\qquad
s_{\mathrm{PP}} \in \{1,\ldots,10\},
\end{equation}
where $\mathcal{E}_{\mathrm{PP}}$ denotes the corresponding LLM-based evaluator. A higher PP score indicates better alignment with the user's preferred presentation style.

For CP and PP, the evaluator is provided with the user history, current input, reference response, and generated response. We report the average score over all evaluation samples:
\begin{equation}
\mathrm{CP}= \frac{1}{N}\sum_{i=1}^{N}s_{\mathrm{CP}}^{(i)},
\qquad
\mathrm{PP}=\frac{1}{N}\sum_{i=1}^{N}s_{\mathrm{PP}}^{(i)},
\end{equation}
where $N$ denotes the number of evaluated samples.

\paragraph{LLM-based Evaluation Protocol.}
For CP and PP, we use \texttt{gpt-4.1-mini} as the LLM-based evaluator. For each evaluation sample, the evaluator receives the user’s historical interactions $H_u$, current input $x$, reference response $y$, and model-generated response $\hat{y}$. The same evaluator model, system prompt, and evaluation procedure are used for all methods and datasets. The evaluator independently assigns integer scores from 1 to 10 for content and presentation personalization, where higher scores indicate stronger personalization. We use deterministic evaluation to reduce randomness in judge outputs. The complete evaluator instruction is provided below for reproducibility.

\begin{quote}
\small
\textbf{System Prompt.}
\par\smallskip

You are an evaluator for personalized text generation. Evaluate the generated response with respect to the user's historical information and the reference response.

Score the response on two dimensions from 1 to 10.

\medskip
\noindent\textbf{1. Content Personalization (CP).}

Evaluate whether the selected content, topics, examples, emphasis, and ordering align with the user's explicit or inferred interests, preferences, and goals.

A score of 1 indicates that the response is generic, user-agnostic, or substantially inconsistent with the user's preferences; a score of 5 indicates moderate personalization, where some content reflects the user's preferences but substantial portions remain generic; and a score of 10 indicates that the content is strongly and consistently tailored to the user's preferences and goals.

\medskip
\noindent\textbf{2. Presentation Personalization (PP).}

Evaluate whether the response matches the user's preferred tone, writing style, organization, level of detail, formatting, and other presentation patterns reflected in the user's history.

A score of 1 indicates little or no alignment with the user's presentation preferences; a score of 5 indicates partial alignment with noticeable generic or mismatched presentation choices; and a score of 10 indicates strong and consistent alignment with the user's preferred presentation style.

\medskip
Use the user history as the primary evidence for personalization and the reference response as additional evidence of the desired response for the current input. Do not reward a response merely for being generally high quality if it does not reflect user-specific preferences.

\medskip
Return only valid JSON in the following format:
\begin{verbatim}
{
  "content_personalization": <integer from 1 to 10>,
  "presentation_personalization": <integer from 1 to 10>
}
\end{verbatim}
\end{quote}

%% file: appendix/app_exp.tex
\begin{table*}[p]
    \centering
    \input{tab/hicupid_performance}
\end{table*}

\begin{table*}[p]
    \centering
    \input{tab/prism_performance}
\end{table*}

\begin{table*}[p]
    \centering
    \input{tab/personamem_performance}
\end{table*}

\begin{table*}[p]
    \centering
    \input{tab/pir_performance}
\end{table*}

\newpage

\section{Additional Experiments}
\label{app:additional_experiments}

\subsection{Extended Personalization Results}
\label{app:additional_performance_results}

The main text presents a subset of representative evaluation metrics for clarity. Here, we provide the full results across all metrics for Llama-3.1-8B-Instruct~\citep{grattafiori2024llama} and further report results for Qwen2.5-7B-Instruct~\citep{qwen2025qwen25technicalreport} to assess whether the observed trends generalize across backbone models.

As shown in Tables~\ref{tab:full_hicupid}-\ref{tab:full_pir}, the full evaluation results are consistent with the observations in the main text. \textit{LatentPersonal} performs particularly well in the low-shot regime on Llama-3.1-8B-Instruct, with strong performance across both lexical and preference-oriented metrics, especially on HiCUPID and PersonaMem. Similar behavior is observed with Qwen2.5-7B-Instruct, although the differences among adaptation-based methods become smaller on several datasets. These results indicate that the observed benefits of latent-guided personalization extend across evaluation metrics and are not specific to a single backbone model. We additionally report 95\% confidence intervals, computed as $\mathrm{mean} \pm 1.96\,\mathrm{SE}$ over held-out test users.

\subsection{Extended Ablation Results}
\label{app:additional_ablation}

Table~\ref{tab:ablation_full} provides the full ablation results across all evaluation metrics and shot settings. Removing the latent representation (\textit{w/o Latent}) leads to a substantial performance degradation across all settings, highlighting the importance of user-specific latent conditioning. Disabling latent navigation (\textit{w/o Navigation}) results in a consistent drop in lexical metrics, particularly in the low-shot regime, while the gap narrows as more observations become available. Removing the information bottleneck (\textit{w/o IB}) causes the largest overall degradation, suggesting that regularizing the latent representation plays an important role in learning effective user-specific adaptation from limited observations. Overall, these results support the complementary roles of latent representation, latent navigation, and information bottleneck regularization in \textit{LatentPersonal}.

\input{tab/ablation_ad.tex}

\newpage

\subsection{Sensitivity Analysis}
\label{app:sensitivity_analysis}

We examine the sensitivity of \textit{LatentPersonal} to two key hyperparameters: the information bottleneck coefficient $\beta$ and the latent dimension $d_z$.

\paragraph{Effect of the Information Bottleneck Coefficient.}
Table~\ref{tab:beta_sensitivity} examines the effect of the information bottleneck coefficient $\beta$ across different numbers of observed interactions. Setting $\beta=0$ leads to a substantial degradation across all metrics and shot settings, highlighting the importance of regularizing the latent representation. With non-zero regularization, performance improves substantially and remains relatively stable across different values of $\beta$, although the optimal value varies across metrics and shot settings. We use $\beta=0.01$ as the default, which provides consistently strong performance, particularly on the lexical metrics.
\input{tab/beta.tex}

\paragraph{Effect of the Latent Dimension.}
Table~\ref{tab:dz_sensitivity} studies the effect of the latent dimension $d_z$ across different numbers of observed interactions. In our LoRA-based instantiation, we set $d_z=r$, such that each latent dimension directly modulates a low-rank adaptation direction. We observe no consistent benefit from increasing the dimensionality: compact latent spaces already achieve strong performance, with $d_z=4$ performing particularly well on several lexical metrics, while larger dimensions provide occasional gains on preference-oriented metrics. Overall, these results suggest that effective personalization does not require a high-dimensional adaptation space, supporting our motivation of navigating a compact latent space.

\input{tab/dz.tex}

\subsection{Test-Time Personalization Efficiency}
\label{app:test_time_efficiency}
We further examine the test-time efficiency of different personalization methods on HiCUPID using Llama-3.1-8B-Instruct. We separately measure \emph{personalization latency}, defined as the method-specific test-time adaptation cost before response generation, and \emph{generation latency}, defined as the subsequent response generation time after personalization. For ICL, personalization consists of constructing the user-specific in-context prompt; for LoRA, it involves gradient-based adaptation on the observed interactions; and for PPlug, Profile-to-PEFT, and \textit{LatentPersonal}, it corresponds to their respective forward personalization procedures. In particular, for \textit{LatentPersonal}, personalization latency measures the forward inference required to obtain the user-specific latent representation from the observed interactions.

\input{tab/time_ad.tex}

As shown in Table~\ref{tab:extended_personalization_efficiency}, \textit{LatentPersonal} introduces only modest personalization overhead, requiring $0.0219$-$0.0676$ seconds across the 1-8-shot settings. Although its personalization latency increases with the number of observed interactions, it remains substantially lower than the gradient-based LoRA adaptation cost across all settings and is lower than Profile-to-PEFT in the 1-4-shot regime. PPlug incurs the lowest personalization latency, reflecting its lightweight forward personalization procedure. Importantly, the generation latency of \textit{LatentPersonal} remains relatively stable as the number of observations increases, indicating that incorporating additional user evidence primarily affects the personalization stage rather than the subsequent generation process. Overall, these results highlight the efficiency advantage of performing user adaptation through forward latent inference without requiring per-user gradient-based optimization.

\subsection{Latent Space Analysis}
\label{app:latent_analysis}

\paragraph{Visualization of the Learned Latent Space}

\begin{figure*}[!th]
    \centering

    \begin{subfigure}[t]{0.48\textwidth}
        \centering
        \includegraphics[width=\linewidth]{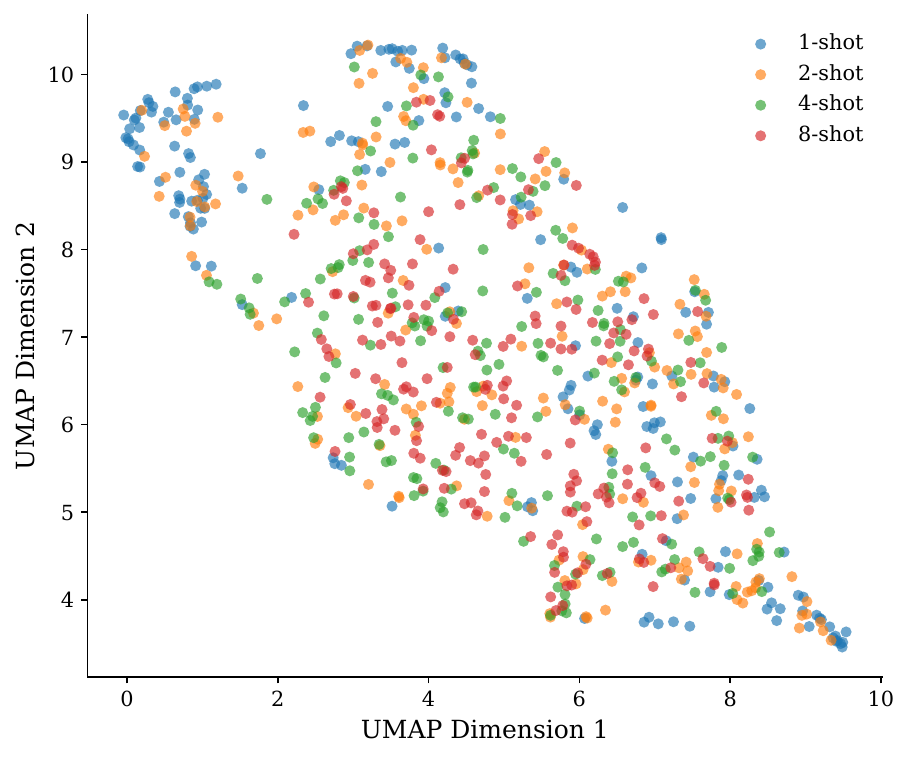}
        \caption{HiCUPID}
    \end{subfigure}
    \hfill
    \begin{subfigure}[t]{0.48\textwidth}
        \centering
        \includegraphics[width=\linewidth]{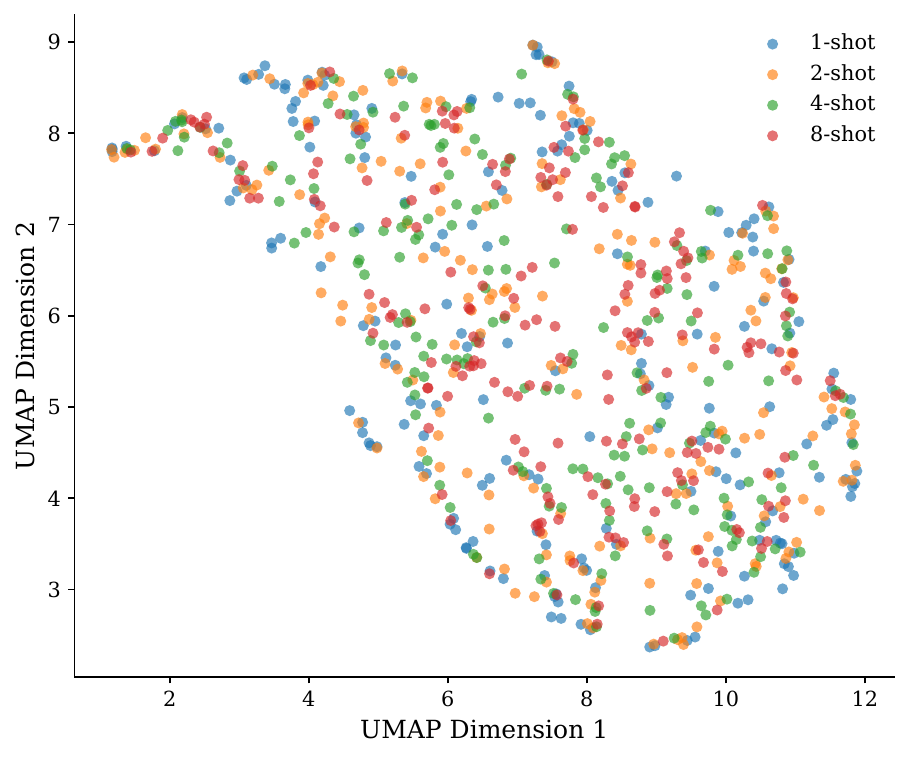}
        \caption{PRISM}
    \end{subfigure}

    \vspace{0.5em}

    \begin{subfigure}[t]{0.48\textwidth}
        \centering
        \includegraphics[width=\linewidth]{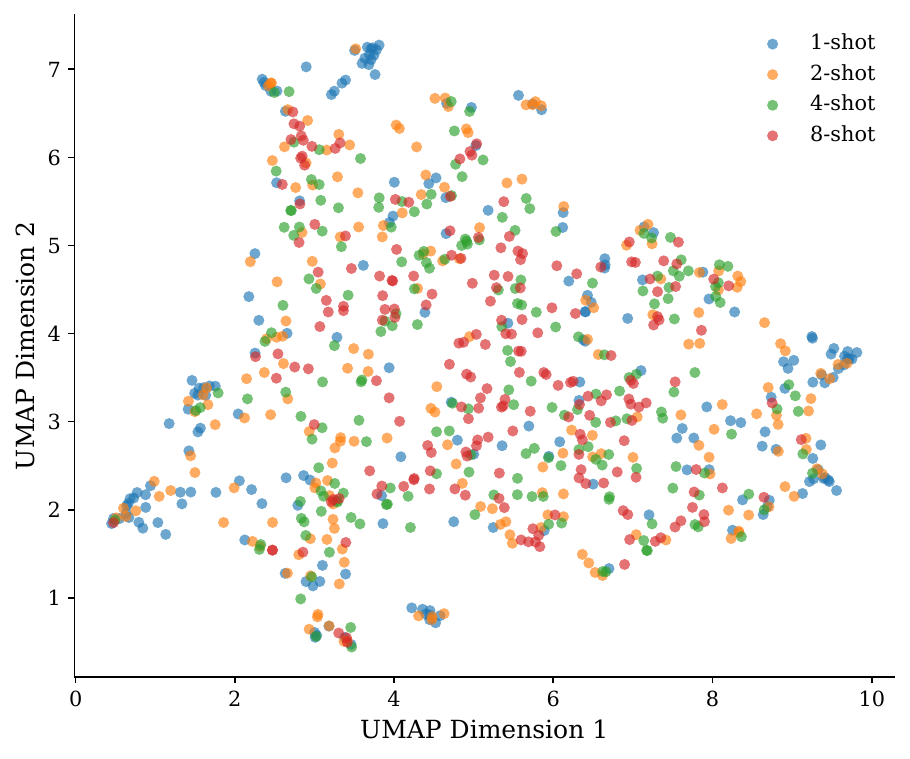}
        \caption{PersonaMem}
    \end{subfigure}
    \hfill
    \begin{subfigure}[t]{0.48\textwidth}
        \centering
        \includegraphics[width=\linewidth]{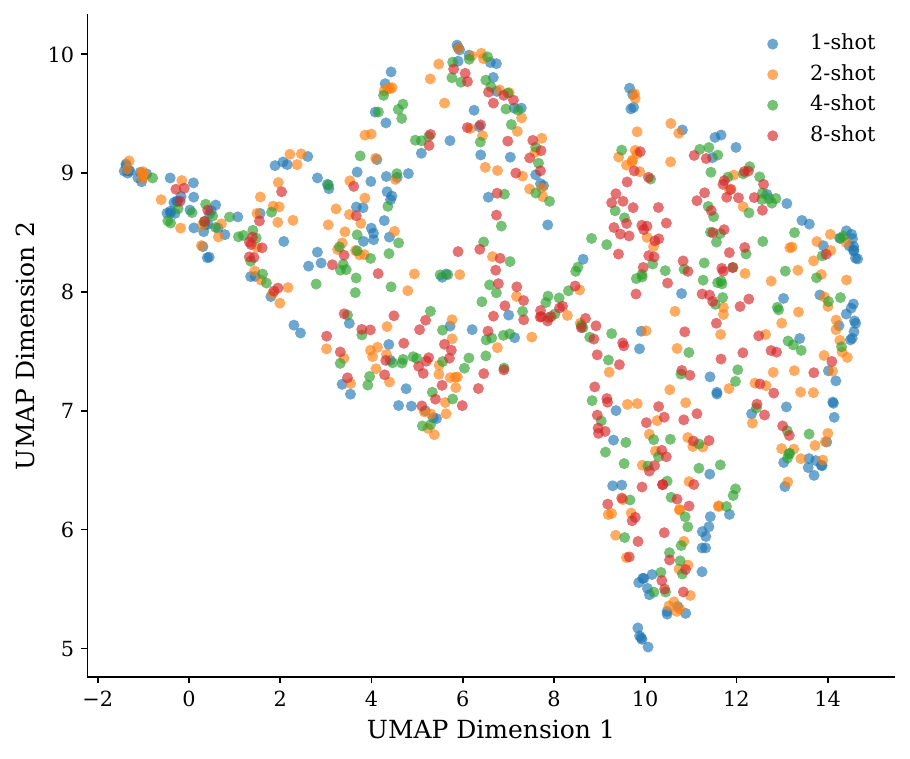}
        \caption{PIR}
    \end{subfigure}

    \caption{
    UMAP visualization of learned user latent representations across four personalization datasets.
    Colors indicate the number of observed user interactions (1, 2, 4, and 8 shots).
    }
    \label{fig:latent_umap}
\end{figure*}

Figure~\ref{fig:latent_umap} visualizes the learned user representations under different numbers of observed interactions. Across all four datasets, representations inferred from different shot settings largely occupy a shared latent structure rather than forming isolated shot-specific clusters. At the same time, their distributions gradually shift as additional interactions are incorporated. This pattern is consistent with our view of personalization as navigation within a shared latent space, where additional user evidence refines the inferred position rather than inducing a separate representation space for each interaction regime.

\begin{figure*}[!th]
    \centering

    \begin{subfigure}[t]{0.48\textwidth}
        \centering
        \includegraphics[width=\linewidth]{pic/box_plot/hicupid.pdf}
        \caption{HiCUPID}
    \end{subfigure}
    \hfill
    \begin{subfigure}[t]{0.48\textwidth}
        \centering
        \includegraphics[width=\linewidth]{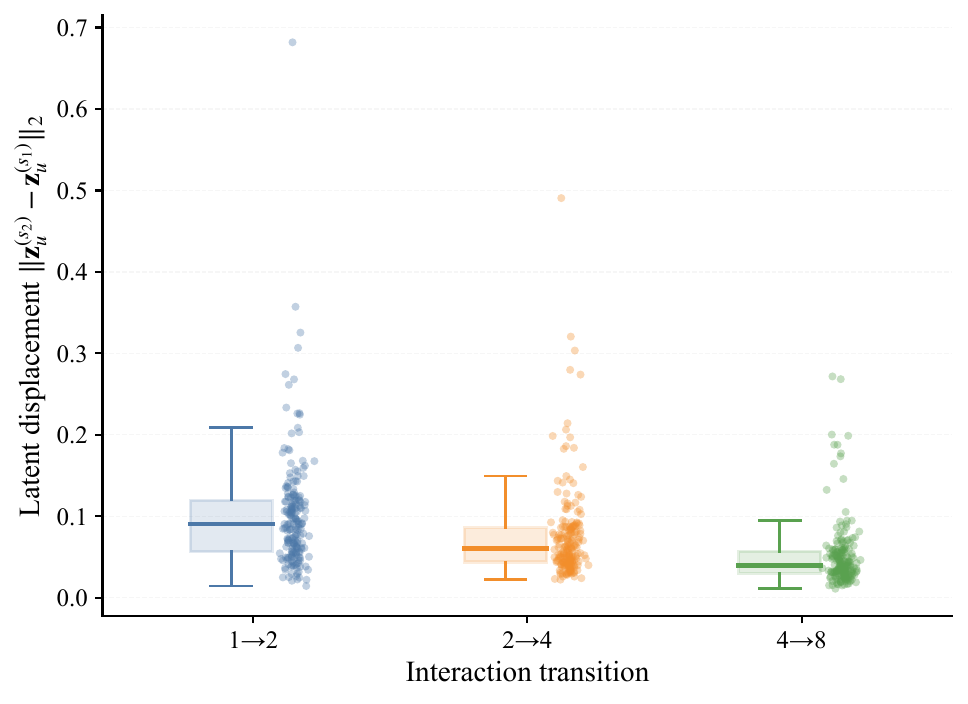}
        \caption{PRISM}
    \end{subfigure}

    \vspace{0.5em}

    \begin{subfigure}[t]{0.48\textwidth}
        \centering
        \includegraphics[width=\linewidth]{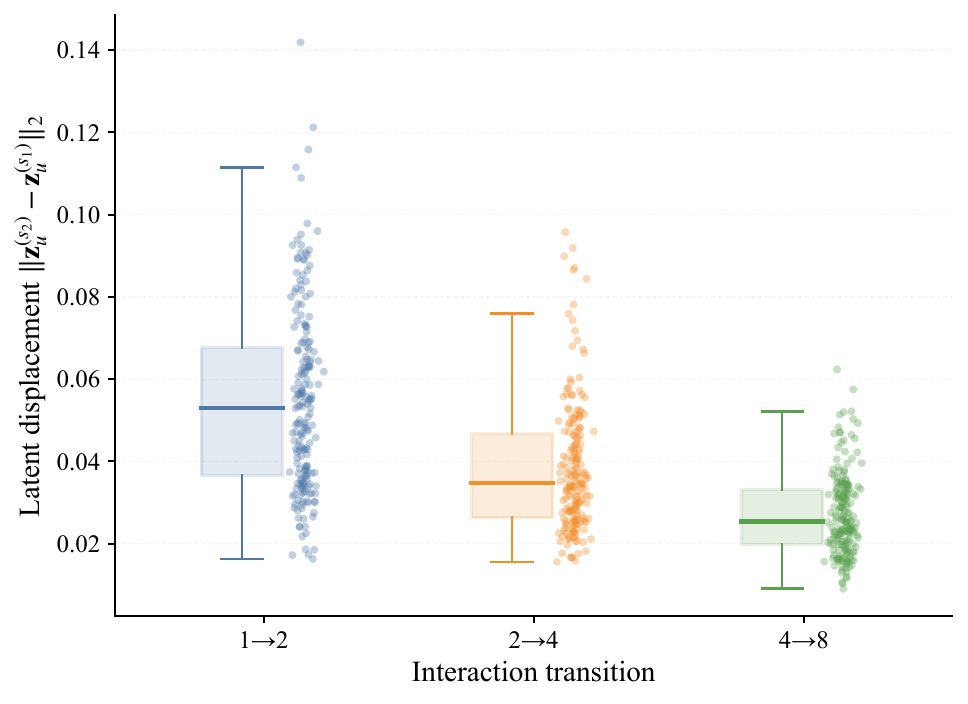}
        \caption{PersonaMem}
    \end{subfigure}
    \hfill
    \begin{subfigure}[t]{0.48\textwidth}
        \centering
        \includegraphics[width=\linewidth]{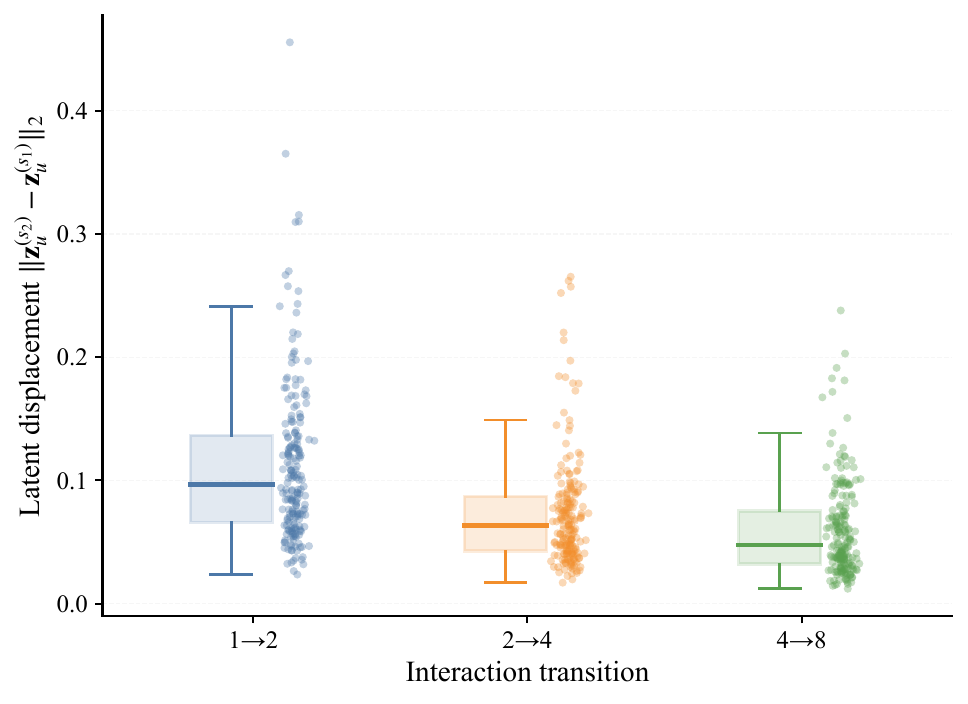}
        \caption{PIR}
    \end{subfigure}

    \caption{
    Latent displacement across successive interaction transitions on four personalization datasets.
    Lower displacement at later transitions indicates that the inferred user representation gradually stabilizes as more interactions are observed.
    }
    \label{fig:latent_displacement}
\end{figure*}

\newpage
\paragraph{Latent Stabilization across Interactions.} To quantify how the inferred user representation evolves as additional interactions are observed, we measure the latent displacement $\|\mathbf{z}_u^{(s_2)}-\mathbf{z}_u^{(s_1)}\|_2$ for each user between consecutive shot settings, i.e., $1\!\rightarrow\!2$, $2\!\rightarrow\!4$, and $4\!\rightarrow\!8$. As shown in Figure~\ref{fig:latent_displacement}, the displacement progressively decreases across interaction transitions on all four datasets, with the largest changes generally occurring in the earliest stage. This suggests that the first few interactions substantially update the inferred user position, whereas subsequent observations increasingly refine an already established latent representation.

\begin{figure*}[!th]
    \centering

    \begin{subfigure}[t]{0.48\textwidth}
        \centering
        \includegraphics[width=\linewidth]{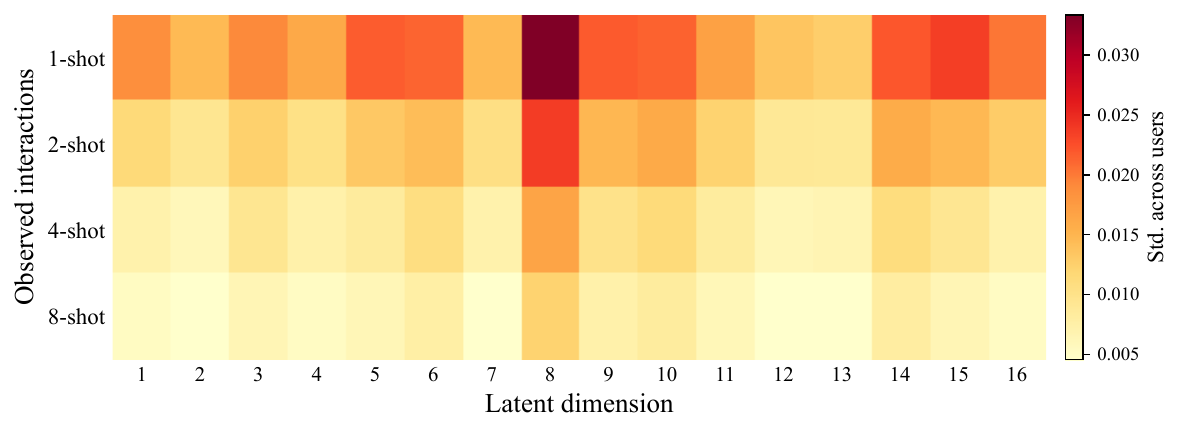}
        \caption{HiCUPID}
    \end{subfigure}
    \hfill
    \begin{subfigure}[t]{0.48\textwidth}
        \centering
        \includegraphics[width=\linewidth]{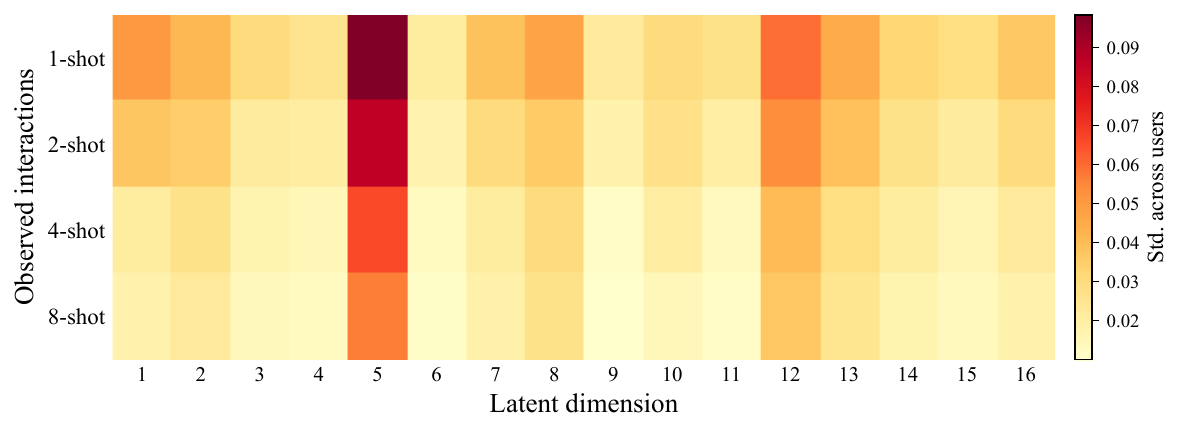}
        \caption{PRISM}
    \end{subfigure}

    \vspace{0.5em}

    \begin{subfigure}[t]{0.48\textwidth}
        \centering
        \includegraphics[width=\linewidth]{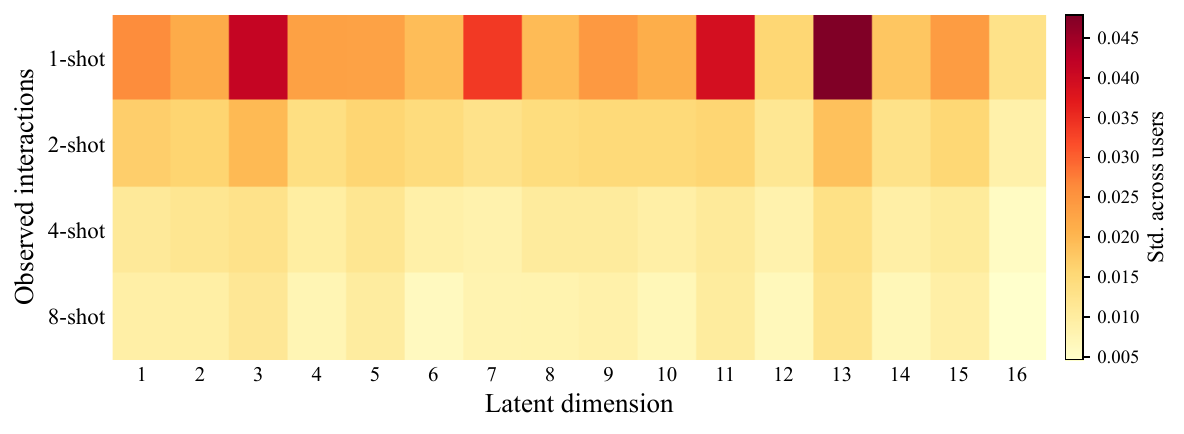}
        \caption{PersonaMem}
    \end{subfigure}
    \hfill
    \begin{subfigure}[t]{0.48\textwidth}
        \centering
        \includegraphics[width=\linewidth]{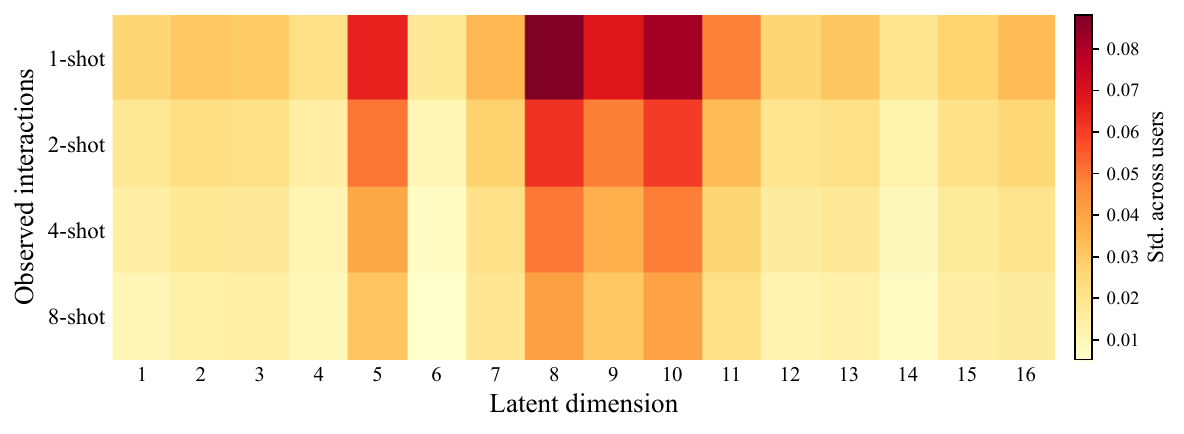}
        \caption{PIR}
    \end{subfigure}

    \caption{
    Dimension-wise variability of learned user latent representations across
    different numbers of observed interactions on four personalization datasets.
    Each cell reports the standard deviation of a latent dimension across users.
    }
    \label{fig:latent_dimension_variability}
\end{figure*}

\paragraph{Dimension-wise Latent Variability.}
We further examine how user-specific variation is distributed across the latent dimensions. For each shot setting and latent dimension, we compute the standard deviation of the corresponding latent coordinate across users. As shown in Figure~\ref{fig:latent_dimension_variability}, the variability is unevenly distributed across dimensions, with a subset of dimensions exhibiting substantially larger cross-user variation. Moreover, the overall variability tends to decrease as more interactions are observed across all four datasets. These results suggest that user-specific variation is concentrated along particular latent coordinates, while the inferred representations become progressively more stable as additional user evidence is incorporated.

\paragraph{Cross-shot Latent Alignment.}

We next examine whether the organization of the latent space is preserved as additional user interactions are incorporated. For each pair of consecutive shot settings, we compute the Pearson correlation across users between every pair of latent dimensions, producing a $d_z\times d_z$ cross-shot correlation matrix. A strong diagonal entry therefore indicates that the same latent coordinate preserves similar user-wise variation across two consecutive interaction regimes, whereas off-diagonal correlations capture dependencies between different coordinates.

\begin{figure*}[p]
    \centering

    \begin{subfigure}[t]{0.30\textwidth}
        \centering
        \includegraphics[width=\linewidth]{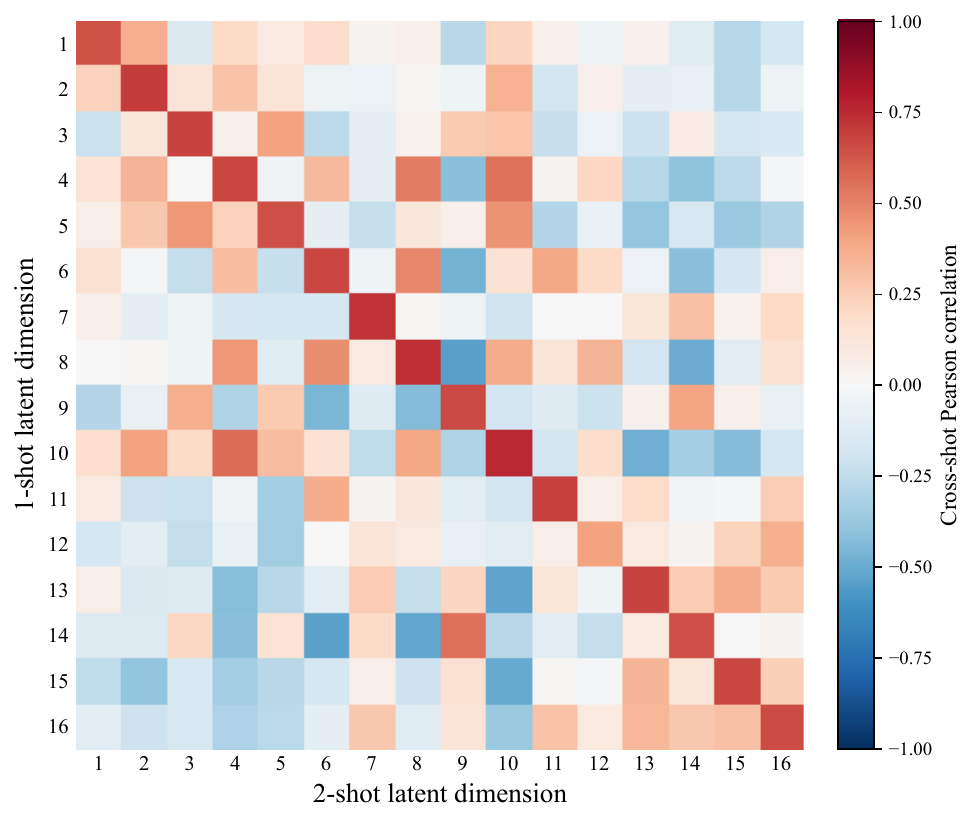}
        \caption{HiCUPID: $1 \rightarrow 2$}
    \end{subfigure}
    \hfill
    \begin{subfigure}[t]{0.30\textwidth}
        \centering
        \includegraphics[width=\linewidth]{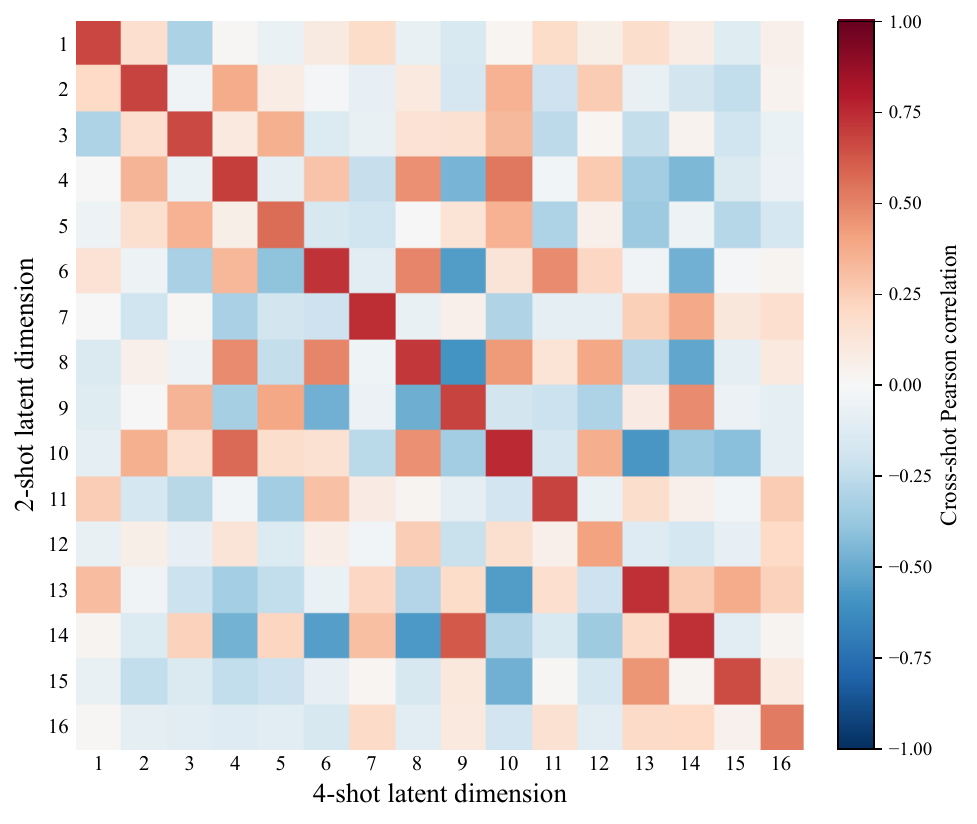}
        \caption{HiCUPID: $2 \rightarrow 4$}
    \end{subfigure}
    \hfill
    \begin{subfigure}[t]{0.30\textwidth}
        \centering
        \includegraphics[width=\linewidth]{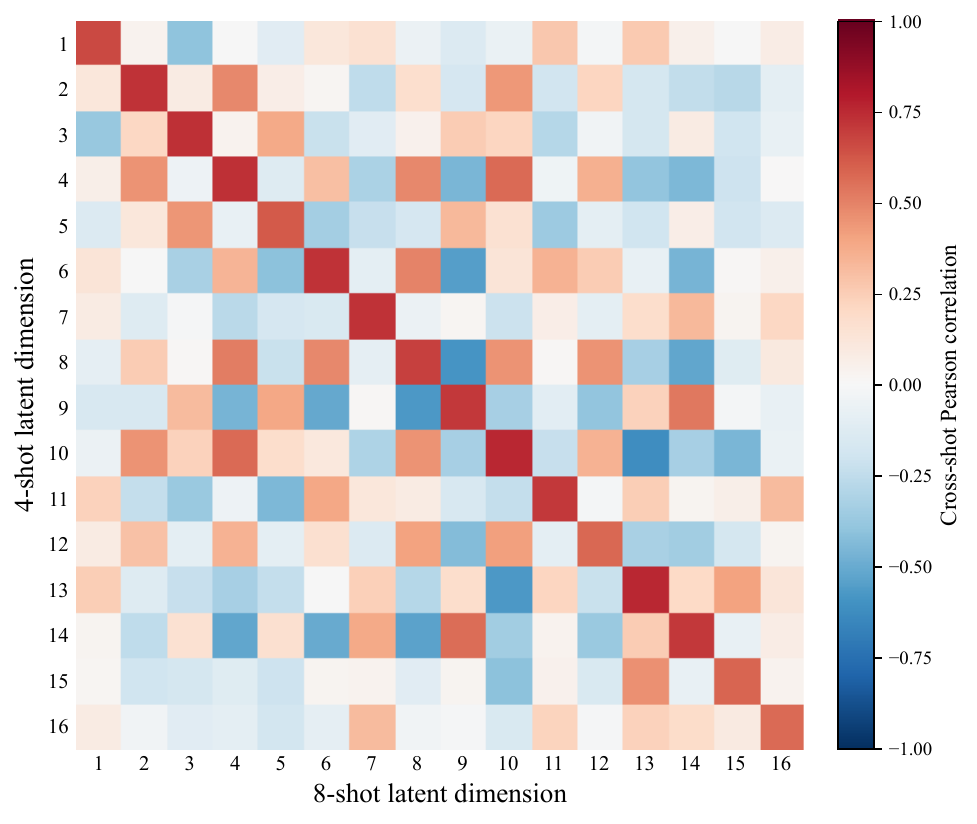}
        \caption{HiCUPID: $4 \rightarrow 8$}
    \end{subfigure}

    \vspace{0.8em}

    \begin{subfigure}[t]{0.30\textwidth}
        \centering
        \includegraphics[width=\linewidth]{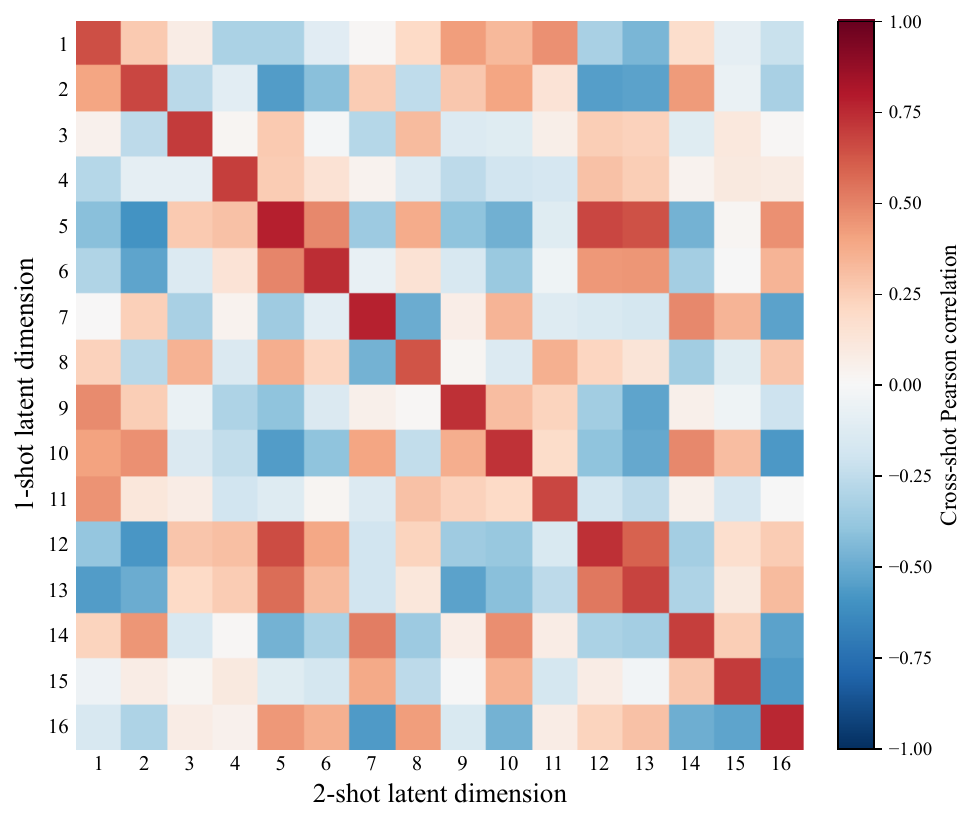}
        \caption{PRISM: $1 \rightarrow 2$}
    \end{subfigure}
    \hfill
    \begin{subfigure}[t]{0.30\textwidth}
        \centering
        \includegraphics[width=\linewidth]{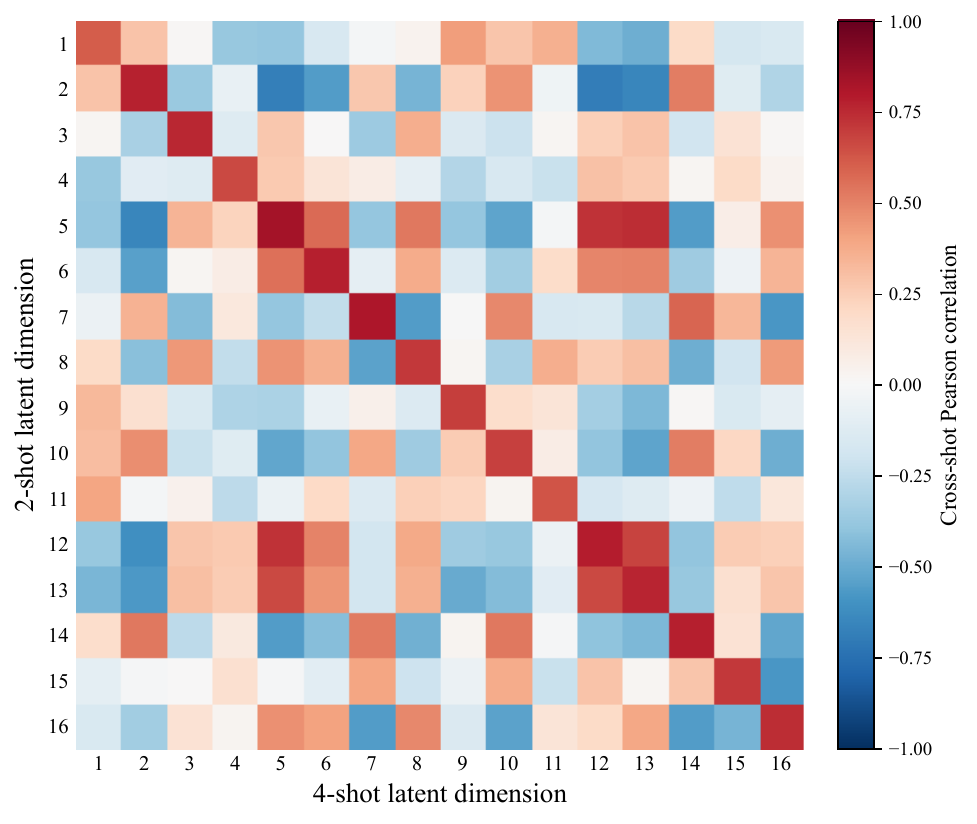}
        \caption{PRISM: $2 \rightarrow 4$}
    \end{subfigure}
    \hfill
    \begin{subfigure}[t]{0.30\textwidth}
        \centering
        \includegraphics[width=\linewidth]{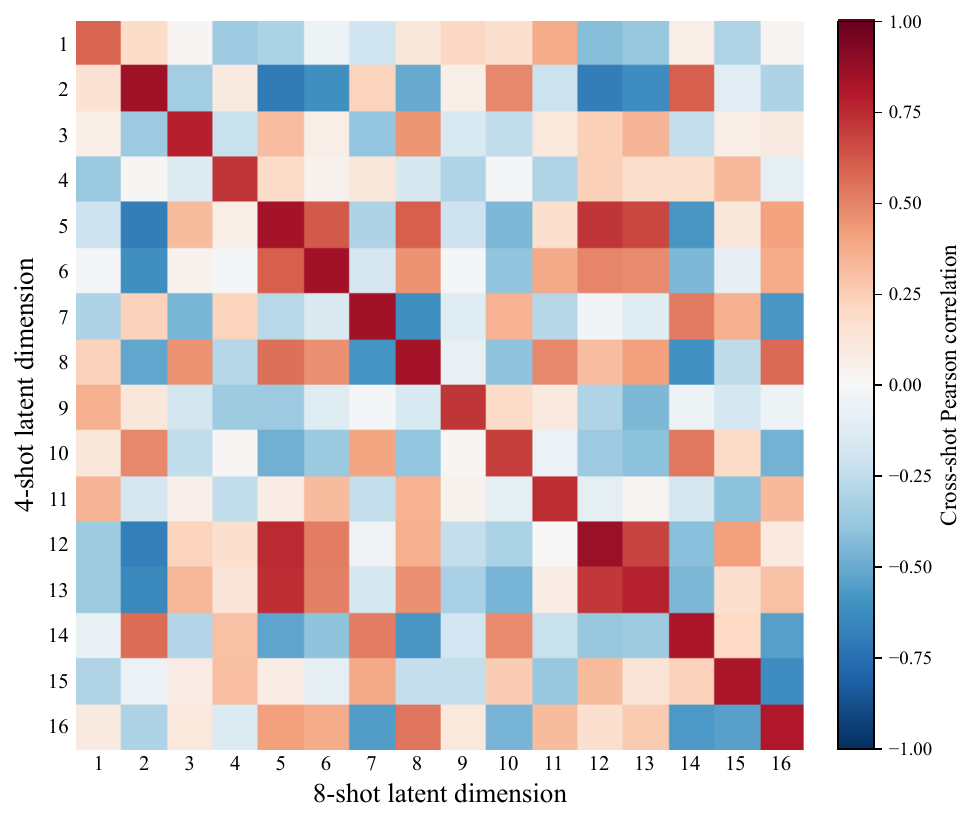}
        \caption{PRISM: $4 \rightarrow 8$}
    \end{subfigure}

    \vspace{0.8em}

    \begin{subfigure}[t]{0.30\textwidth}
        \centering
        \includegraphics[width=\linewidth]{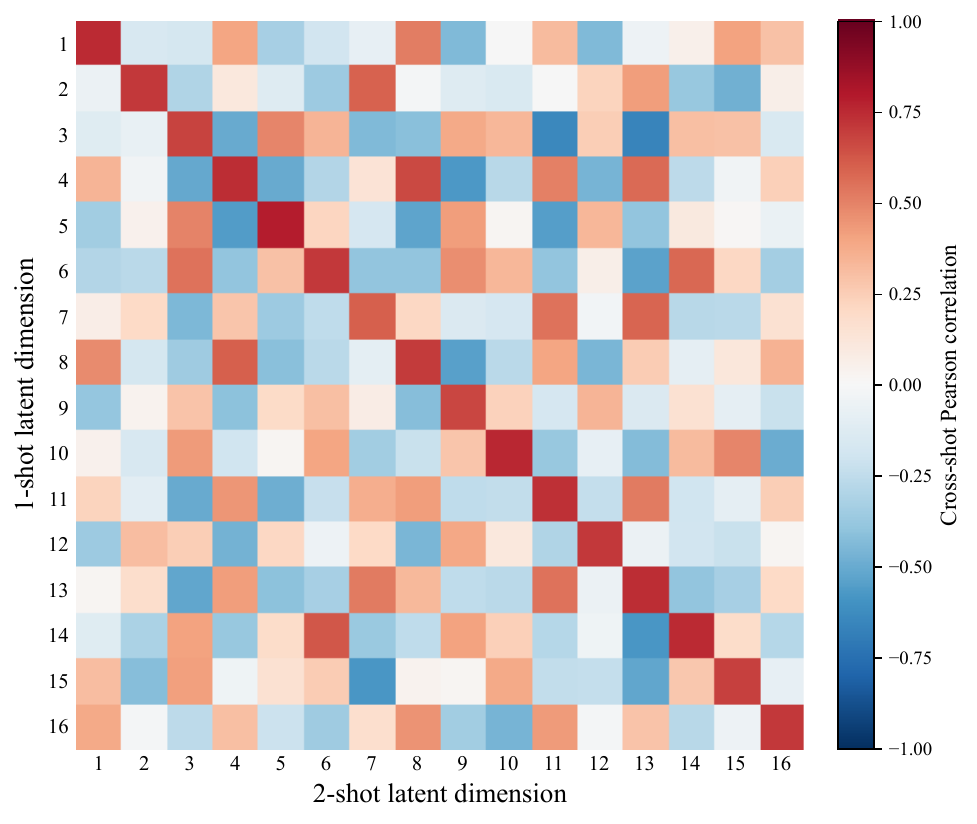}
        \caption{PersonaMem: $1 \rightarrow 2$}
    \end{subfigure}
    \hfill
    \begin{subfigure}[t]{0.30\textwidth}
        \centering
        \includegraphics[width=\linewidth]{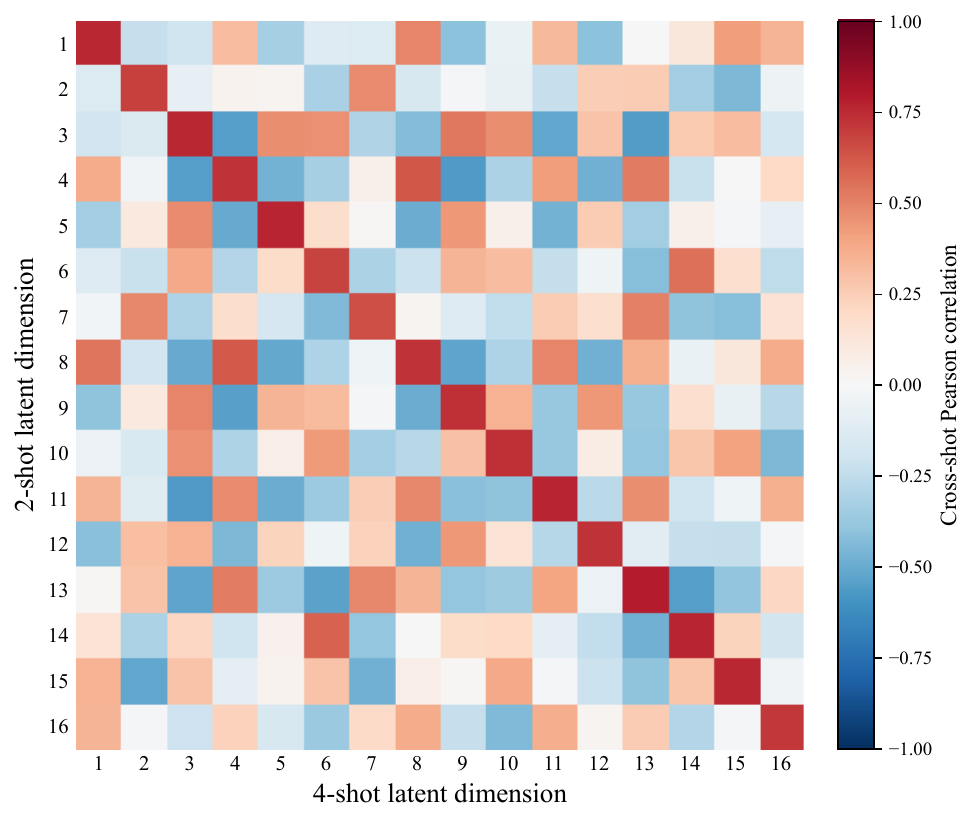}
        \caption{PersonaMem: $2 \rightarrow 4$}
    \end{subfigure}
    \hfill
    \begin{subfigure}[t]{0.30\textwidth}
        \centering
        \includegraphics[width=\linewidth]{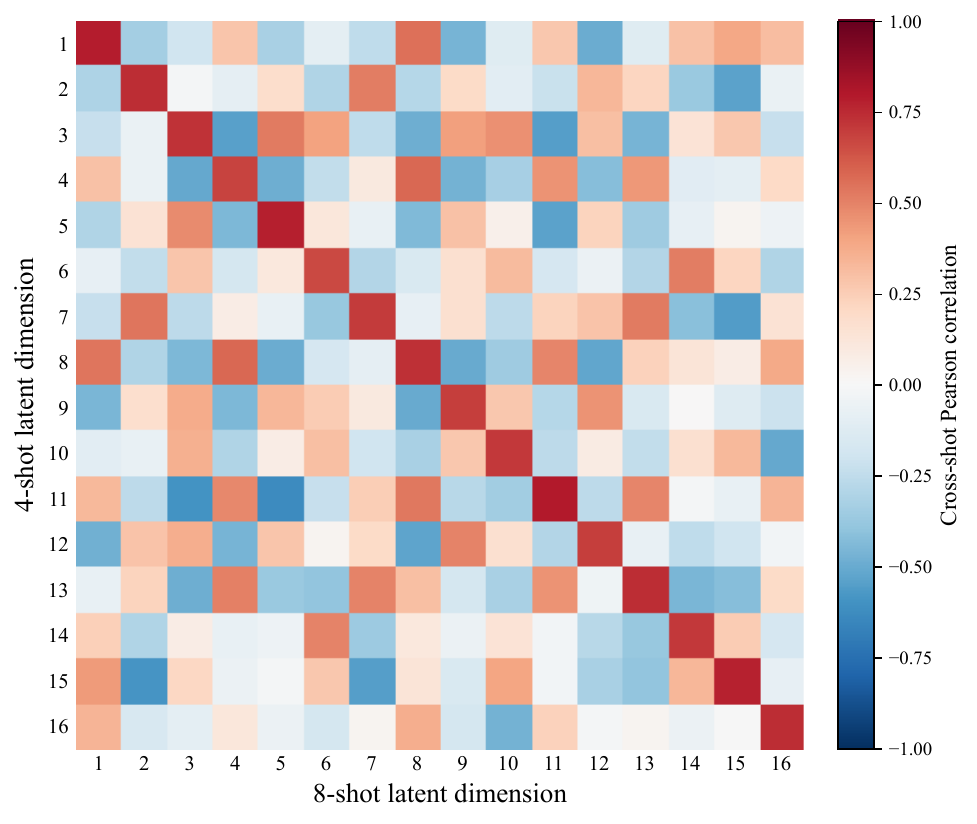}
        \caption{PersonaMem: $4 \rightarrow 8$}
    \end{subfigure}

    \vspace{0.8em}

    \begin{subfigure}[t]{0.30\textwidth}
        \centering
        \includegraphics[width=\linewidth]{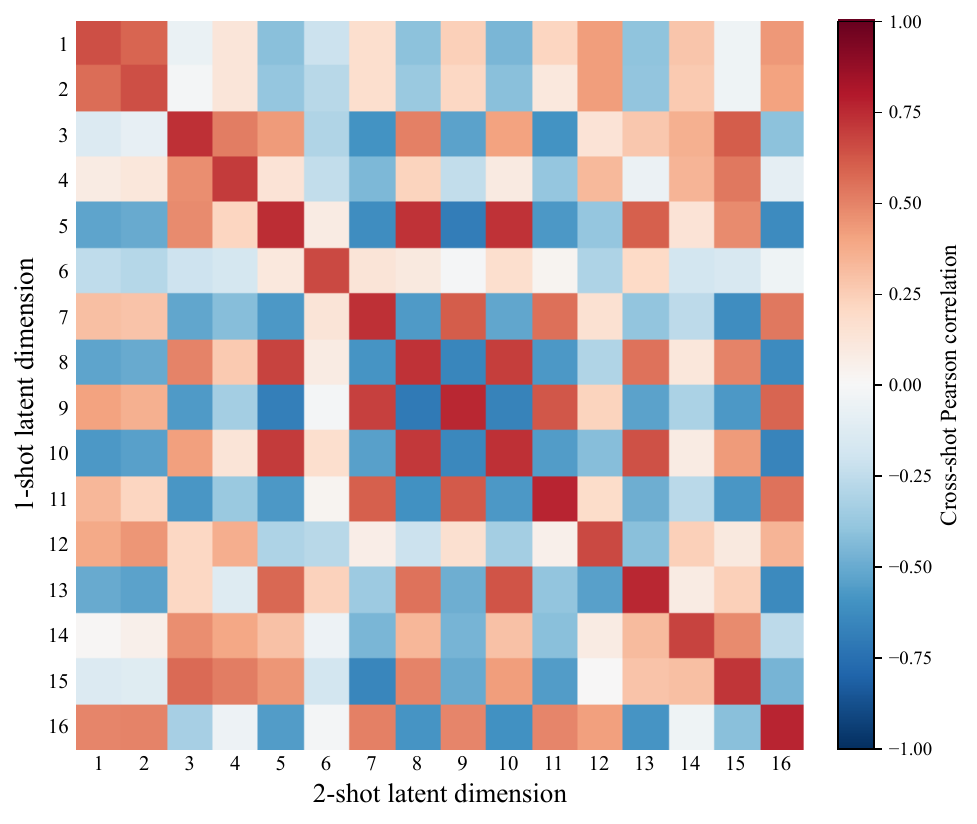}
        \caption{PIR: $1 \rightarrow 2$}
    \end{subfigure}
    \hfill
    \begin{subfigure}[t]{0.30\textwidth}
        \centering
        \includegraphics[width=\linewidth]{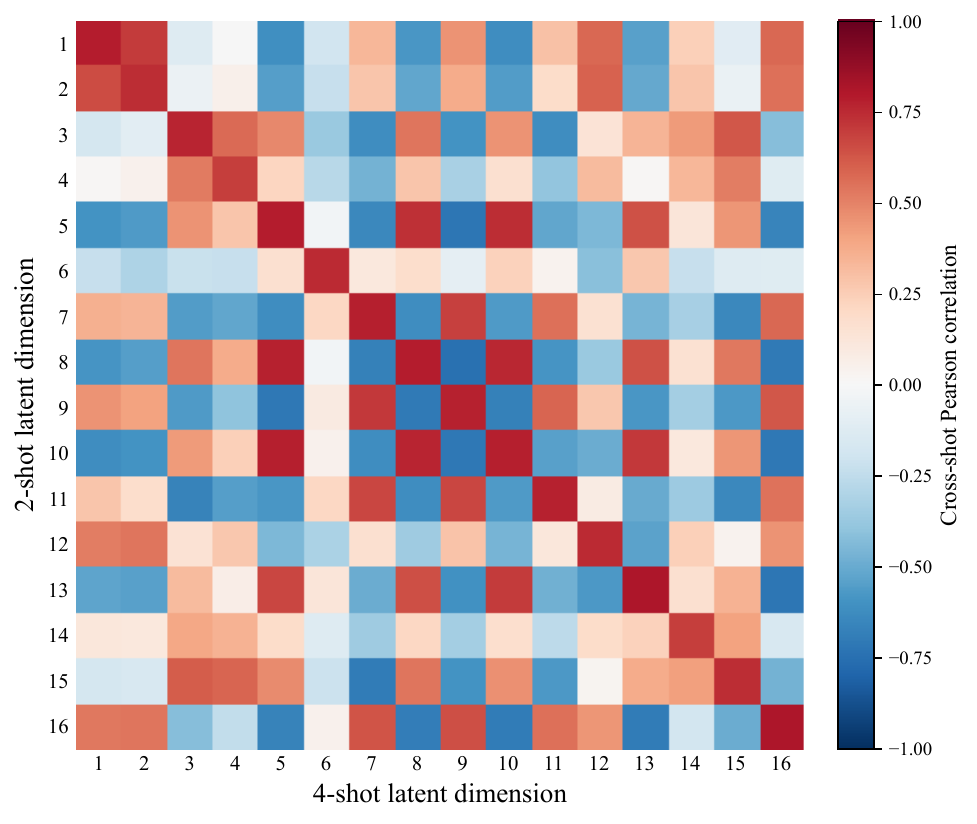}
        \caption{PIR: $2 \rightarrow 4$}
    \end{subfigure}
    \hfill
    \begin{subfigure}[t]{0.30\textwidth}
        \centering
        \includegraphics[width=\linewidth]{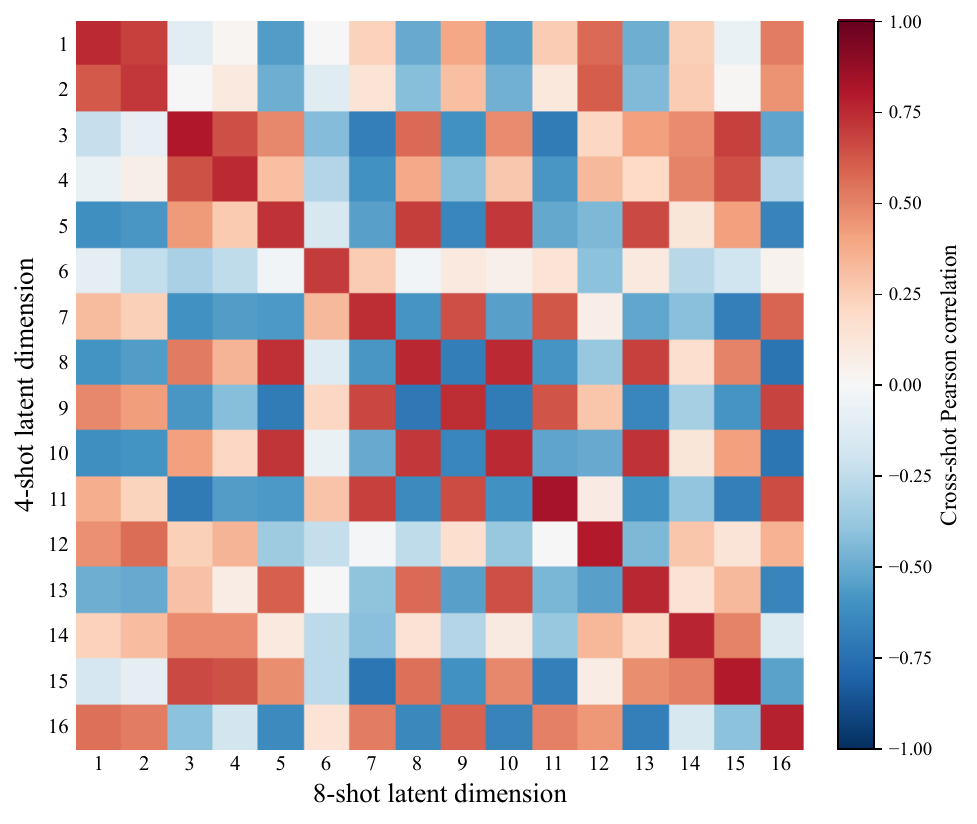}
        \caption{PIR: $4 \rightarrow 8$}
    \end{subfigure}

    \caption{
    Cross-shot alignment of learned latent dimensions across personalization datasets.
    Each matrix reports the Pearson correlation across users between latent dimensions inferred under two consecutive shot settings.
    The diagonal structure indicates consistent alignment of corresponding latent coordinates as additional user interactions are observed.
    }
    \label{fig:cross_shot_alignment}
\end{figure*}

As shown in Figure~\ref{fig:cross_shot_alignment}, a clear diagonal structure persists from $1\rightarrow 2$ through $4 \rightarrow 8$ across the datasets, indicating that corresponding latent coordinates remain aligned rather than being arbitrarily reorganized as more observations are introduced. The strength and surrounding correlation structure nevertheless differ across datasets. HiCUPID exhibits particularly clear diagonal correspondence with comparatively moderate off-diagonal correlations, while PersonaMem also maintains a stable coordinate-wise pattern across successive transitions. PIR exhibits substantially stronger positive and negative off-diagonal structure, suggesting greater coupling among latent coordinates, while still retaining correspondence along the diagonal. These dataset-specific patterns indicate that the latent representation need not consist of independent dimensions; instead, different coordinates can evolve jointly while preserving a recognizable cross-shot organization.

Together with the decreasing latent displacement in Figure~\ref{fig:latent_displacement}, these results provide complementary evidence for a progressively refined representation: additional interactions move the inferred user representation while largely preserving the underlying latent coordinate structure. This behavior is consistent with the view of personalization as navigation within a shared latent adaptation space rather than repeatedly constructing unrelated representations as new user evidence becomes available.

\paragraph{Latent Scaling Analysis.}

We further examine the navigation interpretation underlying \textit{LatentPersonal}. In our formulation, navigation refers to controlling user-specific adaptation through movement in the learned latent space, where the inferred $\mathbf{z}_u$ specifies a user-dependent direction in the shared adaptation space. To directly probe this behavior, we scale the inferred user latent as $\tilde{\mathbf{z}}_u=\alpha\mathbf{z}_u$, where $\alpha\in\{0,0.25,0.5,0.75,1,1.5,2\}$.
This intervention preserves the inferred direction while varying how far the model moves along it: $\alpha=0$ removes the user-specific latent signal, $\alpha=1$ recovers the original representation, and $\alpha>1$ moves beyond the inferred operating point along the same direction. It therefore provides a direct way to examine whether controlled movement in the latent space induces systematic changes in personalized behavior.

As shown in Table~\ref{tab:alpha_scaling}, removing the latent signal ($\alpha=0$) causes a substantial performance degradation across all shot settings and evaluation metrics. Performance recovers rapidly as the model moves along the inferred direction, with $\alpha=0.5$ already approaching the original representation ($\alpha=1$) on several metrics. The inferred operating point at $\alpha=1$ provides consistently strong performance across shot settings, whereas moving further along the same direction ($\alpha>1$) yields no systematic improvement and generally degrades performance. This non-monotonic behavior indicates that personalization depends not only on identifying a user-specific direction but also on the position along that direction: moving away from the origin rapidly recovers personalized behavior, while excessive movement can move the model beyond a favorable personalization region. Taken together, these controlled interventions show that movement along the inferred latent direction produces systematic changes in personalization behavior, providing empirical support for interpreting \textit{LatentPersonal} as navigation in a shared adaptation space.

\input{tab/alpha.tex}

\paragraph{User-Specificity Analysis.}
The latent-space visualization in Figure~\ref{fig:latent_umap} suggests that different users occupy distinct regions within a shared latent structure. We therefore further examine whether these observed differences correspond to genuinely user-specific adaptation, rather than merely reflecting population-level adaptation learned from the training users. For each target test user $u$, we keep the trained model and corresponding held-out evaluation queries fixed and compare four different adaptation conditions: 
(i) \textit{Matched}, using the latent representation $\mathbf{z}_u$ inferred from the target user's support interactions; 
(ii) \textit{Swapped}, replacing $\mathbf{z}_u$ with a latent representation $\mathbf{z}_v$ inferred from another test user $v\neq u$; 
(iii) \textit{Mean}, using a population-level latent representation $\bar{\mathbf{z}}$ as generic conditioning; and 
(iv) \textit{Shared LoRA}, a separately trained LoRA model without any user-specific latent conditioning. 
The first three conditions share exactly the same trained \textit{LatentPersonal} model and differ only in the latent representation supplied during inference. This controlled comparison isolates whether matching the inferred latent representation to the target user provides additional personalization benefits beyond mismatched or population-level conditioning, while the shared LoRA control measures the contribution of user-specific latent modulation beyond shared parameter adaptation alone.

For the \textit{Swapped} condition, we replace the target user's representation $\mathbf{z}_u$ with $\mathbf{z}_v$ inferred from the support interactions of another test user $v\neq u$, while keeping the target user's original held-out evaluation queries unchanged. For each target user, we randomly sample five other users as independent swap sources and average the resulting scores to reduce sensitivity to any particular user pairing. The same target--source user mappings are maintained across all shot settings, so that only the number of support interactions used to infer each representation changes. For the \textit{Mean} condition, we use the average latent representation computed over the training-user population, providing a generic population-level latent-conditioning control without access to the target user's inferred representation. For the \textit{Shared LoRA} condition, we train a separate shared LoRA model on the same training-user data and under the same training budget as \textit{LatentPersonal}, while removing user-specific latent conditioning throughout the entire training process. Equivalently, the latent modulation is fixed to $\mathbf{z}=\mathbf{1}$ for every user, yielding $B\operatorname{diag}(\mathbf{1})A=BA$. This provides a population-level adaptation control that retains the same low-rank adaptation structure but contains no user-specific latent information during personalization.

\input{tab/user.tex}

The matched representation $\mathbf{z}_u$ consistently achieves the strongest performance across all shot settings, while replacing it with either a mismatched representation $\mathbf{z}_v$ or the population mean $\bar{\mathbf{z}}$ leads to clear degradation. Notably, the swapped condition exhibits a substantially larger drop in R-1, R-L, and METEOR than in CP and PP. This discrepancy is consistent with the different sensitivities of these metrics: reference-based metrics directly penalize deviations from the target user’s response, whereas CP and PP can remain relatively high when the generated response still follows broadly plausible personalization patterns shared across users. In contrast, the population-mean representation avoids navigating toward an incorrect individual user and therefore yields more stable performance than the swapped representation on most reference-based metrics. These results support the navigation view of \textit{LatentPersonal}: effective personalization depends not only on entering the learned adaptation space, but on navigating toward the region associated with the target user.

\subsection{Evaluation across PEFT Architectures}
\label{app:peft_evaluation}

To examine whether the effectiveness of \textit{LatentPersonal} depends on the specific PEFT architecture, we further evaluate the framework with Adapter and Prompt Tuning as alternative parameterizations. Their corresponding instantiations are described in Appendix~\ref{app:peft_architectures}.

\input{tab/peft_evaluation.tex}

As shown in Table~\ref{tab:peft_evaluation}, incorporating \textit{LatentPersonal} into Adapter consistently improves the lexical metrics across all shot settings, with particularly pronounced gains in the low-shot regime. For Prompt Tuning, latent conditioning similarly provides substantial improvements in the lexical metrics when only a few user interactions are available, while the advantage generally narrows as more observations are introduced. Preference-oriented metrics exhibit a more mixed pattern, indicating that the benefit of latent conditioning can depend on both the underlying PEFT parameterization and the evaluation criterion. Overall, the improvements observed with both Adapter and Prompt Tuning, particularly under limited user supervision, suggest that the latent personalization framework is not tied to the LoRA-based instantiation used in our main experiments and can be applied to different PEFT adaptation spaces.

%% file: tab/hicupid_performance.tex
\captionof{table}{Full personalization results on HiCUPID under different few-shot settings.}
\label{tab:full_hicupid}

\vspace{2pt}

\scriptsize
\setlength{\tabcolsep}{2.8pt}
\renewcommand{\arraystretch}{1.05}

\begin{tabular}{llccccc}
\toprule

\textbf{Method}
& \textbf{Shot}
& \textbf{R-1}
& \textbf{R-L}
& \textbf{MET.}
& \textbf{CP}
& \textbf{PP} \\

\midrule

\multicolumn{7}{c}{\textbf{Llama-3.1-8B-Instruct}} \\
\midrule

Base
& / & 0.1614 $\pm$ 0.0124 & 0.1097 $\pm$ 0.0083 & 0.2166 $\pm$ 0.0136 & 3.57 $\pm$ 0.30 & 4.29 $\pm$ 0.23 \\

\midrule

ICL
& 1  & 0.1713 $\pm$ 0.0148 & 0.1086 $\pm$ 0.0089 & 0.2377 $\pm$ 0.0158 & 3.85 $\pm$ 0.30 & 4.91 $\pm$ 0.21 \\
& 2  & 0.1682 $\pm$ 0.0152 & 0.1064 $\pm$ 0.0088 & 0.2294 $\pm$ 0.0169 & 4.25 $\pm$ 0.34 & 5.08 $\pm$ 0.25 \\
& 4  & 0.1782 $\pm$ 0.0152 & 0.1185 $\pm$ 0.0103 & 0.2429 $\pm$ 0.0166 & 4.64 $\pm$ 0.36 & 5.20 $\pm$ 0.26 \\
& 8  & 0.1934 $\pm$ 0.0189 & 0.1352 $\pm$ 0.0147 & 0.2730 $\pm$ 0.0209 & 5.52 $\pm$ 0.43 & 5.63 $\pm$ 0.34 \\
& 16 & 0.2114 $\pm$ 0.0221 & 0.1634 $\pm$ 0.0210 & 0.3090 $\pm$ 0.0272 & 6.55 $\pm$ 0.44 & 6.72 $\pm$ 0.38 \\

\midrule

LoRA
& 1  & 0.3886 $\pm$ 0.0412 & 0.3390 $\pm$ 0.0401 & 0.3829 $\pm$ 0.0423 & 5.57 $\pm$ 0.39 & 6.01 $\pm$ 0.26 \\
& 2  & 0.3993 $\pm$ 0.0406 & 0.3541 $\pm$ 0.0396 & 0.3940 $\pm$ 0.0416 & 5.73 $\pm$ 0.39 & 6.19 $\pm$ 0.30 \\
& 4  & 0.3912 $\pm$ 0.0400 & 0.3468 $\pm$ 0.0398 & 0.3923 $\pm$ 0.0413 & 5.61 $\pm$ 0.43 & 6.20 $\pm$ 0.32 \\
& 8  & 0.3984 $\pm$ 0.0392 & 0.3434 $\pm$ 0.0393 & 0.4027 $\pm$ 0.0393 & 6.26 $\pm$ 0.39 & 6.52 $\pm$ 0.31 \\
& 16 & 0.4223 $\pm$ 0.0415 & 0.3729 $\pm$ 0.0426 & 0.4196 $\pm$ 0.0431 & 6.55 $\pm$ 0.44 & 6.83 $\pm$ 0.34 \\

\midrule

PPlug
& 1  & 0.3018 $\pm$ 0.0310 & 0.2395 $\pm$ 0.0278 & 0.2696 $\pm$ 0.0294 & 5.15 $\pm$ 0.41 & 5.42 $\pm$ 0.30 \\
& 2  & 0.3116 $\pm$ 0.0319 & 0.2529 $\pm$ 0.0294 & 0.2813 $\pm$ 0.0303 & 5.37 $\pm$ 0.42 & 5.60 $\pm$ 0.32 \\
& 4  & 0.3276 $\pm$ 0.0356 & 0.2690 $\pm$ 0.0327 & 0.3018 $\pm$ 0.0349 & 5.63 $\pm$ 0.40 & 5.68 $\pm$ 0.32 \\
& 8  & 0.3315 $\pm$ 0.0372 & 0.2715 $\pm$ 0.0331 & 0.3105 $\pm$ 0.0347 & 5.95 $\pm$ 0.38 & 6.14 $\pm$ 0.30 \\
& 16 & 0.3763 $\pm$ 0.0405 & 0.3137 $\pm$ 0.0373 & 0.3515 $\pm$ 0.0387 & 6.38 $\pm$ 0.40 & 6.37 $\pm$ 0.31 \\

\midrule

Profile-to-PEFT
& 1  & 0.4015 $\pm$ 0.0372 & 0.3440 $\pm$ 0.0370 & 0.4022 $\pm$ 0.0397 & 5.64 $\pm$ 0.40 & 6.02 $\pm$ 0.29 \\
& 2  & 0.4018 $\pm$ 0.0404 & 0.3482 $\pm$ 0.0399 & 0.4052 $\pm$ 0.0416 & 5.78 $\pm$ 0.40 & 6.18 $\pm$ 0.32 \\
& 4  & 0.4039 $\pm$ 0.0390 & 0.3481 $\pm$ 0.0400 & 0.4014 $\pm$ 0.0407 & 5.75 $\pm$ 0.40 & 6.12 $\pm$ 0.31 \\
& 8  & 0.3975 $\pm$ 0.0380 & 0.3455 $\pm$ 0.0387 & 0.4078 $\pm$ 0.0405 & 6.17 $\pm$ 0.44 & 6.48 $\pm$ 0.32 \\
& 16 & 0.4338 $\pm$ 0.0426 & 0.3804 $\pm$ 0.0434 & 0.4316 $\pm$ 0.0439 & 6.48 $\pm$ 0.39 & 6.71 $\pm$ 0.33 \\

\midrule

\textbf{LatentPersonal}
& 1  & 0.4162 $\pm$ 0.0408 & 0.3637 $\pm$ 0.0400 & 0.4133 $\pm$ 0.0404 & 5.95 $\pm$ 0.41 & 6.33 $\pm$ 0.31 \\
& 2  & 0.4160 $\pm$ 0.0415 & 0.3656 $\pm$ 0.0404 & 0.4142 $\pm$ 0.0412 & 5.92 $\pm$ 0.40å & 6.28 $\pm$ 0.30 \\
& 4  & 0.4156 $\pm$ 0.0412 & 0.3639 $\pm$ 0.0403 & 0.4137 $\pm$ 0.0412 & 5.94 $\pm$ 0.43 & 6.27 $\pm$ 0.33 \\
& 8  & 0.4162 $\pm$ 0.0412 & 0.3652 $\pm$ 0.0405 & 0.4135 $\pm$ 0.0408 & 6.24 $\pm$ 0.43 & 6.61 $\pm$ 0.32 \\
& 16 & 0.4323 $\pm$ 0.0428 & 0.3851 $\pm$ 0.0428 & 0.4280 $\pm$ 0.0424 & 6.44 $\pm$ 0.43 & 6.72 $\pm$ 0.36 \\

\midrule

\multicolumn{7}{c}{\textbf{Qwen2.5-7B-Instruct}} \\
\midrule

Base
& / & 0.1712 $\pm$ 0.0124 & 0.1072 $\pm$ 0.0072 & 0.2171 $\pm$ 0.0127 & 3.84 $\pm$ 0.31 & 4.90 $\pm$ 0.23 \\

\midrule

ICL
& 1  & 0.1696 $\pm$ 0.0147 & 0.1086 $\pm$ 0.0091 & 0.2347 $\pm$ 0.0160 & 3.77 $\pm$ 0.31 & 4.69 $\pm$ 0.22 \\
& 2  & 0.1713 $\pm$ 0.0149 & 0.1074 $\pm$ 0.0086 & 0.2275 $\pm$ 0.0151 & 4.29 $\pm$ 0.32 & 5.06 $\pm$ 0.23 \\
& 4  & 0.1761 $\pm$ 0.0150 & 0.1175 $\pm$ 0.0109 & 0.2440 $\pm$ 0.0167 & 4.68 $\pm$ 0.38 & 5.24 $\pm$ 0.28 \\
& 8  & 0.1977 $\pm$ 0.0195 & 0.1382 $\pm$ 0.0161 & 0.2772 $\pm$ 0.0223 & 5.36 $\pm$ 0.44 & 5.58 $\pm$ 0.34 \\
& 16 & 0.2185 $\pm$ 0.0202 & 0.1675 $\pm$ 0.0193 & 0.3179 $\pm$ 0.0268 & 6.35 $\pm$ 0.42 & 6.16 $\pm$ 0.37 \\

\midrule

LoRA
& 1  & 0.3762 $\pm$ 0.0422 & 0.3203 $\pm$ 0.0414 & 0.3721 $\pm$ 0.0410 & 5.82 $\pm$ 0.43 & 6.16 $\pm$ 0.33 \\
& 2  & 0.3852 $\pm$ 0.0409 & 0.3321 $\pm$ 0.0413 & 0.3796 $\pm$ 0.0391 & 5.99 $\pm$ 0.45 & 6.34 $\pm$ 0.33 \\
& 4  & 0.3904 $\pm$ 0.0376 & 0.3376 $\pm$ 0.0375 & 0.3880 $\pm$ 0.0370 & 6.06 $\pm$ 0.46 & 6.39 $\pm$ 0.37 \\
& 8  & 0.3838 $\pm$ 0.0415 & 0.3344 $\pm$ 0.0404 & 0.3799 $\pm$ 0.0409 & 6.33 $\pm$ 0.45 & 6.54 $\pm$ 0.37 \\
& 16 & 0.4169 $\pm$ 0.0383 & 0.3607 $\pm$ 0.0395 & 0.4058 $\pm$ 0.0393 & 6.16 $\pm$ 0.51 & 6.50 $\pm$ 0.39 \\

\midrule

PPlug
& 1  & 0.2226 $\pm$ 0.0256 & 0.1638 $\pm$ 0.0191 & 0.2020 $\pm$ 0.0247 & 4.46 $\pm$ 0.36 & 5.01 $\pm$ 0.27 \\
& 2  & 0.2478 $\pm$ 0.0255 & 0.1852 $\pm$ 0.0199 & 0.2131 $\pm$ 0.0228 & 4.53 $\pm$ 0.38 & 4.83 $\pm$ 0.31 \\
& 4  & 0.2594 $\pm$ 0.0221 & 0.1953 $\pm$ 0.0181 & 0.2260 $\pm$ 0.0216 & 4.79 $\pm$ 0.38 & 5.01 $\pm$ 0.33 \\
& 8  & 0.2583 $\pm$ 0.0233 & 0.1989 $\pm$ 0.0196 & 0.2329 $\pm$ 0.0223 & 5.50 $\pm$ 0.41 & 5.64 $\pm$ 0.30 \\
& 16 & 0.2878 $\pm$ 0.0271 & 0.2177 $\pm$ 0.0233 & 0.2626 $\pm$ 0.0282 & 5.69 $\pm$ 0.39 & 5.71 $\pm$ 0.30 \\
\midrule

Profile-to-PEFT
& 1  & 0.3812 $\pm$ 0.0408 & 0.3320 $\pm$ 0.0417 & 0.3791 $\pm$ 0.0405 & 5.80 $\pm$ 0.40 & 6.17 $\pm$ 0.27 \\
& 2  & 0.3841 $\pm$ 0.0418 & 0.3349 $\pm$ 0.0411 & 0.3825 $\pm$ 0.0406 & 5.94 $\pm$ 0.38 & 6.25 $\pm$ 0.27 \\
& 4  & 0.3971 $\pm$ 0.0420 & 0.3434 $\pm$ 0.0422 & 0.3934 $\pm$ 0.0413 & 6.03 $\pm$ 0.42 & 6.25 $\pm$ 0.30 \\
& 8  & 0.3942 $\pm$ 0.0401 & 0.3433 $\pm$ 0.0395 & 0.3903 $\pm$ 0.0390 & 6.22 $\pm$ 0.42 & 6.64 $\pm$ 0.35 \\
& 16 & 0.4116 $\pm$ 0.0391 & 0.3609 $\pm$ 0.0394 & 0.3968 $\pm$ 0.0394 & 6.35 $\pm$ 0.40 & 6.77 $\pm$ 0.32 \\

\midrule

\textbf{LatentPersonal}
& 1  & 0.3906 $\pm$ 0.0420 & 0.3425 $\pm$ 0.0415 & 0.3951 $\pm$ 0.0401 & 6.01 $\pm$ 0.42 & 6.44 $\pm$ 0.26 \\
& 2  & 0.3817 $\pm$ 0.0409 & 0.3431 $\pm$ 0.0396 & 0.3815 $\pm$ 0.0390 & 6.09 $\pm$ 0.43 & 6.46 $\pm$ 0.31 \\
& 4  & 0.3843 $\pm$ 0.0413 & 0.3482 $\pm$ 0.0406 & 0.3898 $\pm$ 0.0388 & 5.91 $\pm$ 0.45 & 6.31 $\pm$ 0.33 \\
& 8  & 0.3844 $\pm$ 0.0405 & 0.3478 $\pm$ 0.0397 & 0.3840 $\pm$ 0.0386 & 6.39 $\pm$ 0.43 & 6.84 $\pm$ 0.31 \\
& 16 & 0.4111 $\pm$ 0.0387 & 0.3562 $\pm$ 0.0400 & 0.3979 $\pm$ 0.0383 & 6.34 $\pm$ 0.43 & 6.72 $\pm$ 0.34 \\

\bottomrule
\end{tabular}

%% file: tab/prism_performance.tex
\captionof{table}{Full personalization results on PRISM under different few-shot settings.}
\label{tab:full_prism}

\vspace{2pt}

\scriptsize
\setlength{\tabcolsep}{2.8pt}
\renewcommand{\arraystretch}{1.05}

\begin{tabular}{llccccc}
\toprule

\textbf{Method}
& \textbf{Shot}
& \textbf{R-1}
& \textbf{R-L}
& \textbf{MET.}
& \textbf{CP}
& \textbf{PP} \\

\midrule

\multicolumn{7}{c}{\textbf{Llama-3.1-8B-Instruct}} \\
\midrule

Base
& / & 0.2558 $\pm$ 0.0229 & 0.1524 $\pm$ 0.0137 & 0.2125 $\pm$ 0.0176 & 4.18 $\pm$ 0.42 & 4.08 $\pm$ 0.36\\

\midrule

ICL
& 1  & 0.2929 $\pm$ 0.0209 & 0.1703 $\pm$ 0.0149 & 0.2372 $\pm$ 0.0199 & 4.11 $\pm$ 0.34 & 4.73 $\pm$ 0.24 \\
& 2  & 0.2917 $\pm$ 0.0204 & 0.1637 $\pm$ 0.0156 & 0.2233 $\pm$ 0.0194 & 4.71 $\pm$ 0.33 & 5.07 $\pm$ 0.24 \\
& 4  & 0.2998 $\pm$ 0.0201 & 0.1706 $\pm$ 0.0146 & 0.2263 $\pm$ 0.0189 & 4.82 $\pm$ 0.33 & 4.98 $\pm$ 0.26 \\
& 8  & 0.3009 $\pm$ 0.0200 & 0.1719 $\pm$ 0.0153 & 0.2276 $\pm$ 0.0198 & 5.84 $\pm$ 0.35 & 5.65 $\pm$ 0.27 \\
& 16 & 0.3146 $\pm$ 0.0204 & 0.1916 $\pm$ 0.0150 & 0.2528 $\pm$ 0.0205 & 6.56 $\pm$ 0.36 & 6.41 $\pm$ 0.27 \\

\midrule

LoRA
& 1  & 0.2843 $\pm$ 0.0253 & 0.1762 $\pm$ 0.0159 & 0.2046 $\pm$ 0.0213 & 4.05 $\pm$ 0.43 & 4.95 $\pm$ 0.36 \\
& 2  & 0.2649 $\pm$ 0.0236 & 0.1672 $\pm$ 0.0151 & 0.1940 $\pm$ 0.0218 & 4.63 $\pm$ 0.47 & 5.10 $\pm$ 0.38 \\
& 4  & 0.2568 $\pm$ 0.0245 & 0.1653 $\pm$ 0.0153 & 0.1873 $\pm$ 0.0208 & 4.97 $\pm$ 0.45 & 5.24 $\pm$ 0.37 \\
& 8  & 0.2638 $\pm$ 0.0243 & 0.1712 $\pm$ 0.0158 & 0.1926 $\pm$ 0.0221 & 5.23 $\pm$ 0.46 & 5.56 $\pm$ 0.36 \\
& 16 & 0.2782 $\pm$ 0.0263 & 0.1832 $\pm$ 0.0192 & 0.2015 $\pm$ 0.0213 & 5.75 $\pm$ 0.51 & 5.95 $\pm$ 0.43 \\

\midrule

PPlug
& 1  & 0.2875 $\pm$ 0.0253 & 0.1830 $\pm$ 0.0167 & 0.2033 $\pm$ 0.0212 & 4.11 $\pm$ 0.43 & 4.76 $\pm$ 0.36 \\
& 2  & 0.2806 $\pm$ 0.0235 & 0.1761 $\pm$ 0.0154 & 0.1951 $\pm$ 0.0205 & 4.55 $\pm$ 0.44 & 4.87 $\pm$ 0.38 \\
& 4  & 0.2853 $\pm$ 0.0235 & 0.1822 $\pm$ 0.0163 & 0.1950 $\pm$ 0.0203 & 4.97 $\pm$ 0.42 & 5.26 $\pm$ 0.37 \\
& 8  & 0.2889 $\pm$ 0.0225 & 0.1835 $\pm$ 0.0165 & 0.1945 $\pm$ 0.0198 & 5.54 $\pm$ 0.43 & 5.69 $\pm$ 0.37 \\
& 16 & 0.2730 $\pm$ 0.0205 & 0.1736 $\pm$ 0.0130 & 0.1851 $\pm$ 0.0170 & 5.65 $\pm$ 0.47 & 5.75 $\pm$ 0.38 \\

\midrule

Profile-to-PEFT
& 1  & 0.2895 $\pm$ 0.0214 & 0.1794 $\pm$ 0.0144 & 0.2200 $\pm$ 0.0192 & 4.15 $\pm$ 0.42 & 5.04 $\pm$ 0.32 \\
& 2  & 0.2888 $\pm$ 0.0223 & 0.1786 $\pm$ 0.0151 & 0.2141 $\pm$ 0.0190 & 4.42 $\pm$ 0.47 & 5.13 $\pm$ 0.38 \\
& 4  & 0.2949 $\pm$ 0.0226 & 0.1839 $\pm$ 0.0146 & 0.2221 $\pm$ 0.0200 & 5.28 $\pm$ 0.44 & 5.73 $\pm$ 0.36 \\
& 8  & 0.2928 $\pm$ 0.0232 & 0.1807 $\pm$ 0.0153 & 0.2187 $\pm$ 0.0206 & 5.68 $\pm$ 0.48 & 5.98 $\pm$ 0.39 \\
& 16 & 0.2941 $\pm$ 0.0226 & 0.1911 $\pm$ 0.0158 & 0.2145 $\pm$ 0.0167 & 6.19 $\pm$ 0.47 & 6.50 $\pm$ 0.40 \\

\midrule

\textbf{LatentPersonal}
& 1  & 0.3081 $\pm$ 0.0231 & 0.2026 $\pm$ 0.0161 & 0.2421 $\pm$ 0.0191 & 4.28 $\pm$ 0.43 & 5.18 $\pm$ 0.37 \\
& 2  & 0.3036 $\pm$ 0.0229 & 0.1996 $\pm$ 0.0155 & 0.2380 $\pm$ 0.0190 & 4.82 $\pm$ 0.44 & 5.46 $\pm$ 0.34 \\
& 4  & 0.3043 $\pm$ 0.0231 & 0.2009 $\pm$ 0.0155 & 0.2385 $\pm$ 0.0177 & 5.30 $\pm$ 0.46 & 5.75 $\pm$ 0.36 \\
& 8  & 0.3049 $\pm$ 0.0234 & 0.2021 $\pm$ 0.0155 & 0.2391 $\pm$ 0.0182 & 5.79 $\pm$ 0.48 & 6.06 $\pm$ 0.36 \\
& 16 & 0.3181 $\pm$ 0.0207 & 0.2048 $\pm$ 0.0164 & 0.2469 $\pm$ 0.0182 & 6.12 $\pm$ 0.47 & 6.43 $\pm$ 0.39 \\

\midrule

\multicolumn{7}{c}{\textbf{Qwen2.5-7B-Instruct}} \\
\midrule

Base
& / & 0.2713 $\pm$ 0.0219 & 0.1571 $\pm$ 0.0135 & 0.2236 $\pm$ 0.0186 & 4.16 $\pm$ 0.41 & 4.37 $\pm$ 0.36 \\

\midrule

ICL
& 1  & 0.2893 $\pm$ 0.0195 & 0.1695 $\pm$ 0.0132 & 0.1994 $\pm$ 0.0181 & 4.16 $\pm$ 0.43 & 4.64 $\pm$ 0.32 \\
& 2  & 0.2899 $\pm$ 0.0196 & 0.1732 $\pm$ 0.0118 & 0.1943 $\pm$ 0.0170 & 4.60 $\pm$ 0.43 & 4.93 $\pm$ 0.32 \\
& 4  & 0.2974 $\pm$ 0.0196 & 0.1701 $\pm$ 0.0118 & 0.2041 $\pm$ 0.0160 & 4.98 $\pm$ 0.40 & 4.95 $\pm$ 0.30 \\
& 8  & 0.3051 $\pm$ 0.0193 & 0.1751 $\pm$ 0.0122 & 0.2051 $\pm$ 0.0157 & 5.78 $\pm$ 0.38 & 5.59 $\pm$ 0.30 \\
& 16 & 0.3087 $\pm$ 0.0188 & 0.1763 $\pm$ 0.0129 & 0.2130 $\pm$ 0.0165 & 6.25 $\pm$ 0.38 & 6.08 $\pm$ 0.31 \\

\midrule

LoRA
& 1  & 0.2488 $\pm$ 0.0218 & 0.1557 $\pm$ 0.0145 & 0.1841 $\pm$ 0.0182 & 3.57 $\pm$ 0.38 & 4.29 $\pm$ 0.34 \\
& 2  & 0.2486 $\pm$ 0.0236 & 0.1517 $\pm$ 0.0150 & 0.1776 $\pm$ 0.0195 & 3.85 $\pm$ 0.43 & 4.46 $\pm$ 0.38 \\
& 4  & 0.2351 $\pm$ 0.0249 & 0.1540 $\pm$ 0.0176 & 0.1712 $\pm$ 0.0208 & 3.83 $\pm$ 0.45 & 4.32 $\pm$ 0.41 \\
& 8  & 0.2454 $\pm$ 0.0240 & 0.1511 $\pm$ 0.0146 & 0.1738 $\pm$ 0.0204 & 4.56 $\pm$ 0.47 & 4.97 $\pm$ 0.38 \\
& 16 & 0.2420 $\pm$ 0.0259 & 0.1553 $\pm$ 0.0167 & 0.1743 $\pm$ 0.0205 & 4.79 $\pm$ 0.53 & 5.12 $\pm$ 0.48 \\

\midrule

PPlug
& 1  & 0.3044 $\pm$ 0.0208 & 0.1812 $\pm$ 0.0144 & 0.2141 $\pm$ 0.0176 & 4.14 $\pm$ 0.43 & 4.65 $\pm$ 0.38 \\
& 2  & 0.2979 $\pm$ 0.0201 & 0.1797 $\pm$ 0.0139 & 0.2084 $\pm$ 0.0180 & 4.45 $\pm$ 0.44 & 4.69 $\pm$ 0.36 \\
& 4  & 0.2908 $\pm$ 0.0201 & 0.1751 $\pm$ 0.0144 & 0.1992 $\pm$ 0.0162 & 4.75 $\pm$ 0.44 & 5.01 $\pm$ 0.35 \\
& 8  & 0.2912 $\pm$ 0.0213 & 0.1760 $\pm$ 0.0151 & 0.1962 $\pm$ 0.0172 & 5.45 $\pm$ 0.46 & 5.56 $\pm$ 0.38 \\
& 16 & 0.2970 $\pm$ 0.0209 & 0.1865 $\pm$ 0.0150 & 0.2082 $\pm$ 0.0186 & 5.57 $\pm$ 0.48 & 5.68 $\pm$ 0.38 \\
\midrule

Profile-to-PEFT
& 1  & 0.2891 $\pm$ 0.0228 & 0.1719 $\pm$ 0.0138 & 0.2223 $\pm$ 0.0193 & 4.25 $\pm$ 0.42 & 5.05 $\pm$ 0.36 \\
& 2  & 0.2925 $\pm$ 0.0244 & 0.1760 $\pm$ 0.0150 & 0.2163 $\pm$ 0.0193 & 4.69 $\pm$ 0.45 & 5.19 $\pm$ 0.37 \\
& 4  & 0.2853 $\pm$ 0.0223 & 0.1724 $\pm$ 0.0129 & 0.2126 $\pm$ 0.0187 & 5.14 $\pm$ 0.44 & 5.53 $\pm$ 0.36 \\
& 8  & 0.2809 $\pm$ 0.0208 & 0.1716 $\pm$ 0.0129 & 0.2050 $\pm$ 0.0177 & 5.57 $\pm$ 0.45 & 5.85 $\pm$ 0.34 \\
& 16 & 0.2907 $\pm$ 0.0192 & 0.1825 $\pm$ 0.0130 & 0.2192 $\pm$ 0.0166 & 5.90 $\pm$ 0.47 & 6.21 $\pm$ 0.38 \\

\midrule

\textbf{LatentPersonal}
& 1  & 0.3008 $\pm$ 0.0229 & 0.1801 $\pm$ 0.0134 & 0.2251 $\pm$ 0.0194 & 4.42 $\pm$ 0.41 & 5.27 $\pm$ 0.34 \\
& 2  & 0.2946 $\pm$ 0.0216 & 0.1794 $\pm$ 0.0142 & 0.2209 $\pm$ 0.0192 & 4.79 $\pm$ 0.43 & 5.28 $\pm$ 0.33 \\
& 4  & 0.3020 $\pm$ 0.0227 & 0.1837 $\pm$ 0.0138 & 0.2255 $\pm$ 0.0193 & 5.33 $\pm$ 0.44 & 5.76 $\pm$ 0.34 \\
& 8  & 0.2977 $\pm$ 0.0223 & 0.1800 $\pm$ 0.0134 & 0.2247 $\pm$ 0.0188 & 5.59 $\pm$ 0.48 & 5.86 $\pm$ 0.36 \\
& 16 & 0.2994 $\pm$ 0.0204 & 0.1869 $\pm$ 0.0144 & 0.2363 $\pm$ 0.0185 & 6.31 $\pm$ 0.46 & 6.36 $\pm$ 0.36 \\

\bottomrule
\end{tabular}

%% file: tab/personamem_performance.tex
\captionof{table}{Full personalization results on PersonaMem under different few-shot settings.}
\label{tab:full_personamem}

\vspace{2pt}

\scriptsize
\setlength{\tabcolsep}{2.8pt}
\renewcommand{\arraystretch}{1.05}

\begin{tabular}{llccccc}
\toprule

\textbf{Method}
& \textbf{Shot}
& \textbf{R-1}
& \textbf{R-L}
& \textbf{MET.}
& \textbf{CP}
& \textbf{PP} \\

\midrule

\multicolumn{7}{c}{\textbf{Llama-3.1-8B-Instruct}} \\
\midrule

Base
& / & 0.2368 $\pm$ 0.0099 & 0.1336 $\pm$ 0.0061 & 0.2103 $\pm$ 0.0101 & 5.01 $\pm$ 0.44 & 5.15 $\pm$ 0.30 \\

\midrule

ICL
& 1  & 0.2514 $\pm$ 0.0136 & 0.1421 $\pm$ 0.0096 & 0.2352 $\pm$ 0.0130 & 4.87 $\pm$ 0.34 & 5.21 $\pm$ 0.28 \\
& 2  & 0.2682 $\pm$ 0.0131 & 0.1522 $\pm$ 0.0089 & 0.2457 $\pm$ 0.0121 & 5.51 $\pm$ 0.33 & 5.65 $\pm$ 0.27 \\
& 4  & 0.2618 $\pm$ 0.0134 & 0.1512 $\pm$ 0.0091 & 0.2457 $\pm$ 0.0124 & 6.07 $\pm$ 0.35 & 6.18 $\pm$ 0.23 \\
& 8  & 0.2759 $\pm$ 0.0132 & 0.1613 $\pm$ 0.0088 & 0.2618 $\pm$ 0.0122 & 6.39 $\pm$ 0.32 & 6.49 $\pm$ 0.25 \\
& 16 & 0.2738 $\pm$ 0.0127 & 0.1659 $\pm$ 0.0084 & 0.2634 $\pm$ 0.0119 & 7.32 $\pm$ 0.32 & 7.29 $\pm$ 0.25 \\

\midrule

LoRA
& 1  & 0.3448 $\pm$ 0.0193 & 0.2311 $\pm$ 0.0174 & 0.2767 $\pm$ 0.0184 & 5.97 $\pm$ 0.38 & 6.16 $\pm$ 0.28 \\
& 2  & 0.3442 $\pm$ 0.0200 & 0.2286 $\pm$ 0.0156 & 0.2740 $\pm$ 0.0202 & 6.09 $\pm$ 0.38 & 6.32 $\pm$ 0.27 \\
& 4  & 0.3447 $\pm$ 0.0188 & 0.2253 $\pm$ 0.0150 & 0.2800 $\pm$ 0.0186 & 6.25 $\pm$ 0.39 & 6.52 $\pm$ 0.26 \\
& 8  & 0.3446 $\pm$ 0.0204 & 0.2302 $\pm$ 0.0176 & 0.2782 $\pm$ 0.0198 & 6.40 $\pm$ 0.36 & 6.69 $\pm$ 0.28 \\
& 16 & 0.3627 $\pm$ 0.0176 & 0.2367 $\pm$ 0.0136 & 0.2955 $\pm$ 0.0188 & 7.66 $\pm$ 0.32 & 7.16 $\pm$ 0.26 \\

\midrule

PPlug
& 1  & 0.2949 $\pm$ 0.0155 & 0.1860 $\pm$ 0.0114 & 0.2265 $\pm$ 0.0139 & 5.17 $\pm$ 0.36 & 5.47 $\pm$ 0.26 \\
& 2  & 0.2902 $\pm$ 0.0139 & 0.1837 $\pm$ 0.0101 & 0.2216 $\pm$ 0.0136 & 5.19 $\pm$ 0.34 & 5.45 $\pm$ 0.29 \\
& 4  & 0.2939 $\pm$ 0.0152 & 0.1860 $\pm$ 0.0117 & 0.2238 $\pm$ 0.0152 & 5.53 $\pm$ 0.36 & 5.76 $\pm$ 0.28 \\
& 8  & 0.2953 $\pm$ 0.0141 & 0.1909 $\pm$ 0.0109 & 0.2226 $\pm$ 0.0140 & 5.88 $\pm$ 0.32 & 5.97 $\pm$ 0.24 \\
& 16 & 0.2892 $\pm$ 0.0139 & 0.1818 $\pm$ 0.0096 & 0.2148 $\pm$ 0.0120 & 6.39 $\pm$ 0.28 & 6.37 $\pm$ 0.24 \\

\midrule

Profile-to-PEFT
& 1  & 0.3398 $\pm$ 0.0189 & 0.2269 $\pm$ 0.0162 & 0.2749 $\pm$ 0.0178 & 5.74 $\pm$ 0.36 & 6.32 $\pm$ 0.25 \\
& 2  & 0.3480 $\pm$ 0.0184 & 0.2286 $\pm$ 0.0160 & 0.2806 $\pm$ 0.0195 & 5.70 $\pm$ 0.37 & 6.51 $\pm$ 0.26 \\
& 4  & 0.3523 $\pm$ 0.0182 & 0.2325 $\pm$ 0.0157 & 0.2827 $\pm$ 0.0186 & 5.51 $\pm$ 0.35 & 6.69 $\pm$ 0.26 \\
& 8  & 0.3561 $\pm$ 0.0189 & 0.2390 $\pm$ 0.0169 & 0.2859 $\pm$ 0.0197 & 6.37 $\pm$ 0.33 & 6.42 $\pm$ 0.24 \\
& 16 & 0.3616 $\pm$ 0.0184 & 0.2397 $\pm$ 0.0151 & 0.2880 $\pm$ 0.0199 & 6.21 $\pm$ 0.30 & 7.55 $\pm$ 0.22 \\

\midrule

\textbf{LatentPersonal}
& 1  & 0.3534 $\pm$ 0.0181 & 0.2336 $\pm$ 0.0161 & 0.2803 $\pm$ 0.0200 & 6.14 $\pm$ 0.37 & 6.52 $\pm$ 0.25 \\
& 2  & 0.3631 $\pm$ 0.0184 & 0.2536 $\pm$ 0.0164 & 0.2930 $\pm$ 0.0202 & 6.27 $\pm$ 0.37 & 6.67 $\pm$ 0.27 \\
& 4  & 0.3647 $\pm$ 0.0178 & 0.2469 $\pm$ 0.0157 & 0.2996 $\pm$ 0.0208 & 6.35 $\pm$ 0.37 & 6.59 $\pm$ 0.27 \\
& 8  & 0.3718 $\pm$ 0.0180 & 0.2542 $\pm$ 0.0156 & 0.3031 $\pm$ 0.0201 & 6.42 $\pm$ 0.38 & 6.48 $\pm$ 0.29 \\
& 16 & 0.3646 $\pm$ 0.0180 & 0.2597 $\pm$ 0.0147 & 0.3196 $\pm$ 0.0193 & 6.64 $\pm$ 0.28 & 6.78 $\pm$ 0.25 \\

\midrule

\multicolumn{7}{c}{\textbf{Qwen2.5-7B-Instruct}} \\
\midrule

Base
& / & 0.2598 $\pm$ 0.0129 & 0.1405 $\pm$ 0.0070 & 0.2303 $\pm$ 0.0125 & 5.64 $\pm$ 0.44 & 5.69 $\pm$ 0.29 \\

\midrule

ICL
& 1  & 0.2245 $\pm$ 0.0121 & 0.1152 $\pm$ 0.0082 & 0.2043 $\pm$ 0.0122 & 4.93 $\pm$ 0.38 & 5.26 $\pm$ 0.25 \\
& 2  & 0.2376 $\pm$ 0.0141 & 0.1235 $\pm$ 0.0100 & 0.2192 $\pm$ 0.0127 & 5.46 $\pm$ 0.36 & 5.61 $\pm$ 0.23 \\
& 4  & 0.2484 $\pm$ 0.0123 & 0.1334 $\pm$ 0.0084 & 0.2294 $\pm$ 0.0111 & 6.07 $\pm$ 0.35 & 6.07 $\pm$ 0.24 \\
& 8  & 0.2440 $\pm$ 0.0138 & 0.1397 $\pm$ 0.0101 & 0.2284 $\pm$ 0.0142 & 6.50 $\pm$ 0.33 & 6.47 $\pm$ 0.26 \\
& 16 & 0.2517 $\pm$ 0.0125 & 0.1596 $\pm$ 0.0081 & 0.2401 $\pm$ 0.0122 & 7.24 $\pm$ 0.31 & 7.13 $\pm$ 0.26 \\

\midrule

LoRA
& 1  & 0.3166 $\pm$ 0.0199 & 0.2114 $\pm$ 0.0160 & 0.2611 $\pm$ 0.0205 & 6.03 $\pm$ 0.40 & 6.20 $\pm$ 0.29 \\
& 2  & 0.3276 $\pm$ 0.0195 & 0.2154 $\pm$ 0.0163 & 0.2652 $\pm$ 0.0194 & 6.19 $\pm$ 0.41 & 6.31 $\pm$ 0.29 \\
& 4  & 0.3399 $\pm$ 0.0198 & 0.2271 $\pm$ 0.0167 & 0.2747 $\pm$ 0.0212 & 6.47 $\pm$ 0.38 & 6.72 $\pm$ 0.28 \\
& 8  & 0.3417 $\pm$ 0.0169 & 0.2250 $\pm$ 0.0147 & 0.2740 $\pm$ 0.0187 & 6.65 $\pm$ 0.40 & 6.73 $\pm$ 0.31 \\
& 16 & 0.3555 $\pm$ 0.0185 & 0.2349 $\pm$ 0.0150 & 0.2834 $\pm$ 0.0202 & 7.32 $\pm$ 0.32 & 7.37 $\pm$ 0.26 \\

\midrule

PPlug
& 1  & 0.2947 $\pm$ 0.0167 & 0.1800 $\pm$ 0.0127 & 0.2332 $\pm$ 0.0171 & 5.16 $\pm$ 0.37 & 5.45 $\pm$ 0.29 \\
& 2  & 0.2950 $\pm$ 0.0167 & 0.1791 $\pm$ 0.0125 & 0.2282 $\pm$ 0.0170 & 5.32 $\pm$ 0.36 & 5.54 $\pm$ 0.29 \\
& 4  & 0.2984 $\pm$ 0.0162 & 0.1768 $\pm$ 0.0118 & 0.2295 $\pm$ 0.0165 & 5.46 $\pm$ 0.36 & 5.74 $\pm$ 0.29 \\
& 8  & 0.3039 $\pm$ 0.0167 & 0.1824 $\pm$ 0.0123 & 0.2277 $\pm$ 0.0163 & 5.68 $\pm$ 0.34 & 5.89 $\pm$ 0.27 \\
& 16 & 0.2990 $\pm$ 0.0150 & 0.1810 $\pm$ 0.0099 & 0.2300 $\pm$ 0.0138 & 6.32 $\pm$ 0.29 & 6.21 $\pm$ 0.28 \\

\midrule

Profile-to-PEFT
& 1  & 0.3230 $\pm$ 0.0197 & 0.2159 $\pm$ 0.0169 & 0.2594 $\pm$ 0.0195 & 6.14 $\pm$ 0.40 & 6.26 $\pm$ 0.29 \\
& 2  & 0.3253 $\pm$ 0.0201 & 0.2172 $\pm$ 0.0163 & 0.2538 $\pm$ 0.0193 & 6.31 $\pm$ 0.37 & 6.40 $\pm$ 0.26 \\
& 4  & 0.3190 $\pm$ 0.0205 & 0.2129 $\pm$ 0.0161 & 0.2592 $\pm$ 0.0193 & 6.60 $\pm$ 0.37 & 6.75 $\pm$ 0.26 \\
& 8  & 0.3223 $\pm$ 0.0204 & 0.2164 $\pm$ 0.0163 & 0.2639 $\pm$ 0.0189 & 6.72 $\pm$ 0.35 & 6.90 $\pm$ 0.25 \\
& 16 & 0.3469 $\pm$ 0.0172 & 0.2288 $\pm$ 0.0136 & 0.2738 $\pm$ 0.0186 & 7.33 $\pm$ 0.30 & 7.43 $\pm$ 0.24 \\
\midrule

\textbf{LatentPersonal}
& 1  & 0.3447 $\pm$ 0.0198 & 0.2268 $\pm$ 0.0162 & 0.2800 $\pm$ 0.0201 & 6.41 $\pm$ 0.37 & 6.39 $\pm$ 0.28 \\
& 2  & 0.3455 $\pm$ 0.0197 & 0.2289 $\pm$ 0.0178 & 0.2781 $\pm$ 0.0210 & 6.38 $\pm$ 0.36 & 6.48 $\pm$ 0.29 \\
& 4  & 0.3486 $\pm$ 0.0192 & 0.2259 $\pm$ 0.0171 & 0.2758 $\pm$ 0.0196 & 6.74 $\pm$ 0.34 & 6.70 $\pm$ 0.25 \\
& 8  & 0.3416 $\pm$ 0.0193 & 0.2240 $\pm$ 0.0164 & 0.2742 $\pm$ 0.0196 & 6.97 $\pm$ 0.36 & 6.96 $\pm$ 0.29 \\
& 16 & 0.3518 $\pm$ 0.0181 & 0.2278 $\pm$ 0.0128 & 0.2796 $\pm$ 0.0180 & 7.49 $\pm$ 0.33 & 7.47 $\pm$ 0.25 \\

\bottomrule
\end{tabular}

%% file: tab/pir_performance.tex
\captionof{table}{Full personalization results on PIR under different few-shot settings.}
\label{tab:full_pir}

\vspace{2pt}

\scriptsize
\setlength{\tabcolsep}{2.8pt}
\renewcommand{\arraystretch}{1.05}

\begin{tabular}{llccccc}
\toprule

\textbf{Method}
& \textbf{Shot}
& \textbf{R-1}
& \textbf{R-L}
& \textbf{MET.}
& \textbf{CP}
& \textbf{PP} \\
\midrule

\multicolumn{7}{c}{\textbf{Llama-3.1-8B-Instruct}} \\
\midrule

Base
& / & 0.2954 $\pm$ 0.0142 & 0.1568 $\pm$ 0.0079 & 0.1966 $\pm$ 0.0087 & 4.08 $\pm$ 0.32	 & 4.12 $\pm$ 0.25 \\

\midrule

ICL
& 1  & 0.3028 $\pm$ 0.0116 & 0.1643 $\pm$ 0.0067 & 0.2066 $\pm$ 0.0085 & 4.19 $\pm$ 0.30 & 4.90 $\pm$ 0.24 \\
& 2  & 0.3090 $\pm$ 0.0114 & 0.1691 $\pm$ 0.0065 & 0.2114 $\pm$ 0.0089 & 5.16 $\pm$ 0.31 & 5.80 $\pm$ 0.24 \\
& 4  & 0.3125 $\pm$ 0.0113 & 0.1707 $\pm$ 0.0065 & 0.2129 $\pm$ 0.0083 & 5.89 $\pm$ 0.33 & 6.24 $\pm$ 0.26 \\
& 8  & 0.3082 $\pm$ 0.0116 & 0.1677 $\pm$ 0.0066 & 0.2082 $\pm$ 0.0080 & 6.51 $\pm$ 0.32 & 6.71 $\pm$ 0.26 \\

\midrule

LoRA
& 1  & 0.3805 $\pm$ 0.0157 & 0.2070 $\pm$ 0.0113 & 0.2962 $\pm$ 0.0140 & 4.28 $\pm$ 0.35 & 5.35 $\pm$ 0.29 \\
& 2  & 0.3795 $\pm$ 0.0152 & 0.2079 $\pm$ 0.0104 & 0.2931 $\pm$ 0.0138 & 5.25 $\pm$ 0.33 & 6.00 $\pm$ 0.28 \\
& 4  & 0.3836 $\pm$ 0.0148 & 0.2059 $\pm$ 0.0086 & 0.2983 $\pm$ 0.0128 & 5.72 $\pm$ 0.33 & 6.47 $\pm$ 0.26 \\
& 8  & 0.3846 $\pm$ 0.0141 & 0.2063 $\pm$ 0.0083 & 0.2977 $\pm$ 0.0126 & 6.67 $\pm$ 0.26 & 7.31 $\pm$ 0.20 \\

\midrule

PPlug
& 1  & 0.3711 $\pm$ 0.0126 & 0.1997 $\pm$ 0.0084 & 0.2495 $\pm$ 0.0107 & 3.96 $\pm$ 0.30 & 4.88 $\pm$ 0.25 \\
& 2  & 0.3633 $\pm$ 0.0128 & 0.1950 $\pm$ 0.0080 & 0.2401 $\pm$ 0.0096 & 5.16 $\pm$ 0.33 & 5.71 $\pm$ 0.26 \\
& 4  & 0.3645 $\pm$ 0.0128 & 0.1929 $\pm$ 0.0078 & 0.2436 $\pm$ 0.0103 & 5.86 $\pm$ 0.32 & 6.02 $\pm$ 0.26 \\
& 8  & 0.3646 $\pm$ 0.0130 & 0.1943 $\pm$ 0.0078 & 0.2409 $\pm$ 0.0096 & 6.56 $\pm$ 0.28 & 6.79 $\pm$ 0.23 \\

\midrule

Profile-to-PEFT
& 1  & 0.3891 $\pm$ 0.0155 & 0.2099 $\pm$ 0.0103 & 0.3066 $\pm$ 0.0137 & 4.08 $\pm$ 0.33 & 5.45 $\pm$ 0.31 \\
& 2  & 0.3952 $\pm$ 0.0153 & 0.2161 $\pm$ 0.0110 & 0.3111 $\pm$ 0.0140 & 5.13 $\pm$ 0.31 & 6.15 $\pm$ 0.25 \\
& 4  & 0.3857 $\pm$ 0.0153 & 0.2168 $\pm$ 0.0121 & 0.3024 $\pm$ 0.0138 & 5.75 $\pm$ 0.33 & 6.77 $\pm$ 0.27 \\
& 8  & 0.3880 $\pm$ 0.0152 & 0.2144 $\pm$ 0.0112 & 0.3045 $\pm$ 0.0133 & 6.63 $\pm$ 0.32 & 7.35 $\pm$ 0.23 \\

\midrule

\textbf{LatentPersonal}
& 1  & 0.3932 $\pm$ 0.0146 & 0.2092 $\pm$ 0.0113 & 0.3087 $\pm$ 0.0136 & 4.38 $\pm$ 0.31 & 5.62 $\pm$ 0.27 \\
& 2  & 0.3926 $\pm$ 0.0144 & 0.2089 $\pm$ 0.0111 & 0.3094 $\pm$ 0.0128 & 5.30 $\pm$ 0.34 & 6.25 $\pm$ 0.27 \\
& 4  & 0.3912 $\pm$ 0.0150 & 0.2101 $\pm$ 0.0117 & 0.3084 $\pm$ 0.0139 & 5.90 $\pm$ 0.33 & 6.57 $\pm$ 0.27 \\
& 8  & 0.3852 $\pm$ 0.0149 & 0.2166 $\pm$ 0.0098 & 0.3163 $\pm$ 0.0128 & 6.58 $\pm$ 0.31 & 7.34 $\pm$ 0.24 \\

\midrule

\multicolumn{7}{c}{\textbf{Qwen2.5-7B-Instruct}} \\
\midrule

Base
& / & 0.2871 $\pm$ 0.0123 & 0.1581 $\pm$ 0.0069 & 0.1919 $\pm$ 0.0093 & 3.86 $\pm$ 0.31 & 4.28 $\pm$ 0.25 \\

\midrule

ICL
& 1  & 0.2981 $\pm$ 0.0129 & 0.1483 $\pm$ 0.0073 & 0.2012 $\pm$ 0.0093 & 4.08 $\pm$ 0.32 & 4.98 $\pm$ 0.28 \\
& 2  & 0.3014 $\pm$ 0.0135 & 0.1583 $\pm$ 0.0071 & 0.2194 $\pm$ 0.0099 & 5.06 $\pm$ 0.31 & 5.66 $\pm$ 0.24 \\
& 4  & 0.3123 $\pm$ 0.0143 & 0.1688 $\pm$ 0.0070 & 0.2123 $\pm$ 0.0101 & 5.81 $\pm$ 0.30 & 6.16 $\pm$ 0.25 \\
& 8  & 0.3103 $\pm$ 0.0139 & 0.1691 $\pm$ 0.0071 & 0.2217 $\pm$ 0.0101 & 6.50 $\pm$ 0.31 & 6.78 $\pm$ 0.29 \\

\midrule

LoRA
& 1  & 0.3756 $\pm$ 0.0163 & 0.1995 $\pm$ 0.0108 & 0.2902 $\pm$ 0.0150 & 4.17 $\pm$ 0.33 & 5.45 $\pm$ 0.28 \\
& 2  & 0.3827 $\pm$ 0.0158 & 0.2033 $\pm$ 0.0101 & 0.2945 $\pm$ 0.0143 & 5.13 $\pm$ 0.33 & 6.12 $\pm$ 0.28 \\
& 4  & 0.3868 $\pm$ 0.0167 & 0.2058 $\pm$ 0.0109 & 0.2954 $\pm$ 0.0157 & 6.12 $\pm$ 0.30 & 6.71 $\pm$ 0.27 \\
& 8  & 0.3806 $\pm$ 0.0157 & 0.2020 $\pm$ 0.0099 & 0.2933 $\pm$ 0.0140 & 6.51 $\pm$ 0.30 & 7.30 $\pm$ 0.23 \\

\midrule

PPlug
& 1  & 0.3404 $\pm$ 0.0100 & 0.1754 $\pm$ 0.0064 & 0.2272 $\pm$ 0.0082 & 4.00 $\pm$ 0.29 & 4.87 $\pm$ 0.23 \\
& 2  & 0.3351 $\pm$ 0.0105 & 0.1740 $\pm$ 0.0056 & 0.2255 $\pm$ 0.0086 & 4.88 $\pm$ 0.33 & 5.45 $\pm$ 0.27 \\
& 4  & 0.3397 $\pm$ 0.0113 & 0.1744 $\pm$ 0.0063 & 0.2288 $\pm$ 0.0096 & 5.79 $\pm$ 0.29 & 6.13 $\pm$ 0.25 \\
& 8  & 0.3483 $\pm$ 0.0107 & 0.1811 $\pm$ 0.0064 & 0.2324 $\pm$ 0.0089 & 6.35 $\pm$ 0.30 & 6.56 $\pm$ 0.26 \\

\midrule

Profile-to-PEFT
& 1  & 0.3958 $\pm$ 0.0160 & 0.2096 $\pm$ 0.0107 & 0.3073 $\pm$ 0.0151 & 4.18 $\pm$ 0.32 & 5.31 $\pm$ 0.26 \\
& 2  & 0.3891 $\pm$ 0.0142 & 0.2049 $\pm$ 0.0096 & 0.2993 $\pm$ 0.0121 & 5.20 $\pm$ 0.29 & 6.08 $\pm$ 0.22 \\
& 4  & 0.3871 $\pm$ 0.0152 & 0.2056 $\pm$ 0.0095 & 0.2988 $\pm$ 0.0134 & 5.85 $\pm$ 0.32 & 6.62 $\pm$ 0.27 \\
& 8  & 0.3903 $\pm$ 0.0159 & 0.2066 $\pm$ 0.0097 & 0.3015 $\pm$ 0.0131 & 6.47 $\pm$ 0.29 & 7.16 $\pm$ 0.22 \\

\midrule

\textbf{LatentPersonal}
& 1  & 0.3975 $\pm$ 0.0156 & 0.2074 $\pm$ 0.0097 & 0.3099 $\pm$ 0.0126 & 4.42 $\pm$ 0.32 & 5.75 $\pm$ 0.27 \\
& 2  & 0.3940 $\pm$ 0.0163 & 0.2100 $\pm$ 0.0100 & 0.3094 $\pm$ 0.0141 & 5.23 $\pm$ 0.32 & 6.20 $\pm$ 0.25 \\
& 4  & 0.3935 $\pm$ 0.0162 & 0.2099 $\pm$ 0.0100 & 0.3031 $\pm$ 0.0141 & 5.94 $\pm$ 0.33 & 6.73 $\pm$ 0.28 \\
& 8  & 0.3946 $\pm$ 0.0156 & 0.2079 $\pm$ 0.0096 & 0.2990 $\pm$ 0.0133 & 6.48 $\pm$ 0.29 & 7.48 $\pm$ 0.23 \\

\bottomrule
\end{tabular}

%% file: tab/ablation_ad.tex
\begin{table*}[!htbp]
\centering
\caption{
Full ablation results of \textit{LatentPersonal} on HiCUPID with
Llama-3.1-8B-Instruct across different numbers of observed user interactions.
}
\label{tab:ablation_full}

\small
\setlength{\tabcolsep}{2.8pt}
\renewcommand{\arraystretch}{1.08}

\begin{tabular}{lc*{5}{c}}
\toprule
\textbf{Variant}
& \textbf{Shot}
& \textbf{R-1}
& \textbf{R-L}
& \textbf{MET.}
& \textbf{CP}
& \textbf{PP} \\
\midrule

\multirow{5}{*}{\textit{LatentPersonal}}
& 1  & 0.4162 $\pm$ 0.0408 & 0.3637 $\pm$ 0.0400 & 0.4133 $\pm$ 0.0404 & 5.95 $\pm$ 0.41 & 6.33 $\pm$ 0.31 \\
& 2  & 0.4160 $\pm$ 0.0415 & 0.3656 $\pm$ 0.0404 & 0.4142 $\pm$ 0.0412 & 5.92 $\pm$ 0.40 & 6.28 $\pm$ 0.30 \\
& 4  & 0.4156 $\pm$ 0.0412 & 0.3639 $\pm$ 0.0403 & 0.4137 $\pm$ 0.0412 & 5.94 $\pm$ 0.43 & 6.27 $\pm$ 0.33 \\
& 8  & 0.4162 $\pm$ 0.0412 & 0.3652 $\pm$ 0.0405 & 0.4135 $\pm$ 0.0408 & 6.24 $\pm$ 0.43 & 6.61 $\pm$ 0.32 \\
& 16 & 0.4323 $\pm$ 0.0428 & 0.3851 $\pm$ 0.0428 & 0.4280 $\pm$ 0.0424 & 6.44 $\pm$ 0.43 & 6.72 $\pm$ 0.36 \\

\midrule

\multirow{5}{*}{w/o Latent}
& 1  & 0.3550 $\pm$ 0.0260 & 0.3011 $\pm$ 0.0226 & 0.3667 $\pm$ 0.0231 & 4.72 $\pm$ 0.40 & 5.85 $\pm$ 0.34 \\
& 2  & 0.3499 $\pm$ 0.0230 & 0.2925 $\pm$ 0.0199 & 0.3841 $\pm$ 0.0231 & 4.86 $\pm$ 0.38 & 5.92 $\pm$ 0.33 \\
& 4  & 0.3308 $\pm$ 0.0163 & 0.2757 $\pm$ 0.0112 & 0.3423 $\pm$ 0.0141 & 5.29 $\pm$ 0.36 & 6.16 $\pm$ 0.35 \\
& 8  & 0.3479 $\pm$ 0.0187 & 0.2742 $\pm$ 0.0142 & 0.3475 $\pm$ 0.0184 & 5.51 $\pm$ 0.35 & 6.21 $\pm$ 0.32 \\
& 16 & 0.3478 $\pm$ 0.0158 & 0.2807 $\pm$ 0.0112 & 0.3531 $\pm$ 0.0150 & 5.36 $\pm$ 0.36 & 6.02 $\pm$ 0.36 \\

\midrule

\multirow{5}{*}{w/o Navigation}
& 1  & 0.3812 $\pm$ 0.0418 & 0.3298 $\pm$ 0.0411 & 0.3777 $\pm$ 0.0404 & 5.66 $\pm$ 0.45 & 6.12 $\pm$ 0.27 \\
& 2  & 0.3910 $\pm$ 0.0423 & 0.3452 $\pm$ 0.0420 & 0.3929 $\pm$ 0.0402 & 5.71 $\pm$ 0.44 & 6.26 $\pm$ 0.31 \\
& 4  & 0.3909 $\pm$ 0.0420 & 0.3415 $\pm$ 0.0401 & 0.3992 $\pm$ 0.0407 & 5.92 $\pm$ 0.45 & 6.07 $\pm$ 0.36 \\
& 8  & 0.3838 $\pm$ 0.0389 & 0.3347 $\pm$ 0.0377 & 0.3870 $\pm$ 0.0402 & 6.12 $\pm$ 0.49 & 6.52 $\pm$ 0.36 \\
& 16 & 0.4149 $\pm$ 0.0410 & 0.3643 $\pm$ 0.0418 & 0.4112 $\pm$ 0.0415 & 6.49 $\pm$ 0.39 & 6.88 $\pm$ 0.34 \\

\midrule

\multirow{5}{*}{w/o IB}
& 1  & 0.2684 $\pm$ 0.0079 & 0.2510 $\pm$ 0.0051 & 0.3423 $\pm$ 0.0114 & 4.57 $\pm$ 0.31 & 4.72 $\pm$ 0.27 \\
& 2  & 0.2703 $\pm$ 0.0083 & 0.2511 $\pm$ 0.0052 & 0.3422 $\pm$ 0.0118 & 4.65 $\pm$ 0.30 & 4.64 $\pm$ 0.28 \\
& 4  & 0.2695 $\pm$ 0.0079 & 0.2515 $\pm$ 0.0050 & 0.3444 $\pm$ 0.0112 & 4.74 $\pm$ 0.33 & 4.55 $\pm$ 0.29 \\
& 8  & 0.2674 $\pm$ 0.0077 & 0.2502 $\pm$ 0.0049 & 0.3413 $\pm$ 0.0112 & 4.64 $\pm$ 0.30 & 4.33 $\pm$ 0.27 \\
& 16 & 0.2740 $\pm$ 0.0072 & 0.2541 $\pm$ 0.0043 & 0.3504 $\pm$ 0.0101 & 4.45 $\pm$ 0.30 & 4.12 $\pm$ 0.27 \\

\bottomrule
\end{tabular}
\end{table*}

%% file: tab/beta.tex
\begin{table}[!th]
\centering
\caption{
Sensitivity analysis of the information bottleneck coefficient $\beta$
on HiCUPID with Llama-3.1-8B-Instruct across different numbers of
observed user interactions.
}
\label{tab:beta_sensitivity}

\small
\setlength{\tabcolsep}{5pt}
\renewcommand{\arraystretch}{1.08}

\begin{tabular}{cc*{5}{c}}
\toprule
$\boldsymbol{\beta}$
& \textbf{Shot}
& \textbf{R-1}
& \textbf{R-L}
& \textbf{MET.}
& \textbf{CP}
& \textbf{PP} \\
\midrule

\multirow{5}{*}{0}
& 1  & 0.2684 & 0.2510 & 0.3423 & 4.57 & 4.72 \\
& 2  & 0.2703 & 0.2511 & 0.3422 & 4.65 & 4.64 \\
& 4  & 0.2695 & 0.2515 & 0.3444 & 4.74 & 4.55 \\
& 8  & 0.2674 & 0.2502 & 0.3413 & 4.64 & 4.33 \\
& 16 & 0.2740 & 0.2541 & 0.3504 & 4.45 & 4.12 \\
\midrule

\multirow{5}{*}{0.1}
& 1  & 0.3848 & 0.3386 & 0.3957 & 6.28 & 6.40 \\
& 2  & 0.3858 & 0.3379 & 0.3918 & 6.30 & 6.47 \\
& 4  & 0.3899 & 0.3428 & 0.3983 & 5.94 & 6.34 \\
& 8  & 0.3835 & 0.3400 & 0.3945 & 6.51 & 6.74 \\
& 16 & 0.4349 & 0.3774 & 0.4318 & 6.60 & 6.83 \\
\midrule

\multirow{5}{*}{0.01}
& 1  & 0.4162 & 0.3637 & 0.4133 & 5.95 & 6.33 \\
& 2  & 0.4160 & 0.3656 & 0.4142 & 5.92 & 6.28 \\
& 4  & 0.4156 & 0.3639 & 0.4137 & 5.94 & 6.27 \\
& 8  & 0.4162 & 0.3652 & 0.4135 & 6.24 & 6.61 \\
& 16 & 0.4323 & 0.3851 & 0.4280 & 6.44 & 6.72 \\
\midrule

\multirow{5}{*}{0.001}
& 1  & 0.3901 & 0.3381 & 0.4017 & 5.64 & 6.13 \\
& 2  & 0.3881 & 0.3360 & 0.3953 & 6.05 & 6.31 \\
& 4  & 0.3887 & 0.3369 & 0.3978 & 5.87 & 6.36 \\
& 8  & 0.3898 & 0.3371 & 0.3980 & 6.20 & 6.62 \\
& 16 & 0.4265 & 0.3702 & 0.4300 & 6.15 & 6.61 \\

\bottomrule
\end{tabular}
\end{table}

%% file: tab/dz.tex
\begin{table}[!ht]
\centering
\caption{
Sensitivity analysis of the latent adaptation dimension ($d_z=r$) on HiCUPID with Llama-3.1-8B-Instruct across different numbers of observed user interactions.
}
\label{tab:dz_sensitivity}

\small
\setlength{\tabcolsep}{5pt}
\renewcommand{\arraystretch}{1.08}

\begin{tabular}{cc*{5}{c}}
\toprule
$\boldsymbol{d_z}$
& \textbf{Shot}
& \textbf{R-1}
& \textbf{R-L}
& \textbf{MET.}
& \textbf{CP}
& \textbf{PP} \\
\midrule

\multirow{5}{*}{4}
& 1  & 0.4169 & 0.3681 & 0.4217 & 6.12 & 6.52 \\
& 2  & 0.4176 & 0.3702 & 0.4214 & 5.87 & 6.17 \\
& 4  & 0.4195 & 0.3719 & 0.4268 & 5.91 & 6.37 \\
& 8  & 0.4153 & 0.3677 & 0.4204 & 6.32 & 6.64 \\
& 16 & 0.4345 & 0.3837 & 0.4330 & 6.27 & 6.68 \\
\midrule

\multirow{5}{*}{8}
& 1  & 0.3981 & 0.3483 & 0.4191 & 6.02 & 6.47 \\
& 2  & 0.4002 & 0.3497 & 0.4193 & 6.25 & 6.54 \\
& 4  & 0.4027 & 0.3532 & 0.4219 & 6.21 & 6.58 \\
& 8  & 0.3986 & 0.3476 & 0.4172 & 6.55 & 6.88 \\
& 16 & 0.4240 & 0.3775 & 0.4282 & 6.53 & 6.83 \\
\midrule

\multirow{5}{*}{16}
& 1  & 0.4162 & 0.3637 & 0.4133 & 5.95 & 6.33 \\
& 2  & 0.4160 & 0.3656 & 0.4142 & 5.92 & 6.28 \\
& 4  & 0.4156 & 0.3639 & 0.4137 & 5.94 & 6.27 \\
& 8  & 0.4162 & 0.3652 & 0.4135 & 6.24 & 6.61 \\
& 16 & 0.4323 & 0.3851 & 0.4280 & 6.44 & 6.72 \\

\midrule

\multirow{5}{*}{32}
& 1  & 0.3938 & 0.3455 & 0.4003 & 6.16 & 6.42 \\
& 2  & 0.3919 & 0.3411 & 0.3988 & 6.17 & 6.37 \\
& 4  & 0.3889 & 0.3436 & 0.4006 & 6.34 & 6.56 \\
& 8  & 0.3903 & 0.3423 & 0.3990 & 6.37 & 6.67 \\
& 16 & 0.4316 & 0.3789 & 0.4286 & 6.57 & 6.74 \\
\midrule

\multirow{5}{*}{64}
& 1  & 0.3962 & 0.3483 & 0.4120 & 6.10 & 6.51 \\
& 2  & 0.3975 & 0.3514 & 0.4173 & 6.20 & 6.37 \\
& 4  & 0.3989 & 0.3498 & 0.4151 & 6.19 & 6.55 \\
& 8  & 0.4010 & 0.3550 & 0.4203 & 6.54 & 6.96 \\
& 16 & 0.4318 & 0.3784 & 0.4390 & 6.55 & 6.75 \\

\bottomrule
\end{tabular}
\end{table}

%% file: tab/time_ad.tex
\begin{table*}[htbp]
\centering
\caption{
Extended test-time personalization efficiency results on HiCUPID with
Llama-3.1-8B-Instruct. Personalization latency measures the cost required
to incorporate observed user interactions before response generation,
while generation latency measures the subsequent response generation cost.
Lower is better.
}
\label{tab:extended_personalization_efficiency}

\small
\setlength{\tabcolsep}{5.5pt}
\renewcommand{\arraystretch}{1.08}

\begin{tabular}{lcccccccc}
\toprule

\multirow{2}{*}{\textbf{Method}}
& \multicolumn{4}{c}{\textbf{Personalization Latency (s)}}
& \multicolumn{4}{c}{\textbf{Generation Latency (s)}} \\

\cmidrule(lr){2-5}
\cmidrule(lr){6-9}

& \textbf{1-shot}
& \textbf{2-shot}
& \textbf{4-shot}
& \textbf{8-shot}
& \textbf{1-shot}
& \textbf{2-shot}
& \textbf{4-shot}
& \textbf{8-shot} \\

\midrule

ICL
& / & / & / & /
& 2.0896 & 2.1081 & 2.1571 & 2.2302 \\

LoRA
& 0.3718 & 0.3622 & 0.3692 & 0.3640
& 0.6647 & 0.6993 & 0.7357 & 0.7550 \\

PPlug
& 0.0004 & 0.0006 & 0.0009 & 0.0014
& 0.4597 & 0.4283 & 0.4861 & 0.4818 \\

Profile-to-PEFT
& 0.0539 & 0.0538 & 0.0539 & 0.0544
& 0.8758 & 0.9091 & 0.8278 & 0.9599 \\

\textit{LatentPersonal}
& 0.0219
& 0.0304
& 0.0404
& 0.0676
& 0.8996
& 0.9145
& 0.8907
& 0.8838 \\

\bottomrule
\end{tabular}

\vspace{1mm}

\end{table*}

%% file: tab/alpha.tex
\begin{table}[thbp]
\centering
\caption{
Effect of latent scaling factor $\alpha$ across different numbers of observed user interactions on HiCUPID.
}
\label{tab:alpha_scaling}

\small
\setlength{\tabcolsep}{5.5pt}
\renewcommand{\arraystretch}{1.05}

\begin{tabular}{cc*{5}{c}}
\toprule
$\boldsymbol{\alpha}$
& \textbf{Shot}
& \textbf{R-1}
& \textbf{R-L}
& \textbf{MET.}
& \textbf{CP}
& \textbf{PP} \\
\midrule

$0$
& / & 0.1614 & 0.1097 & 0.2166 & 3.57 & 4.29 \\

\midrule

$0.25$
& 1 & 0.3491 & 0.2958 & 0.3513 & 5.12 & 5.60 \\
& 2 & 0.3596 & 0.3109 & 0.3648 & 5.42 & 5.92 \\
& 4 & 0.3575 & 0.3088 & 0.3633 & 5.28 & 5.63 \\
& 8 & 0.3602 & 0.3077 & 0.3644 & 5.67 & 6.05 \\
\midrule

$0.5$
& 1 & 0.3941 & 0.3487 & 0.3972 & 5.75 & 6.42 \\
& 2 & 0.3896 & 0.3470 & 0.3945 & 5.75 & 6.22 \\
& 4 & 0.3989 & 0.3533 & 0.4002 & 6.07 & 6.33 \\
& 8 & 0.3922 & 0.3458 & 0.3922 & 6.08 & 6.67 \\
\midrule

$0.75$
& 1 & 0.3832 & 0.3407 & 0.3891 & 5.72 & 6.28 \\
& 2 & 0.3875 & 0.3466 & 0.3893 & 5.82 & 6.28 \\
& 4 & 0.3847 & 0.3437 & 0.3897 & 5.82 & 6.23 \\
& 8 & 0.3871 & 0.3462 & 0.3907 & 6.13 & 6.72 \\
\midrule

$1$
& 1  & 0.4162 & 0.3637 & 0.4133 & 5.95 & 6.33 \\
& 2  & 0.4160 & 0.3656 & 0.4142 & 5.92 & 6.28 \\
& 4  & 0.4156 & 0.3639 & 0.4137 & 5.94 & 6.27 \\
& 8  & 0.4162 & 0.3652 & 0.4135 & 6.24 & 6.61 \\
\midrule

$1.5$
& 1 & 0.3810 & 0.3413 & 0.3885 & 5.63 & 6.07 \\
& 2 & 0.3838 & 0.3420 & 0.3918 & 5.45 & 6.03 \\
& 4 & 0.3823 & 0.3403 & 0.3899 & 5.78 & 6.23 \\
& 8 & 0.3855 & 0.3446 & 0.3929 & 5.93 & 6.58 \\
\midrule

$2$
& 1 & 0.3879 & 0.3408 & 0.3869 & 5.63 & 6.20 \\
& 2 & 0.3739 & 0.3304 & 0.3774 & 5.72 & 6.30 \\
& 4 & 0.3735 & 0.3293 & 0.3704 & 5.77 & 6.07 \\
& 8 & 0.3698 & 0.3247 & 0.3760 & 5.97 & 6.50 \\

\bottomrule
\end{tabular}
\end{table}

%% file: tab/user.tex
\begin{table}[thbp]
\centering
\caption{
User-specificity analysis across different numbers of observed user interactions on HiCUPID.
}
\label{tab:user_specificity}

\small
\setlength{\tabcolsep}{5.5pt}
\renewcommand{\arraystretch}{1.05}

\begin{tabular}{lc*{5}{c}}
\toprule
\textbf{Latent}
& \textbf{Shot}
& \textbf{R-1}
& \textbf{R-L}
& \textbf{MET.}
& \textbf{CP}
& \textbf{PP} \\
\midrule

\textbf{Matched} $\mathbf{z}_u$
& 1 & 0.4162 & 0.3637 & 0.4133 & 5.95 & 6.33 \\
& 2 & 0.4160 & 0.3656 & 0.4142 & 5.92 & 6.28 \\
& 4 & 0.4156 & 0.3639 & 0.4137 & 5.94 & 6.27 \\
& 8 & 0.4162 & 0.3652 & 0.4135 & 6.24 & 6.61 \\

\midrule

Swapped $\mathbf{z}_v$
& 1 & 0.3373 & 0.2965 & 0.3501 & 5.48 & 5.77 \\
& 2 & 0.3330 & 0.2841 & 0.3478 & 5.54 & 5.84 \\
& 4 & 0.3114 & 0.2818 & 0.3441 & 5.45 & 5.72 \\
& 8 & 0.3084 & 0.2765 & 0.3329 & 5.42 & 5.79 \\

\midrule

Mean $\bar{\mathbf{z}}$
& 1 & 0.3792 & 0.3245 & 0.3781 & 5.64 & 5.89 \\
& 2 & 0.3841 & 0.3285 & 0.3804 & 5.62 & 5.94 \\
& 4 & 0.3828 & 0.3398 & 0.3881 & 5.69 & 5.91 \\
& 8 & 0.3853 & 0.3357 & 0.3956 & 5.63 & 5.87 \\

\midrule

Shared LoRA
& / & 0.3712 & 0.3039 & 0.3425 & 5.38 & 5.34 \\

\bottomrule
\end{tabular}
\end{table}

%% file: tab/peft_evaluation.tex
\begin{table*}[p]
\centering
\caption{
Evaluation with alternative PEFT architectures on HiCUPID with
Llama-3.1-8B-Instruct across different numbers of observed user interactions.
}
\label{tab:peft_evaluation}

\small
\setlength{\tabcolsep}{5pt}
\renewcommand{\arraystretch}{1.08}

\begin{tabular}{lc*{5}{c}}
\toprule
\textbf{Method}
& \textbf{Shot}
& \textbf{R-1}
& \textbf{R-L}
& \textbf{MET.}
& \textbf{CP}
& \textbf{PP} \\
\midrule

\multirow{4}{*}{Adapter}
& 1  & 0.3627 & 0.3063 & 0.3591 & 6.04 & 6.39 \\
& 2  & 0.3848 & 0.3272 & 0.3749 & 6.34 & 6.48 \\
& 4  & 0.3917 & 0.3373 & 0.3876 & 6.35 & 6.48 \\
& 8  & 0.3944 & 0.3346 & 0.3748 & 6.23 & 6.59 \\
\midrule

\multirow{4}{*}{Adapter + \textit{LatentPersonal}}
& 1  & 0.3980 & 0.3468 & 0.4139 & 6.19 & 6.41 \\
& 2  & 0.4074 & 0.3536 & 0.4191 & 6.23 & 6.43 \\
& 4  & 0.4018 & 0.3521 & 0.4127 & 6.29 & 6.39 \\
& 8  & 0.3978 & 0.3483 & 0.4112 & 6.29 & 6.50 \\
\midrule

\multirow{4}{*}{Prompt Tuning}
& 1  & 0.3215 & 0.2940 & 0.2978 & 5.54 & 5.47 \\
& 2  & 0.3534 & 0.3323 & 0.3231 & 5.76 & 5.59 \\
& 4  & 0.3362 & 0.3114 & 0.3064 & 6.00 & 5.66 \\
& 8  & 0.3757 & 0.3455 & 0.3364 & 6.05 & 5.64 \\
\midrule

\multirow{4}{*}{Prompt Tuning + \textit{LatentPersonal}}
& 1  & 0.3712 & 0.3469 & 0.3399 & 5.76 & 5.76 \\
& 2  & 0.3691 & 0.3476 & 0.3359 & 5.85 & 5.66 \\
& 4  & 0.3686 & 0.3417 & 0.3327 & 5.86 & 5.58 \\
& 8  & 0.3729 & 0.3524 & 0.3393 & 5.94 & 5.42 \\
\bottomrule
\end{tabular}
\end{table*}

%% file: appendix/app_peft_instantiations.tex
\newpage

\section{\textit{LatentPersonal} across PEFT Architectures}
\label{app:peft_architectures}

While our main implementation adopts LoRA as the primary PEFT parameterization, the latent navigation principle of \textit{LatentPersonal} is not tied to a specific PEFT parameterization. In this section, we briefly illustrate how the same formulation can be instantiated with other representative PEFT architectures, including Adapter and Prompt Tuning modules.

\subsection{Adapter Instantiation}
\label{app:adapter_instantiation}

For a standard Adapter, the hidden representation at layer $l$ is updated as
\begin{equation}
\tilde{\mathbf{h}}^{(l)}
=
\mathbf{h}^{(l)}
+
W_{\mathrm{up}}^{(l)}
\sigma\left(
W_{\mathrm{down}}^{(l)}\mathbf{h}^{(l)}
\right),
\end{equation}
where $W_{\mathrm{down}}^{(l)}$ and $W_{\mathrm{up}}^{(l)}$ denote the down- and up-projection matrices, respectively, and $\sigma(\cdot)$ denotes a nonlinear activation function.

The Adapter architecture naturally induces a low-dimensional intermediate space through $W_{\mathrm{down}}^{(l)}$, which is directly compatible with the latent representation in \textit{LatentPersonal}. We therefore instantiate \textit{LatentPersonal} with Adapter by introducing the user latent $z_u$ into this intermediate space:
\begin{equation}
\tilde{\mathbf{h}}_u^{(l)}
=
\mathbf{h}^{(l)}
+
W_{\mathrm{up}}^{(l)}
\operatorname{diag}(z_u)
\sigma\left(
W_{\mathrm{down}}^{(l)}\mathbf{h}^{(l)}
\right).
\end{equation}
The dimensionality of $z_u$ is matched to the Adapter bottleneck dimension, such that each latent component modulates the corresponding dimension of the intermediate representation. The preference encoder for obtaining $z_u$, the training objective, and the optimization procedure remain identical to those used in the LoRA instantiation.

\subsection{Prompt Tuning Instantiation}
\label{app:prompt_instantiation}

For standard Prompt Tuning, a set of learnable continuous prompt embeddings is introduced at the input embedding level:
\begin{equation}
P
=
\left[
p_1;
p_2;
\ldots;
p_m
\right]
\in
\mathbb{R}^{m \times d},
\end{equation}
where $m$ denotes the number of soft prompt tokens and $d$ is the hidden dimension of the pretrained model. Given an input sequence $x$ with token embeddings $E(x)$, the model input is $[P;E(x)]$. During adaptation, only $P$ is optimized while the model parameters remain frozen.

The soft prompt provides a natural component space for incorporating the latent representation in \textit{LatentPersonal}. We therefore introduce the user latent $z_u$ along the prompt-token dimension:
\begin{equation}
P_u
=
\operatorname{diag}(z_u)P,
\end{equation}
or equivalently,
\begin{equation}
P_u
=
\left[
z_{u,1}p_1;
z_{u,2}p_2;
\ldots;
z_{u,m}p_m
\right].
\end{equation}
The dimensionality of $z_u$ is matched to the number of soft prompt tokens, such that each latent component modulates the corresponding prompt embedding. The resulting $P_u$ replaces $P$ in the input representation, while the preference encoder for obtaining $z_u$, the training objective, and the optimization procedure remain identical to those used in the LoRA instantiation.

%% file: appendix/app_limitation.tex
\section{SCOPE AND LIMITATIONS}

Our results support latent navigation as an effective formulation for few-shot personalization, but they do not distinguish whether the underlying personalization-relevant structure is inherited from pretraining or emerges during population-level personalization training. In addition, our evaluation focuses on benchmark-based personalization settings with limited observed interactions; understanding how the learned latent representations behave under longer-term and evolving user preferences remains an important direction for future work.